\documentclass[10pt,twocolumn,letterpaper]{article}

\usepackage[pagenumbers]{cvpr} % To force page numbers, e.g. for an arXiv version

\usepackage[dvipsnames]{xcolor}

\definecolor{cvprblue}{rgb}{0.21,0.49,0.74}
\usepackage[pagebackref,breaklinks,colorlinks,citecolor=cvprblue]{hyperref}
\usepackage{multirow}
\usepackage{tikz}
\usepackage{makecell}

\def\paperID{4095} % *** Enter the Paper ID here
\def\confName{CVPR}
\def\confYear{2024}
\newcommand{\st}[2]{#2}

\title{DiaLoc: An Iterative Approach to Embodied Dialog Localization}

\author{Chao Zhang
\qquad  Mohan Li
\qquad  Ignas Budvytis
\qquad  Stephan Liwicki
\\
Toshiba Europe Ltd \\
{\tt\small firstname.lastname@toshiba.eu}
}

\begin{document}
\maketitle%%%%%%%%% ABSTRACT
\begin{abstract}
    % Submission: Nov 17th, 2023 
    Multimodal learning has advanced the performance for many vision-language tasks. However, most existing works in embodied dialog research focus on navigation and leave the localization task understudied. 
    \st{Furthermore, existing}{The few existing} dialog-based localization approaches assume the availability of entire dialog prior to localizaiton, which \st{contrasts with human behavior}{is impractical for deployed dialog-based localization}. In this paper, we propose DiaLoc, a new dialog-based localization framework which aligns with \st{real-world application perspectives}{a real human operator behavior}. \st{}{Specifically, we produce an iterative refinement of location predictions which can visualize current pose believes after each dialog turn.} DiaLoc effectively utilizes the multimodal data for multi-shot localization, where a fusion encoder fuses vision and dialog information iteratively. We achieve state-of-the-art results on embodied dialog-based localization task, in single-shot (+$7.08\%$ in Acc5@valUnseen) and multi-shot settings (+$10.85\%$ in Acc5@valUnseen). DiaLoc narrows the gap between simulation and real-world applications, opening doors for future research on collaborative localization and navigation.
    
\end{abstract}

%%%%%%%%% BODY TEXT
% \input{section/todo}
\section{Introduction}

% problem definition
Picture yourself becoming lost within a novel building during a visit to a friend's place of residence or work place. In this scenario, you reach out to your friend and detail the elements of your environment. Through a series of exchanges, it is anticipated that your friend will eventually figure out your whereabouts.  The completion of this collaborative endeavor depends on the skillful use of targeted inquiries and unequivocal linguistic responses. 
\st{Within this framework, we assume the involvement of two agents:}{In this study, we formulate this task as an iterative embodied dialog localization problem involving two agents:} an Observer, randomly situated within an environment, and a Locator,  whose job is to localize the observer with a provided top-down map through dialog.

\st{The task of employing dialogue and a top-down map to determine an agent's precise location presents two challenges. Firstly, its utilization of multimodal information to overcome information imbalances stemming from egocentric and top-down viewpoints. Secondly, the dataset WAY~\cite{hahn2020you}, sourced from human annotations, is rather limited in scale. Consequently, this data scarcity restrics the capacity of trained models to generalize to unfamiliar environments. Despite these challenges, this task bears the potential for numerous real-world applications, including cooperative navigation~\cite{wang2021collaborative}\cite{yu2022learning} and the utilization of mobile robots for search and rescue operations~\cite{queralta2020collaborative}. Nonetheless, it is worth noting that dialog-based localization, in comparison to navigation, is still a relatively understudied area.}{Iterative embodied dialog localization is of clear relevance to numerous real-world applications, including the utilization of mobile robots for search and rescue operations~\cite{queralta2020collaborative, manuel2022robot}. Nevertheless, it is worth noting that dialog-based localization, in comparison to navigation, is still a relatively understudied area. 
%\textcolor{red}{TODO: why is localization alone relevant? Would navigation not solve it all?} 
The dataset WAY  sourced dialogs from human annotations and formulated dialog-based localization through employing a full dialogue and a top-down map to determine an agent's precise location. The sparsity and problem formulation pose however two challenges: Firstly, its utilization of multimodal information to overcome information imbalances stemming from egocentric and top-down viewpoints. Secondly, the dataset's scarcity restricts the capacity of trained models to generalize to unfamiliar environments.}

%%%%%%%%%%%%%%%%%%%%%%%%%%%%%%%%%%%%%%%%%%%%%%%%%%%%%%%%%%%%%
\begin{figure}[t]
    \centering
    \includegraphics[width=0.9\linewidth]{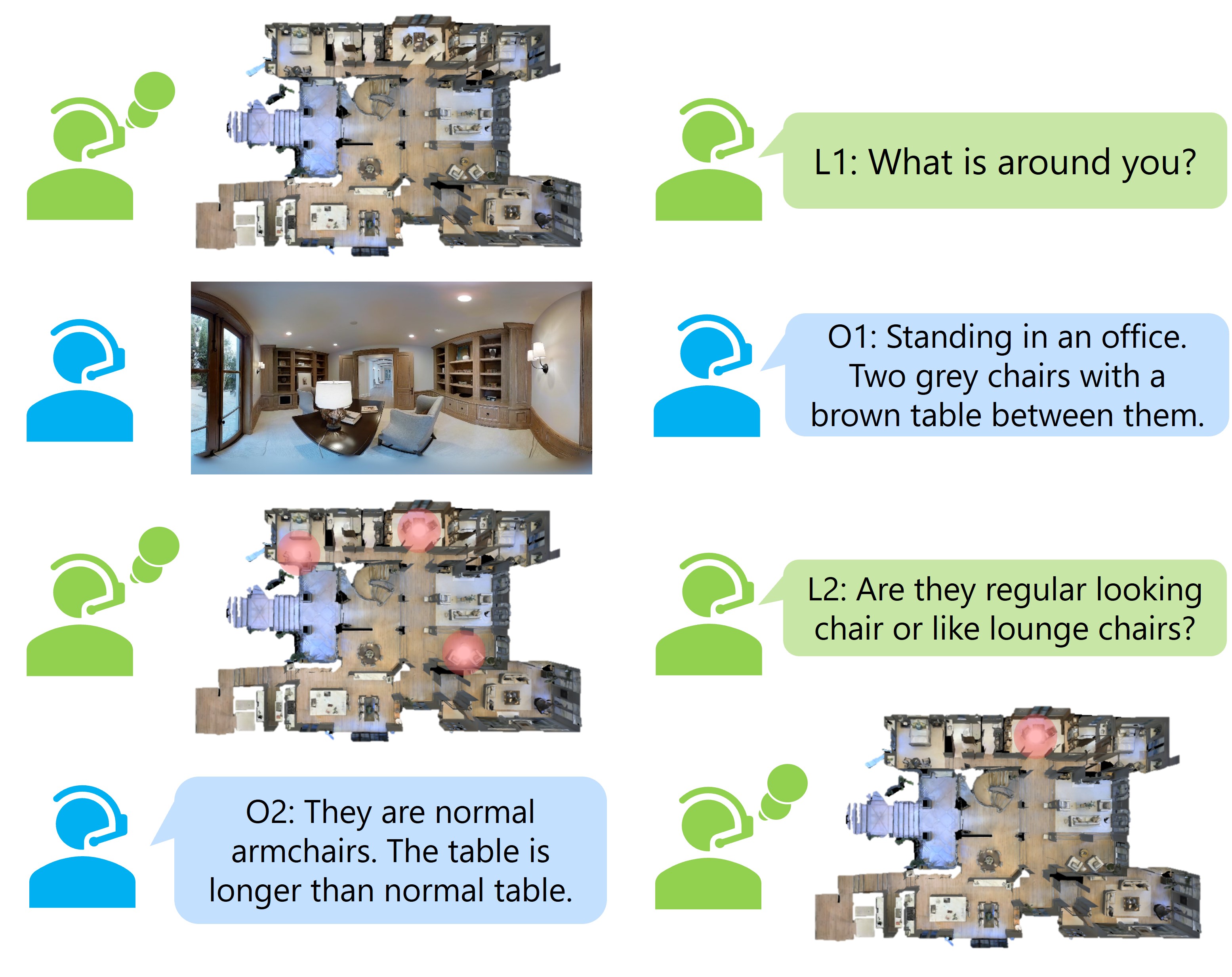}
    \caption{\textbf{Illustration of iterative embodied dialog localization.} The locator with top-down map and the observer with egocentric view engage in a cooperative dialog to assist in determining the observer's location. The locator iteratively forms predictions to enhance the estimations. The preceding dialog exchanges and cumulative predictions also influence the manner in which the locator poses new questions. As depicted, the locator predicts 3 possible locations based on the first turn. The second question is asked to disambiguate the predictions and the correct location is predicted given the answer from the observer.  }
    \label{fig:teaser}
\end{figure}
%%%%%%%%%%%%%%%%%%%%%%%%%%%%%%%%%%%%%%%%%%%%%%%%%%%%%%%%%%%%%

% reformulat the problem 
In an initial attempt to address the challenge of localization through embodied dialog, Hahn \etal \cite{hahn2020you} introduced a novel approach by framing the task as an image-to-image problem. They adopted LingUNet network to predict localization distribution. Language embedding generated by BiLSTM~\cite{huang2015bidirectional} is used as conditional input to guide the learning of visual features. Leveraging a dataset of human annotations, their approach assumes the availability of the complete dialog prior to making predictions. However, this approach contrasts with human intuition, as human localization predictions are characterized by incremental refinement. As illustrated in Figure~\ref{fig:teaser}, the locator actively makes location guesses as new dialog information arrives. This dynamic process is also reflected in questioning: the locator poses new questions based on prior dialog exchanges and their cumulative predictions. Therefore, developing a localization technique capable of continual location estimation stands as a fundamental cornerstone for realizing the practical application of cooperative search and rescue.

% the benefits
\st{In this study, we introduce a novel task: iterative embodied dialog localization}{Inspired by cooperative search and rescue we therefore introduce a novel approach which we call \emph{iterative embodied dialog localization}}. This formulation requires the locator to possess the capability of finding out the observer's location following each turn within the dialog. When compared to existing approaches that consider the complete dialog, our iterative framework offers several merits. Firstly, it operates with greater efficiency. During the inference phase, it only processes a single turn of the dialog, as opposed to the entire conversation. In practical application, this efficiency can translate into time savings, or life savings, as the location can be estimated or narrowed down to a few points before the conversation finishes. Secondly, our approach exhibits enhanced generalization to novel locations. The risk of over-fitting to training data is mitigated by not relying on complete dialogs during the training process. Lastly, the intermediate predictions provide vital cues for assessing uncertainty and are indispensable for tasks that involve dialog generation,  such as embodied visual dialog and cooperative localization. 
In summary, this transition towards an iterative dialog-based localization approach not only aligns with engineering perspectives but also paves the way for improved performance and broader applicability across various related tasks.

% the approach: 
Multimodal learning, particularly through the integration of vision and language,  has emerged as a prominent topic in deep learning era. The paradigm of Vision-Language Pre-training (VLP)~\cite{li2022blip, li2021align} has advanced the performance of numerous vision-language tasks. VLP employs Transformers to acquire either unimodal representations or fused multimodal representations. 
Inspired by VLP's success in multiple vision language understanding and generation tasks,  such as image captioning and Visual Question Answering(VQA), we embrace the Transformer-based encoders. This forms the backbone for our novel multimodal multi-shot localizer, which becomes the state of the art. 
%BLIP \cite{li2022blip} was a VLP framework which successfully transfers to multiple vision language understanding and generation tasks, such as image-text retrieval, image captioning, and visual question answering. Following BLIP, we use Transformer-based encode-decoder architecture for our method.  

% sum up contributions
In summary, we make the following \textbf{contributions}: 
(i) We introduce an iterative approach towards practical embodied dialog localization. This formulation aligns with \st{that the}{the intuitive behavior of a} human locator \st{}{who} employs incremental predictions to improve localization performance. 
(ii) We propose a novel multimodal multi-shot localizer DiaLoc based on Transformer encoders.
(iii) We demonstrate state-of-the-art performance for embodied dialog localization. Our proposed iterative solution exhibits enhanced generalization capabilities.
(iv) We showcase that the proposed approach posses the ability to rectify previous prediction with new information \st{}{and may help a human operator in question formulation for search and rescue applications}. 
%%%%%%%%%%%%%%%%%%%%%%%%%%%%%%%%%%%%%%%%%%%%%%%%%%%%%%%%%%%%%

\section{Related Work}

% visual language learning: VLN (CE), blip, flamingo, robotic 
\paragraph{Vision and Language}
%\st{}{TODO I would swap these 2. First talk about Vision and Language, then about Embodied Localization... or Vision and Language for Localization should come at end of this part}
Vision and language understanding has been extensively explored across tasks such as image captioning, visual question answering (VQA) and visual grounding~\cite{brown2020language, alayrac2022flamingo, wang2023image, driess2023palm}. The fusion of language and vision for recognition is also an active domain. Notably, CLIP~\cite{radford2021learning} effectively learns visual concepts through natural language supervision. The fundamental insight is to employ natural language as a versatile space for predictions, facilitating zero-shot generalization and transfer learning.

In the context of language-driven tasks, LSeg~\cite{li2022languagedriven} was proposed for language-driven semantic image segmentation using flexible text embeddings. It allows LSeg to generalize to unseen categories at test time. Delving into the intersection of vision and language with navigation, Visual and language navigation (VLN) combines visual and linguistic inputs to enhance robot navigation in building-scale environments~\cite{anderson2018vision}. Moreover, vision and language navigation in continuous environments (VLN-CE) is an instruction-guided navigation task. It relaxes the assumptions inherent in the original VLN task, striving to bridge the gap between simulated environments and real-world scenarios. Instead of utilizing a labeled dataset with trajectories annotations, LM-Nav~\cite{shah2023lm} shows that a robotic navigation system can be constructed purely from pre-trained models using CLIP and GPT-3~\cite{brown2020language} without any fine-tuning or language-annotated robot data. 

% multi-modal localization 
\noindent\textbf{Embodied Localization}
Estimating an agent's position within the environment forms a vital aspect of applications in embodied AI. 
Hahn \etal \cite{hahn2020you} introduced the task of localization from embodied dialog (LED), leveraging the WAY dataset containing approximately 10,000 dialogs curated by human operators. The authors also proposed a baseline for LED based on LingUnet~\cite{misra2018mapping}, using a top-down map as input to predict location. Choosing an effective map representation is a key consideration in this task. Unlike \cite{hahn2020you}, Hann \etal \cite{hahn2022transformer} proposed an Transformer-based method using navigation graph extracted from Matterport3D~\cite{chang2017matterport3d}. In a similar vein to LED, Cooperative Vision-and-Dialog Navigation (CVDN) \cite{thomason2020vision} introduces a dataset consisting of dialogues where an oracle assists a navigator in completing a navigation task. The baseline approach adopts a  sequence-to-sequence approach. Both WAY and CVDN are constructed upon the Matterport3D dataset, albeit with different focuses: WAY concentrates on localization, while CVDN focus on navigation. Further, RobotSlang~\cite{banerjee2021robotslang} brings a dataset that contains 169 dialogs between a human directing a robot and another human offering guidance towards navigation goals. Beyond multimodal localization,
Text2Pos~\cite{kolmet2022text2pos} delves into cross-modal text to point cloud localization based on Kitti360~\cite{liao2022kitti}. Within this context, the proposed approach identifies positions specified by template-based language descriptions of immediate surroundings within the environment represented as a point cloud. 
% heatmap regression
%Human pose estimation could also be solved as heatmaps estimation. Existing methods typically improve the quality of heatmaps with customized architectures, such as high-resolution representation~\cite{sun2019deep, cheng2020higherhrnet} and vision transformers~\cite{yang2021transpose, xu2022vitpose}. Diffusion models are also used in pose estimation for 2D and 3D~\cite{gong2023diffpose, qiu2023learning}. 
%Another related application is image segmentation. Methods using diffusion models are SegDiff and MedSegDiff.
%DiffusionDet formulates object detection as a denoising diffusion process from noisy boxes to object boxes.

%\st{}{TODO: Why localization and not navigation. What Vision language approach to use for localization?}

%%%%%%%%%%%%%%%%%%%%%%%%%%%%%%%%%%%%%%%%%%%%%%%%%%%%%%%%%%%%%

\section{Iterative Embodied Dialog Localization}
We first describe the task of iterative embodied dialog based localization. Then, the architecture of our proposed model is introduced. These are followed by the training objectives and implementation details.

%%%%%%%%%%%%%%%%%%%%%%%%%%%%%%%%%%%%%%%%%%%%%%%%%%%%%%%%%%%%%
\begin{figure*}[t]
    \centering
    \includegraphics[width=0.85\linewidth]{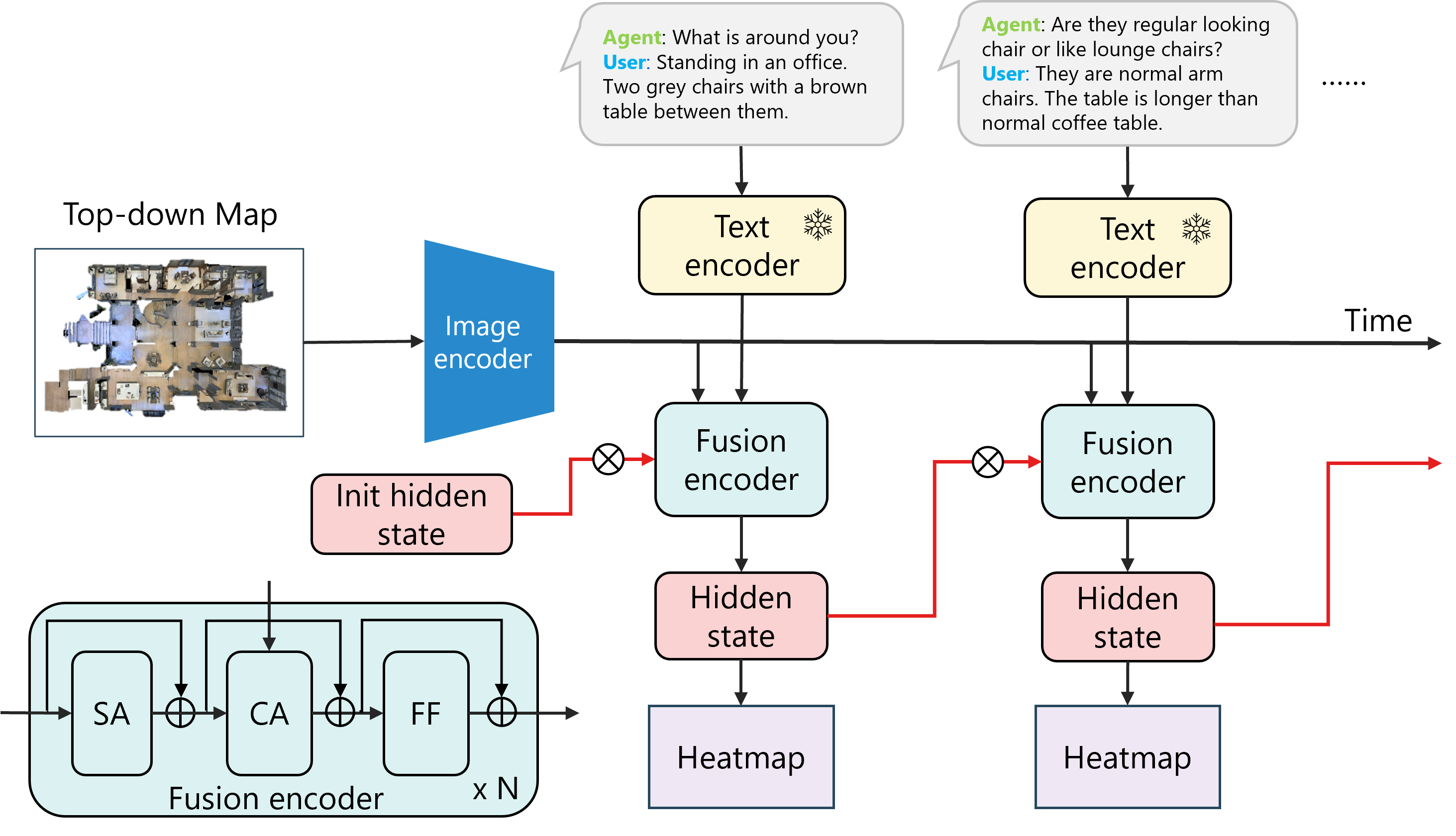}
    \caption{\textbf{DiaLoc-e: the proposed multi-shot multimodal architecture for embodied dialog localization.} The approach employs the image encoder and the frozen text encoder to derive visual and linguistic unimodal embeddings. The fusion encoder integrates the unimodal inputs to update the hidden state. Multi-shot predictions are produced using the hidden state at varying timesteps. The fusion encoder comprises $N$ blocks of Transformer encoder with cross attention layer. }
    \label{fig:arch1}
\end{figure*}
%%%%%%%%%%%%%%%%%%%%%%%%%%%%%%%%%%%%%%%%%%%%%%%%%%%%%%%%%%%%%

\subsection{Task and Evaluation Metrics}
%WAY dataset is used to evaluate our proposed method against multiple baselines. To provide a human-performance baseline and gather training data for agents, the dataset was collected from human localization dialogs. 2020 episodes across 87 environments are generated by rejection sampling to avoid spatial redundancy. For each environment, start locations were iteratively sample, rejecting ones that are within 5m of already-sampled positions. Using this dataset, models are trained with KL-divergence loss to minimize the predicted location distribution and the ground-truth location. To evaluate the performance, localization error (LE) defined as the geodesic distance in meters is used. Furthermore, binary success rates for 3m and 5m are also reported. 

% the new task: 
\st{We perform embodied localization using dialogs $D$ and the global map $M$. The human dialog is represented by $T$ turns between the locator and the observer $(L_1, O_1, ..., L_{T}, O_{T})$.}{We perform embodied localization using dialogs $D = (L_1, O_1, ..., L_{T}, O_{T})$ and the global map $e$, here the human dialog is represented by $T$ turns between the utterance of the locator $L_i$ and the observer $O_i$.}
The goal is to predict the observer's final location $p_T$. Different from the previous approaches making a single prediction at the end of the whole dialog, 
the proposed iterative approach outputs multiple predictions for each turn $p_1,..., p_{T}$.  The goal is to enhance communication efficacy and spatial understanding through a series of information exchanges between the locator (top-down view) and the observer (egocentric). 
% evaluations metrics
To evaluate the localization error (LE), we follow \cite{hahn2022transformer} to use geodesic distance instead of euclidean distance, as geodesic distance is more meaningful for determining error across rooms. Pixel coordinate $p_T$ of the location is snapped to the nearest way-point node $g_T$ before the calculation of localization error as: $\text{LE}=||g_T - \hat{g}_T||^{2}$, where $g_T$ is the ground-truth and $\hat{g}_T$ is the prediction. 
Binary success rate for LE at threshold $k$ is also reported for represented values such as 0m and 5m. We draw Cumulative Matching Characteristics (CMC) curves to compare different approaches to characterize the holistic performance.

% dataset split
The WAY dataset~\cite{hahn2020you} is used to evaluate the proposed method. It is split into train (9,955), valSeen (305), valUnseen (579) and test sets (1,200). There exists two evaluation protocols: floor-level ~\cite{hahn2020you} and building-level~\cite{hahn2022transformer}. 
As opposed to non-iterative methods, our proposed method focus on exploiting the sequential context from a dialog and has the potential of stopping the dialogs when predictions are accurate. Achieving the early-stopping requires the turn-based locations since the observer is allowed to move during the dialogs. To further improve the training efficiency, our method follows ~\cite{hahn2020you} and use the floor-based setting assuming the known floor. We found that neither the target floor information or the per-turn location is available from the test split and hence not feasible for the evaluation purpose. As such, all methods are trained using train split and validated using valSeen, and the performance on valUnSeen is used to show the generalization in novel scenes.

There are two advantages of evaluating on the dialogs with known floor and turn-based dense locations: (1) Comparing to methods that use all floors during training, evaluating using the target floor map is efficient and easier for us to investigate the iterative improvement; (2) With corresponding locations at each turn of the dialog, we could measure the localization performance at different timesteps. This is useful to quantify the potential of early termination.

% advantages
Compared to non-iterative embodied dialog localization, the new iterative approach offers improved adaptability and accuracy. It enables agents to progressively refine their spatial understanding through ongoing dialogs, accommodating dynamic environments representation and nuanced contextual information. This contrasts with the previous single-shot methods, and allows for more effective training, better generalization to unseen environments, and ultimately more practical embodied applications.

\subsection{Architecture}\label{sec:arch}
% overview
Previously, LED~\cite{hahn2020you} formalized embodied dialog localization as one image-to-image translation problem. The whole dialog is used as conditional input to the encoder-decoder network. Unlike LED, we argue that multi-turn predictions happened at the end of each turn are crucial for the applications such as search and rescue. This also resembles the way how human performs dialog-based localization. To reduce this gap, we adopt Transformer~\cite{vaswani2017attention} backbone for the new task.  Inspired by recent work on multimodal learning~\cite{li2022blip, li2021align}. Our design features an image backbone which interacts with dialog input iteratively. Location predictions are made at multiple timesteps. We now describe the network architecture details \st{}{as shown in Figure~\ref{fig:arch1}}.

\textbf{Unimodal Encoders.} To obtain unimodal embeddings, we use Transformer-based encoder for both vision and language inputs. Vision transformers (ViT) pre-trained on ImageNet is used as the image encoder for top-down map $e$. We fine-tune ViT to generate the visual embedding $V\in R^{M, C}$, where $M=196$ is the number of visual tokens. Input map is resized to $224\times 224$. To encode dialog $D=(L_1, O_1,\cdots, L_T, O_T)$ of $T$ turns, we adopt the pre-trained Bert~\cite{devlin2018bert} as the text encoder. The dialog $D$ is represented as features $L\in R^{N, C}$, where $N=100$ is the max token length. Note that the feature dimension $C$ is 768 for both ViT and BERT.  

\textbf{Multimodal Encoder.} Given visual and text embeddings $V$ and $L$, we employ a stack of Transformer  blocks as the multimodal encoder $\Phi$. Each block consists of 12 sequential modules of self-attention (SA), cross-attention (CA) and feed-forward (FF) layers. The cross-attention layer is introduced for the purpose of multimodal fusion. Specifically, text embedding of current turn $L_t$ is used to create query $Q$ via a MLP, while the map embedding $V$ provides key $K$ and value $V$. The output of multimodal encoder is hidden state $S\in R^{M, C}$ and has the same dimension as visual embedding $V$. A detailed illustration of the fusion encoder is shown in Figure \ref{fig:arch1}. 

\textbf{Prediction Head.} Multimodal encoder fuses dialog input and visual map representations to output the current state $S_t$ at turn $t$. Using $S$ as input, a prediction head including multiple convolution and deconvolution layers is trained to produce location heatmap $H\in R^{h, w}$. The heatmap $H$ is finally up-sampled to the same size as target $\Tilde{H}\in R^{h_0, w_0}$ for loss caculation. At inference stage, the predicted location in image space is obtained via $\hat{p}=\text{Argmax}(\text{Softmax}(H))$.

\begin{figure}[t]
    \centering
    \includegraphics[width=0.98\linewidth]{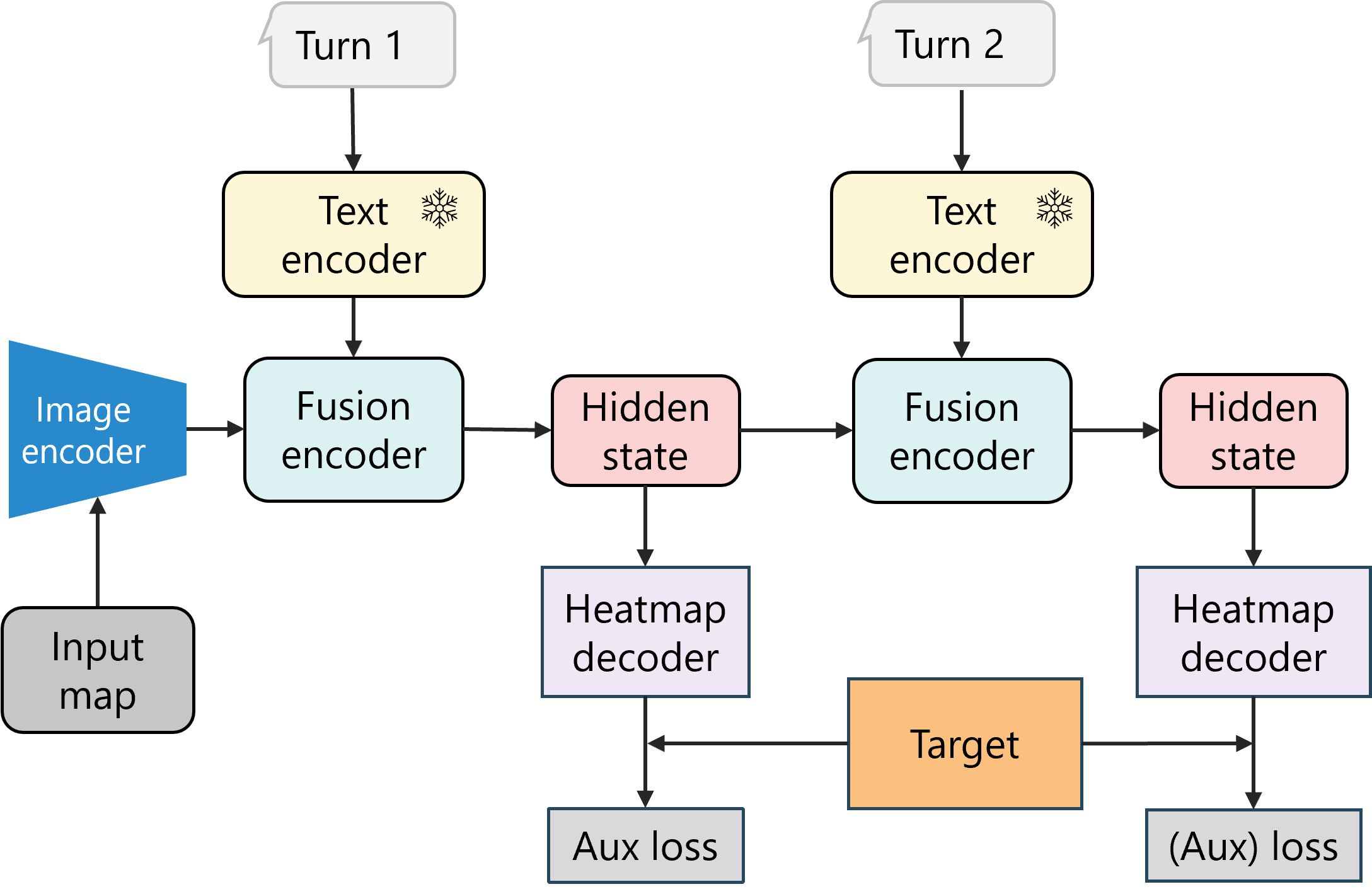}
    \caption{\textbf{DiaLoc-i: the proposed localizer variant with implicit fusion design.} The hidden state is continuously updated with dialog information at each timestep.  }
    \label{fig:arch2}
\end{figure}

\textbf{Multi-shot Multimodal Localizer.}
We now discuss two fusion variants for multi-shot localization. The distinction mainly lies in how previous hidden state $S_{t-1}$ are intergrated with new dialog input $L_t$. In Figure \ref{fig:arch1}, the \emph{explicit} variant,  dubbed \textbf{DiaLoc-e},  is illustrated. The map embedding $V$ is directly processed at each timestep and is fused with the hidden state $S$ before being fed into the fusion encoder via: $S_t = \Phi(V\odot S_{t-1}, L_t)$. The hidden state $S$ is initialized to $\mathbf{1}$ and serves as the prior for the visual map.  For the \emph{implicit} variant, \textbf{DiaLoc-i}, depicted in Figure~\ref{fig:arch2}, it iteratively updates the hidden state via $S_t=\Phi(S_{t-1}, L_t)$ thus ensuring that the final state $S_T$ are conditioned on all dialog turns $(L_1,..,L_T)$.

%%%%%%%%%%%%%%%%%%%%%%%%%%%%%%%%%%%%%%%%%%%%%%%%%%%%%%%%%%%%%
%%%%%%%%%%%%%%%%%%%%%%%%%%%%%%%%%%%%%%%%%%%%%%%%%%%%%%%%%%%%%

\subsection{Loss Function}
Starting with single-shot case, given the prediction $H$ and the target $\Tilde{H}$, we train the model to minimize the KL-divergence between the predicted location distribution and the ground-truth location. Similar to \cite{hahn2020you}, Gaussian smoothing with standard deviation of 3m is applied to the ground-truth $\Tilde{H}$. The single-shot loss function is: 
%%%%%%%%%%%%%%%%%%%%%%%%%%%%%%%%%%%%%%%%%%%%%%%%%%%%%%%%%%%%%
\begin{equation}
    L_{ss}(H, \Tilde{H}) = \log(\Tilde{H}) (\log\Tilde{H}-\log (\text{Softmax}(H)))
\end{equation}
%%%%%%%%%%%%%%%%%%%%%%%%%%%%%%%%%%%%%%%%%%%%%%%%%%%%%%%%%%%%%

 To adapt to our multi-shot approach, we simply apply single-shot loss to all predictions $\textbf{H}=H_{t=1}^T$ made at $t\in [1, T]$. Additionally, the multiple losses are weighted summed to provide the following multi-shot loss: 

%%%%%%%%%%%%%%%%%%%%%%%%%%%%%%%%%%%%%%%%%%%%%%%%%%%%%%%%%%%%%
\begin{equation}
    L_{ms}(\textbf{H}, \Tilde{H}) = \frac{1}{T} \sum_{t=1}^T L_{ss}(H_t, \Tilde{H}) \alpha^{T-t}
\end{equation}
%%%%%%%%%%%%%%%%%%%%%%%%%%%%%%%%%%%%%%%%%%%%%%%%%%%%%%%%%%%%%
where $\alpha \in [0,1]$ is the decay factor. The idea of applying decay is to penalize less for early predictions due to incomplete dialog context.

However, there is still a risk of over-fitting when intermediate predictions $H_t$ where $t<T$ are toned down with the decay factor. \st{}{Specifically we note that it is often unreasonable to find the true location after a single, ambiguous dialog turn. Unfortunately, however,} The KL-divergence loss with the ground-truth target encourages peaky predictions and leads to over-confident prediction at early timesteps. To alleviate this issue, we introduce an auxiliary loss $L_{aux}$. The aim of this term is to promote the diversity of early predictions \st{. In turn this enables more false negatives survive}{and avoid unreasonable early filtering of false positives}. The loss function is defined as:
%%%%%%%%%%%%%%%%%%%%%%%%%%%%%%%%%%%%%%%%%%%%%%%%%%%%%%%%%%%%%
\begin{equation}
    L_{aux}(\textbf{H}, \Tilde{H}) = \frac{1}{T} \sum_{t=1}^T L_{mse}(\text{Sigmoid}(H_t)\odot \Tilde{H}_{mask}, \Tilde{H})
\end{equation}
\label{eq:aux}
%%%%%%%%%%%%%%%%%%%%%%%%%%%%%%%%%%%%%%%%%%%%%%%%%%%%%%%%%%%%%
where $\Tilde{H}_{mask}$ is the binary mask of target where $\Tilde{H}>0$. Sigmoid function is applied to $H_t$ to produce the probability scores of location. Mean squared loss $L_{mse}$ is used to measure the prediction error. 

We combine the multi-shot loss $L_{ms}$ and the auxiliary loss $L_{aux}$ to get the final loss:
%%%%%%%%%%%%%%%%%%%%%%%%%%%%%%%%%%%%%%%%%%%%%%%%%%%%%%%%%%%%%
\begin{equation}
    L(\textbf{H}, \Tilde{H}) = L_{ms}(\textbf{H}, \Tilde{H}) + \beta L_{aux}(\textbf{H}, \Tilde{H})
\end{equation}
%%%%%%%%%%%%%%%%%%%%%%%%%%%%%%%%%%%%%%%%%%%%%%%%%%%%%%%%%%%%%
where $\beta$ as the weight factor. We set $\beta=1.0$ for our method except ablation studies.

%Considering the limitations of LED approach, we reformulate the localization task as a sequential prediction task. Given a dialog and the corresponding top-down map, our method makes predictions at the end of each turn. This differs with the LED method which makes one prediction after seeing the whole dialog. Our model is inspired by BLIP, a multi-modal architecture for image captioning task. During training, we adopt weighted loss based on multiple predictions. We penalize less for the early predictions due to the partial information.

\subsection{Training Details}
In training, both top-down map and ground-truth target are resized to $224\times 224$.  Color jittering, random cropping with ratio $[0,9, 1.0]$ and scale $[0,75, 1.0]$, and random rotation of $180^{\circ}$ are used for data augmentation. ViT-base pretrained on ImageNet-21k and Bert-base-uncased are used as unimodal encoders. We keep BERT encoder frozen in training. We train all models up to 30 epochs, with batch size 16. We use AdamW~\cite{loshchilov2017decoupled} as the optimizer and set the learning rate to $2e-5$. 
% Following official implementation\footnote{\url{https://github.com/meera1hahn/Graph_LED/tree/main/src/lingunet}} of LingUNet, the best checkpoint for the model is selected based on Acc0 on valUnseen set.
\section{Experiments}

First, we conduct several ablation studies aimed at validating the design choices, including the depth of multimodal encoders, fusion variants,  and the effect of the decay factor. Additionally, we delve into the evaluation of the auxiliary loss to support its efficacy. From the perspective of data augmentation, we observe that incorporating synthetic dialogs generated by ChatGPT API \st{}{\footnote{\url{https://api.openai.com/v1/chat/completions}}} contributes to performance enhancement. 
Second, we proceed to compare our proposed methods with the baseline approaches in both single-shot and multi-shot modes. This include quantitative results as well as qualitative samples. Lastly, we provide a detailed examination of multi-shot performance, offering a fine-grained analysis of the results. 

\subsection{Ablations}
\paragraph{Depth of multimodal encoder.} 
We first study the effect of depth of the multimodal encoder in Table~\ref{tab:ablate_depth}. We ablate this with the implicit fusion variant as shown in Figure~\ref{fig:arch1}. Results show that depth=3 gives the best performance in unseen set when vision encoder is frozen. When ViT is fine-tuned, the performance using more blocks drops. The  best unseen Acc5 is 47.09 using depth=1. Overall, fine-tuning ViT improves the performance a lot because ViT needs to adapt to the top-down map visual input.  
%%%%%%%%%%%%%%%%%%%%%%%%%%%%%%%%%%%%%%%%%%%%%%%%%%%
\begin{table}[ht]
\centering
\begin{tabular}{c|c|c|c|c|c}
\hline
\multirow{2}{*}{ViT}& \multirow{2}{*}{Depth} & \multicolumn{2}{c|}{valSeen} & \multicolumn{2}{c}{valUnseen} \\
 &  & LE$\downarrow$ & Acc5$\uparrow$ &  LE$\downarrow$ &  Acc5$\uparrow$ \\
\hline
\multirow{3}{*}{Frozen}    & 1 & 9.29 & 47.18 & 10.54 & 33.27 \\
                           & 3 & 9.45 & 43.43 & 9.26  & 38.23 \\
                           & 6 & 9.41 & 47.50 & 9.54  & 37.72 \\
                           \hline
\multirow{3}{*}{Fine-tune} & 1 & \textbf{7.62} & 57.37 & \textbf{9.09}  & \textbf{47.09} \\
                           & 3 & 7.73 & \textbf{58.01} & 9.35  & 42.06 \\
                           & 6 & 8.44 & 55.12 & 10.02  & 39.61 \\
                        
\hline
\end{tabular}
\vspace{5pt}
\caption{\textbf{Ablation on depth of multimodal encoders.} DiaLoc-i is used in this ablation with decay $\alpha=0$.  }
\label{tab:ablate_depth}
\end{table}
%%%%%%%%%%%%%%%%%%%%%%%%%%%%%%%%%%%%%%%%%%%%%%%%%%%

\textbf{Fusion variants and decay factor.} We have proposed two variants to perform multimodal fusion in Section~\ref{sec:arch}. DiaLoc-i updates the visual hidden state using dialog input recursively, while DiaLoc-e fuses the original map embedding with the hidden state using multiply fusion operation. To investigate how the two fusion variants perform, we report the results in Table~\ref{tab:ablate_fusion_decay}. For both variants (depth=3), we fine-tune the ViT. The results show that implicit fusion performs better on valSeen set, while explicit fusion works better on valUnseen. Implicit fusion leverages shared hidden state and has higher risk of over-fitting. Using fixed map embedding and learnable hidden state, explicit fusion shows better generalization. We also find that using non-zero decay factor leads to lower accuracy due to the early convergence. 

%%%%%%%%%%%%%%%%%%%%%%%%%%%%%%%%%%%%%%%%%%%%%%%%%%%
\begin{table}[ht]
\centering
\begin{tabular}{c|c|c|c|c|c}
\hline
\multirow{2}{*}{Method}& \multirow{2}{*}{Decay} & \multicolumn{2}{c|}{valSeen} & \multicolumn{2}{c}{valUnseen} \\
 &  & LE$\downarrow$ & Acc5$\uparrow$ &  LE$\downarrow$ &  Acc5$\uparrow$ \\
\hline
\multirow{3}{*}{\makecell{DiaLoc-i}}    & 0.0 & \textbf{7.73} & \textbf{58.01} & 9.35 & \textbf{42.06} \\
                             & 0.5 & 7.95 & 54.68 & 9.74 & 36.48 \\
                             & 1.0 & 8.20 & 55.31 & 9.75 & 33.61 \\
                             \hline
\multirow{3}{*}{\makecell{DiaLoc-e}}    & 0.0 & 8.99 & 50.32 & \textbf{9.10} & 41.60 \\
                             & 0.5 & 9.51 & 54.68 & 9.76 & 41.10 \\
                             & 1.0 & 8.91 & 51.56 & 9.21 & 37.95 \\
\hline
\end{tabular}
\vspace{5pt}
\caption{\textbf{Ablation on fusion schemes and the decay factor.} Both DiaLoc variants use depth=3 without auxiliary loss. } 
\label{tab:ablate_fusion_decay}
\end{table}
%%%%%%%%%%%%%%%%%%%%%%%%%%%%%%%%%%%%%%%%%%%%%%%%%%%

%\multirow{2}{*}{reuse} & \multirow{2}{*}{mse0} & 0.5 & 6.76 & 63.46 & 9.58 & 40.01 \\
% & & 1.0 & 7.92 & 58.65 & 8.99 & 42.06 \\

\textbf{Impact of auxiliary loss.} We ablate the impact of using auxiliary loss as defined in Eq~\ref{eq:aux}. We train the model with depth as 1 with varying decay factor of KL-divergence term. For each decay choice, the same model is trained with ($\beta$=1) and without ($\beta$=0) using auxiliary loss. Results are shown in Table~\ref{tab:ablate_auxloss}, combining the two losses helps to improve the performance consistently across all setups. 

%%%%%%%%%%%%%%%% ablation on Decay x Aux %%%%%%%%%%%%%%%%%%%%%
\begin{table}[ht]
%\captionsetup{font=scriptsize}
\small
\centering
\scalebox{0.75}{
\begin{tabular}{c|c|c|c|c|c|c|c|c|c}
\hline
\multirow{3}{*}{\makecell{Decay\\$\alpha$}} & \multirow{3}{*}{\makecell{Aux\\$\beta$}} & \multicolumn{4}{c|}{\color{blue}DiaLoc-e} & \multicolumn{4}{c}{\color{red}DiaLoc-i} \\
& & \multicolumn{2}{c|}{valSeen} & \multicolumn{2}{c|}{valUnseen} & \multicolumn{2}{c|}{valSeen} & \multicolumn{2}{c}{valUnseen} \\
 &  & LE$\downarrow$ & Acc5$\uparrow$ &  LE$\downarrow$ &  Acc5$\uparrow$ & LE$\downarrow$ & Acc5$\uparrow$ &  LE$\downarrow$ &  Acc5$\uparrow$ \\
\hline
\multirow{2}{*}{0.0}     & 0.0 & 7.81 & 57.18 & 9.99 & \color{blue}37.89 & \color{red}7.54 & \color{red}58.65 & \color{red}9.61 & 34.19 \\
                         & 1.0 & 7.76 & 58.01 & 9.56 & \color{blue}40.23 & \color{red}7.39 & \color{red}60.57 & \color{red}9.19 & 38.58\\
                         \hline
\multirow{2}{*}{0.5}     & 0.0 & 7.95 & 54.68 & \color{blue}9.74 & \color{blue}36.48 & \color{red}7.16 & \color{red}58.97 & 9.98 & 36.41\\
                         & 1.0 & \color{blue}6.76 & \color{blue}63.46 & \color{blue}9.58 & \color{blue}40.01 & 7.79 & 60.25 & 10.54 & 32.53 \\
                         \hline
\multirow{2}{*}{1.0}     & 0.0 & 8.20 & 55.31 & \color{blue}9.75 & 33.61 & \color{red}7.99 & \color{red}56.41 & 11.01 & \color{red}34.36\\
                         & 1.0 & 7.92 & 58.65 & \color{blue}8.99 & \color{blue}42.06 & \color{red}7.55 & \color{red}59.29 & 10.42 & 36.47 \\
    
\hline
\end{tabular}}
% \vspace{-10pt}
\caption{\textbf{Ablation on auxiliary loss using explicit fusion variant DiaLoc-e.} We vary the decay factor $\alpha$ to train multiple versions of the model,  with and without the proposed auxiliary loss. DiaLoc-e generalizes (9/12 valUnseen) better than DiaLoc-i (10/12 valSeen)}
% \vspace{-5pt}
\label{tab:ablate_auxloss}
\end{table}
%%%%%%%%%%%%%%%%%%%%%%%%%%%%%%%%%%%%%%%%%%%%%%%%%%%
% %%%%%%%%%%%%%%%%%%%%%%%%%%%%%%%%%%%%%%%%%%%%%%%%%%%
% \begin{table}[ht]
% \centering
% \begin{tabular}{c|c|c|c|c|c}
% \hline
% \multirow{2}{*}{\makecell{Decay\\$\alpha$}}& \multirow{2}{*}{\makecell{Aux\\$\beta$}} & \multicolumn{2}{c|}{valSeen} & \multicolumn{2}{c}{valUnseen} \\
%  &  & LE$\downarrow$ & Acc5$\uparrow$ &  LE$\downarrow$ &  Acc5$\uparrow$ \\
% \hline
% \multirow{2}{*}{0.0}     & 0.0 & 7.81 & 57.18 & 9.99 & 37.89 \\
%                          & 1.0 & 7.76 & 58.01 & 9.56 & 40.23 \\
%                          \hline
% \multirow{2}{*}{0.5}     & 0.0 & 7.95 & 54.68 & 9.74 & 36.48 \\
%                          & 1.0 & \textbf{6.76} & \textbf{63.46} & 9.58 & 40.01 \\
%                          \hline
% \multirow{2}{*}{1.0}     & 0.0 & 8.20 & 55.31 & 9.75 & 33.61 \\
%                          & 1.0 & 7.92 & 58.65 & \textbf{8.99} & \textbf{42.06} \\
    
% \hline
% \end{tabular}
% \vspace{5pt}
% \caption{\textbf{Ablation on auxiliary loss using explicit fusion variant DiaLoc-e.} We vary the decay factor $\alpha$ to train multiple versions of the model,  with and without the proposed auxiliary loss. }
% \label{tab:ablate_auxloss}
% \end{table}
% %%%%%%%%%%%%%%%%%%%%%%%%%%%%%%%%%%%%%%%%%%%%%%%%%%%

\textbf{Dialog augmentation using LLM.} We hypothesize that increasing the dialog diversity during training could improve the performance because text encoder is frozen in our work. We leverage GPT to study the effect of dialog augmentation on the task. For each training sample, we prompt GPT API to paraphrase the ground-truth dialog. During training, either the GPT version or the ground-truth dialog is chosen randomly. Results in Table~\ref{tab:ablate_gpt} show that using extra data helps improve the performance for seen and unseen maps. 
%%%%%%%%%%%%%%%%%%%%%%%%%%%%%%%%%%%%%%%%%%%%%%%%%%%
\begin{table}[ht]
\centering
\begin{tabular}{c|c|c|c|c|c}
\hline
\multirow{2}{*}{Method} & \multirow{2}{*}{Dialog} & \multicolumn{2}{c|}{valSeen} & \multicolumn{2}{c}{valUnseen} \\
&  & LE$\downarrow$ & Acc5$\uparrow$ &  LE$\downarrow$ &  Acc5$\uparrow$ \\
\hline
%multiply,d3 & 7.80 & 57.81 & 8.94 & 39.24 \\
\multirow{2}{*}{DiaLoc-e} & GT     & 8.91 & 51.56 & 9.21 & 37.95 \\
                        & GT+GPT & \textbf{7.07} & \textbf{60.00} & \textbf{9.09} & \textbf{40.71} \\
\hline
\end{tabular}
\vspace{5pt}
\caption{\textbf{Ablation on using extra dialogs generated by GPT.} DiaLoc-e is configured to use depth=3, $\alpha=1$, and $\beta=0$}
\label{tab:ablate_gpt}
\end{table}
\begin{figure*}[t]
    \centering
    \setlength{\tabcolsep}{1pt}
    \renewcommand{\arraystretch}{0.4}
    \begin{tabular}{p{10pt} c c c c c}
         % (a) & (b) & (c) & (d) & (e) \\
        % seen, 67
       \rotatebox{90}{LingUNet} &
        \begin{tikzpicture}
            \draw (0, 0) node[inner sep=0]
            {\includegraphics[width=0.2\linewidth]{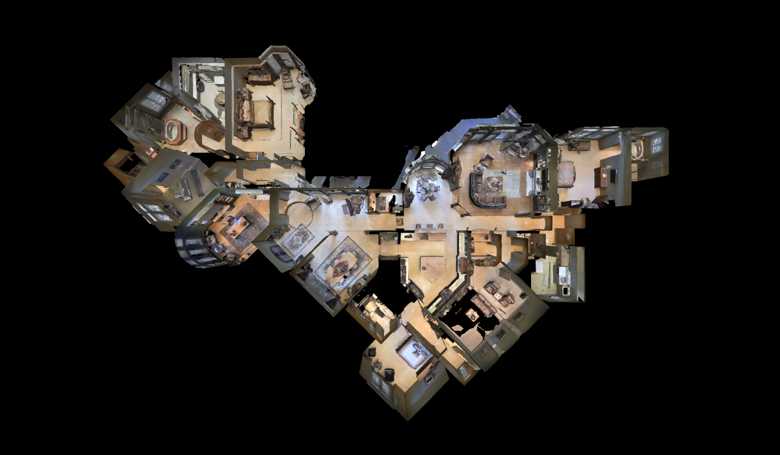}};
            \draw (1.4, 0.8) node[color=white,font=\small] {Map};
        \end{tikzpicture} &
        \begin{tikzpicture}
            \draw (0, 0) node[inner sep=0]
            {\includegraphics[width=0.2\linewidth]{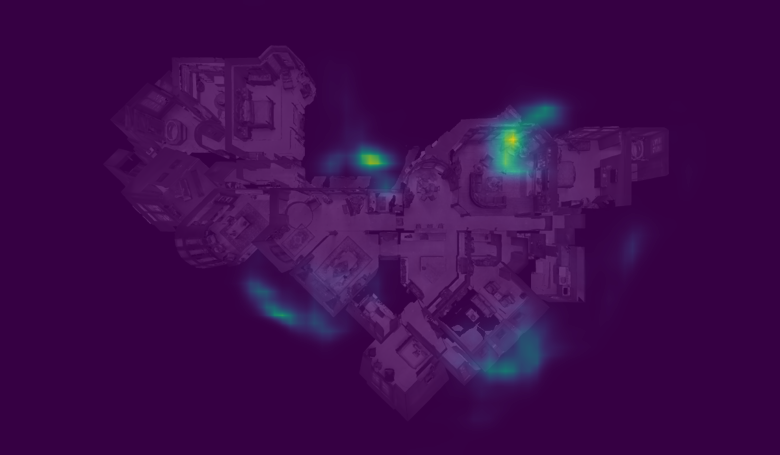}};
            \draw (1.2, 0.8) node[color=red,font=\small] {1.63m};
        \end{tikzpicture} &
        \begin{tikzpicture}
            \draw (0, 0) node[inner sep=0]
            {\includegraphics[width=0.2\linewidth]{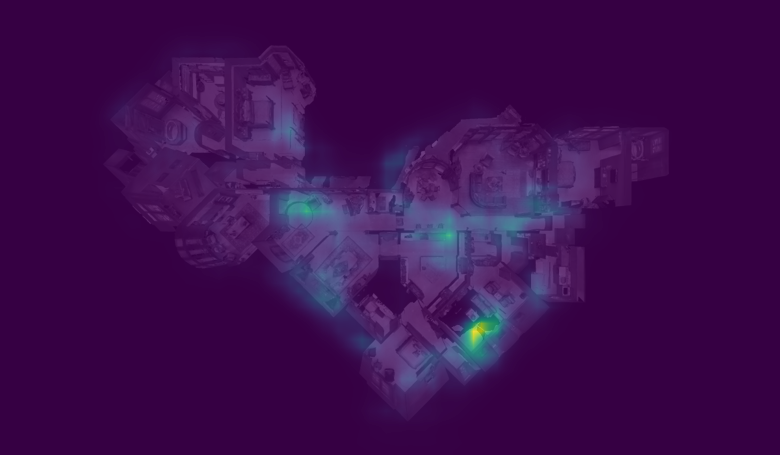}};
            \draw (1.2, 0.8) node[color=cyan,font=\small] {14.42m};
        \end{tikzpicture} &
        \begin{tikzpicture}
            \draw (0, 0) node[inner sep=0]
            {\includegraphics[width=0.2\linewidth]{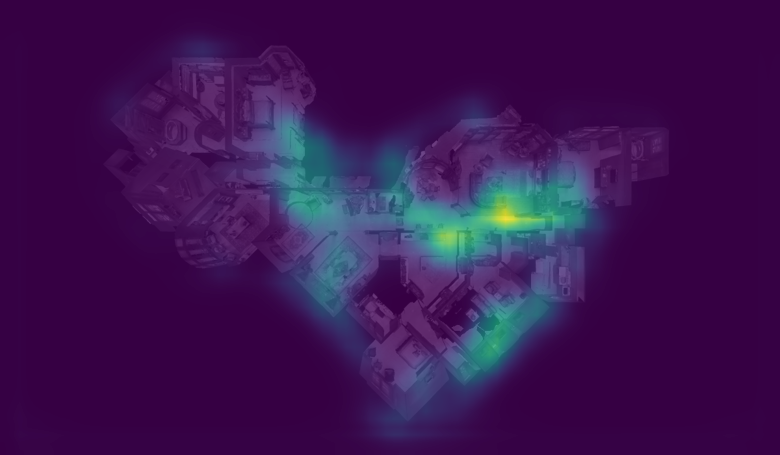}};
            \draw (1.2, 0.8) node[color=cyan,font=\small] {5.19m};
        \end{tikzpicture} &
        \begin{tikzpicture}
            \draw (0, 0) node[inner sep=0]
            {\includegraphics[width=0.2\linewidth]{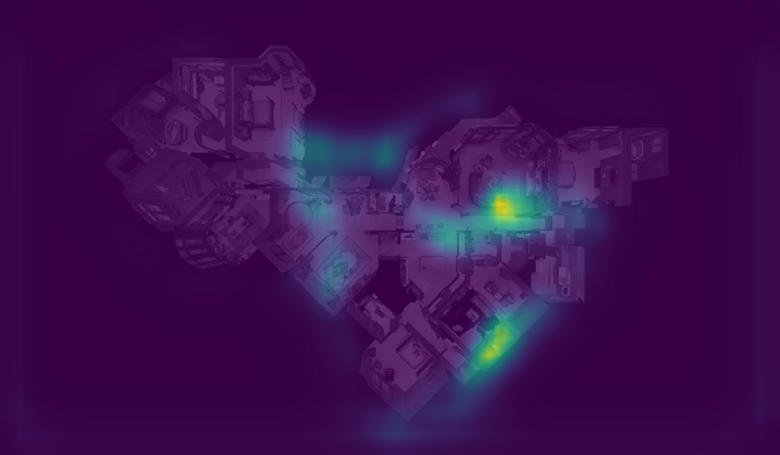}};
            \draw (1.2, 0.8) node[color=cyan,font=\small] {6.01m};
        \end{tikzpicture} 
         \\
        \rotatebox{90}{DiaLoc} &
        \begin{tikzpicture}
            \draw (0, 0) node[inner sep=0]
            {\includegraphics[width=0.2\linewidth]{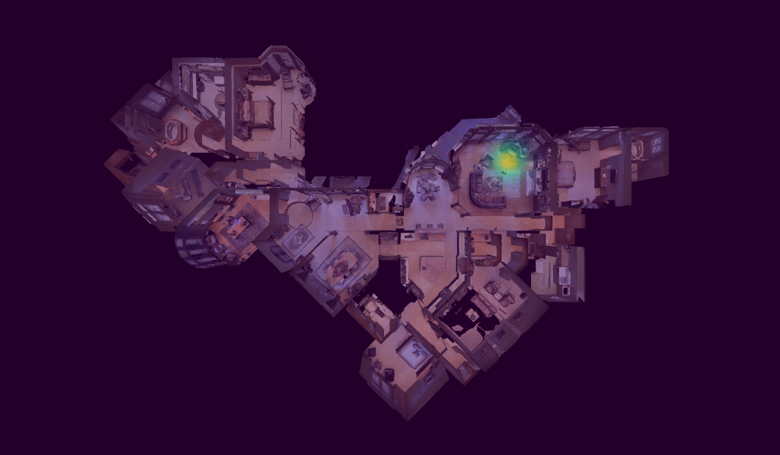}};
            \draw (1.4, 0.8) node[color=white,font=\small] {GT};
         \end{tikzpicture}  
         &
         \begin{tikzpicture}
            \draw (0, 0) node[inner sep=0]
            {\includegraphics[width=0.2\linewidth]{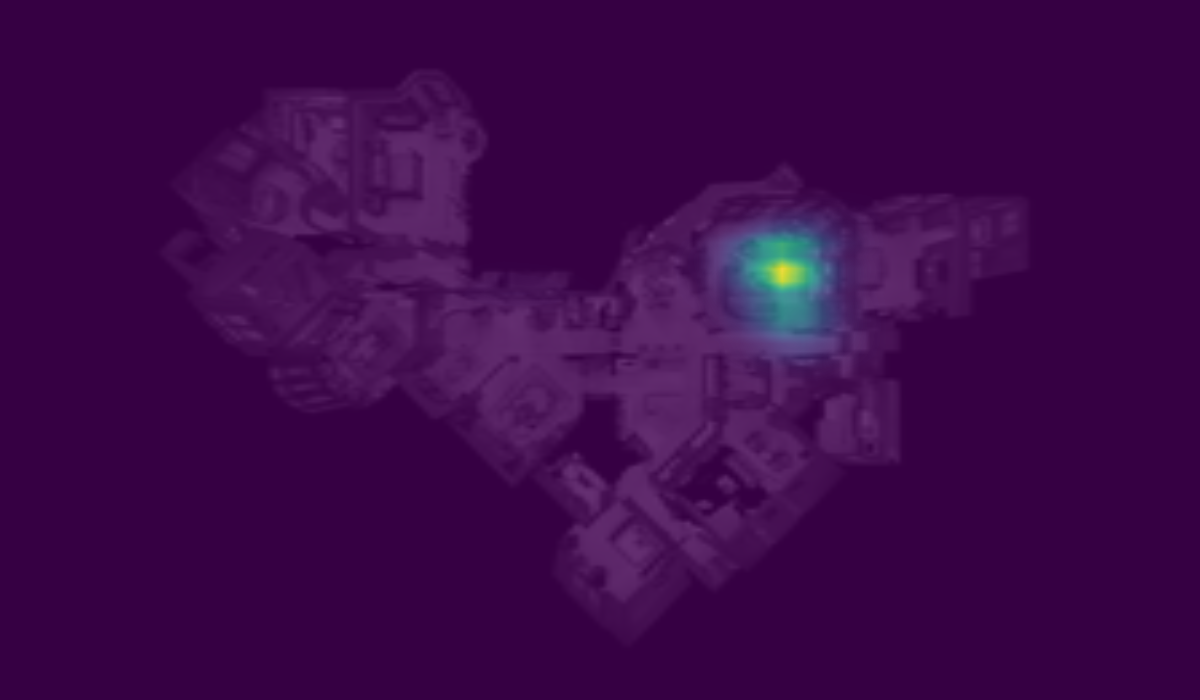}};
            \draw (1.2, 0.8) node[color=red,font=\small] {1.88m};
         \end{tikzpicture} &
         \begin{tikzpicture}
            \draw (0, 0) node[inner sep=0]
            {\includegraphics[width=0.2\linewidth]{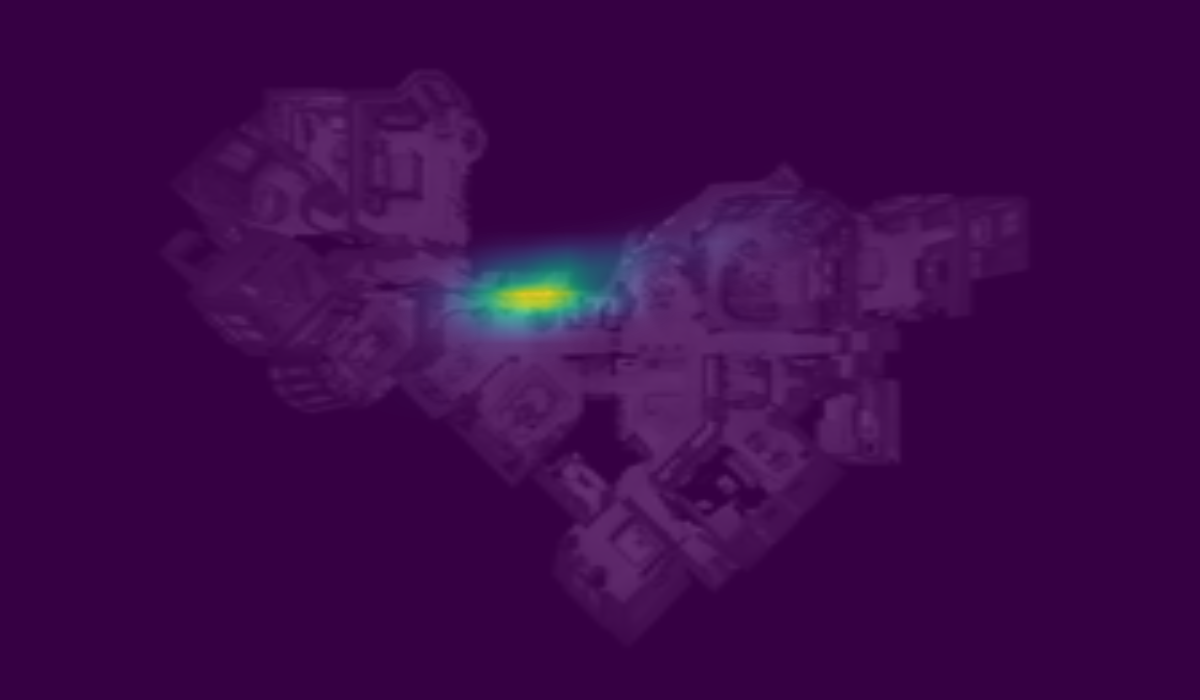}};
            \draw (1.2, 0.8) node[color=cyan,font=\small] {16.58m};
         \end{tikzpicture} &
         \begin{tikzpicture}
            \draw (0, 0) node[inner sep=0]
            {\includegraphics[width=0.2\linewidth]{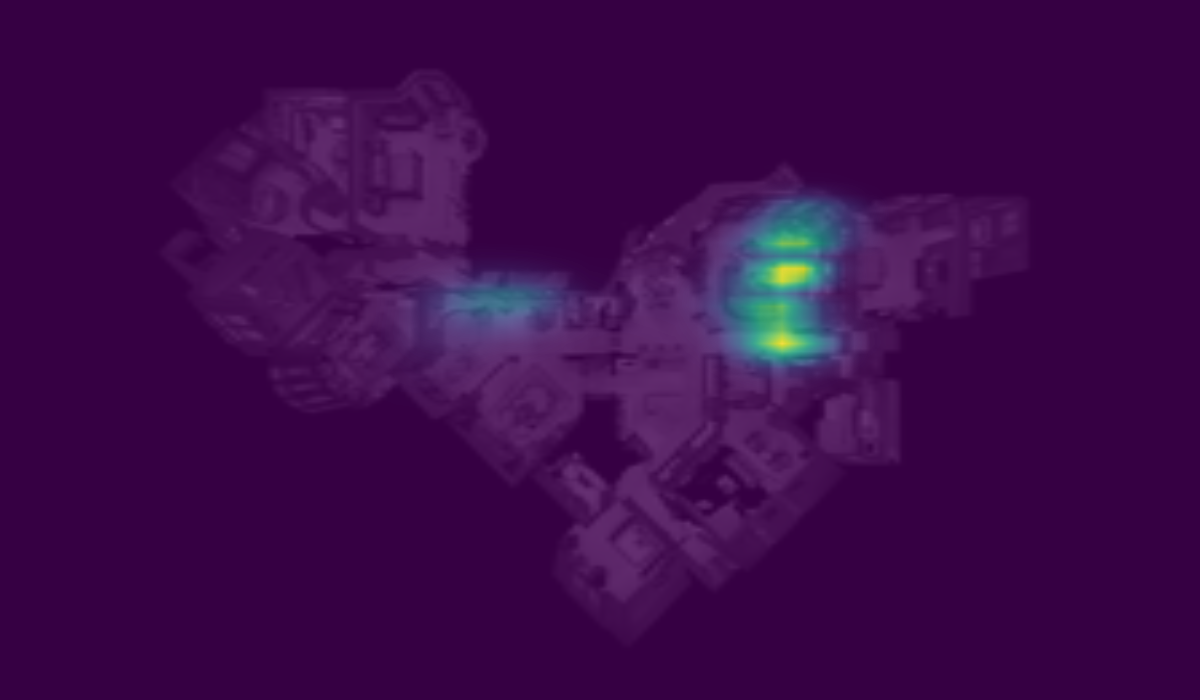}};
            \draw (1.2, 0.8) node[color=cyan,font=\small] {5.19m};
         \end{tikzpicture} &
         \begin{tikzpicture}
            \draw (0, 0) node[inner sep=0]
            {\includegraphics[width=0.2\linewidth]{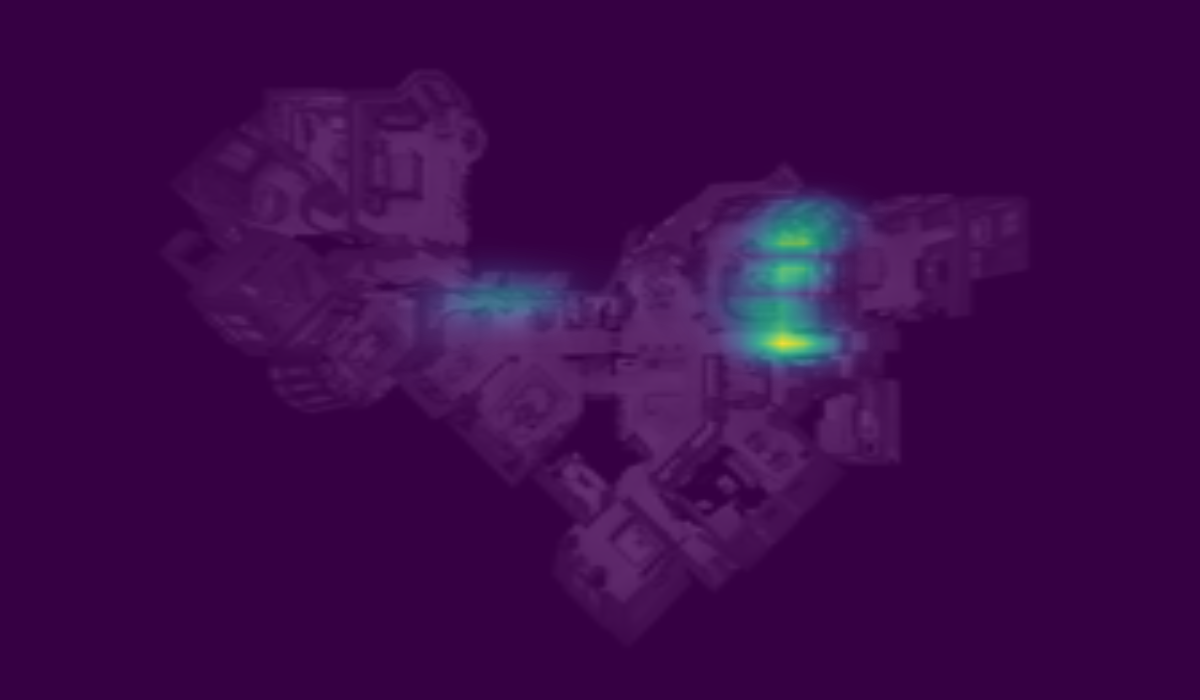}};
            \draw (1.2, 0.8) node[color=cyan,font=\small] {4.66m};
         \end{tikzpicture}  
         \\
         & val-seen 67 & {\color{red}Single-shot} & {\color{cyan}Multi-shot:} $1/T$ & $2/T$ & $3/T$ \\
         % unseen, 245
         \rotatebox{90}{LingUNet} &
         \begin{tikzpicture}
            \draw (0, 0) node[inner sep=0] 
            {\includegraphics[width=0.2\linewidth]{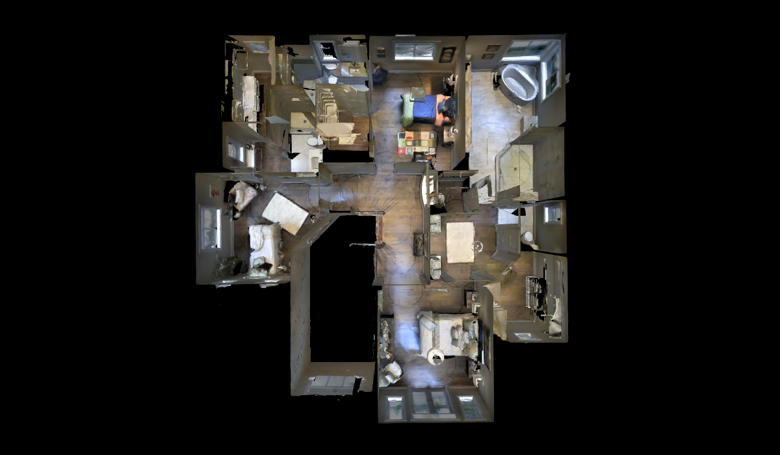}};
            \draw (1.2, 0.8) node[color=white,font=\small] {Map};
         \end{tikzpicture} &
         \begin{tikzpicture}
            \draw (0, 0) node[inner sep=0] 
            {\includegraphics[width=0.2\linewidth]{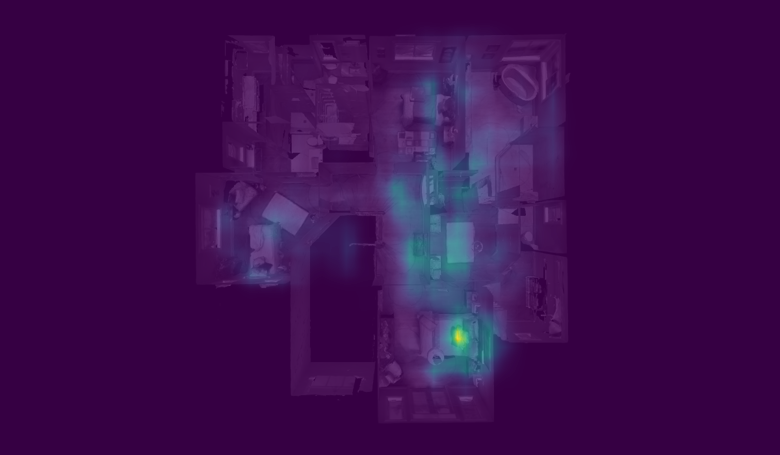}};
            \draw (1.2, 0.8) node[color=red,font=\small] {9.72m};
         \end{tikzpicture} &
         \begin{tikzpicture}
            \draw (0, 0) node[inner sep=0] 
            {\includegraphics[width=0.2\linewidth]{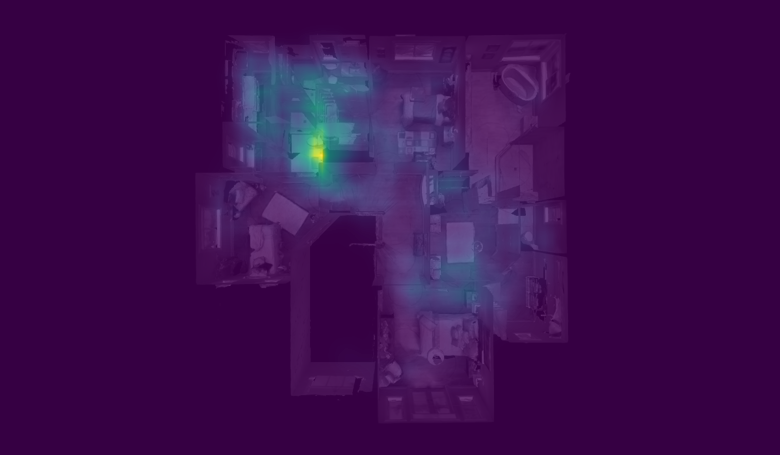}};
            \draw (1.2, 0.8) node[color=cyan,font=\small] {2.08m};
         \end{tikzpicture} &
         \begin{tikzpicture}
            \draw (0, 0) node[inner sep=0] 
            {\includegraphics[width=0.2\linewidth]{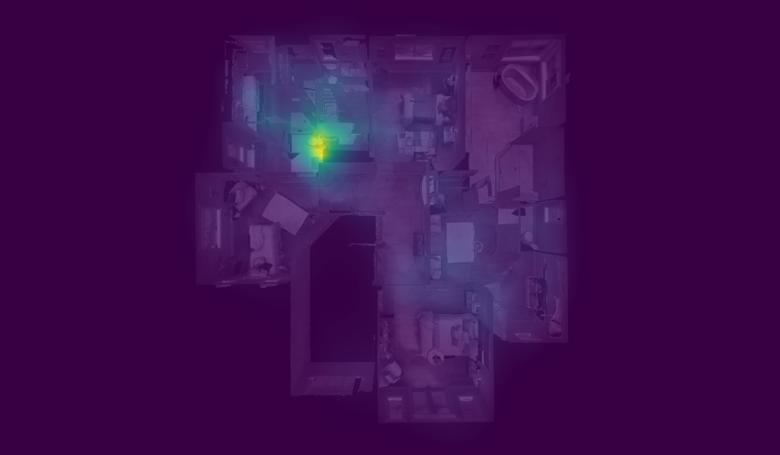}};
            \draw (1.2, 0.8) node[color=cyan,font=\small] {2.08m};
         \end{tikzpicture} &
         \begin{tikzpicture}
            \draw (0, 0) node[inner sep=0] 
            {\includegraphics[width=0.2\linewidth]{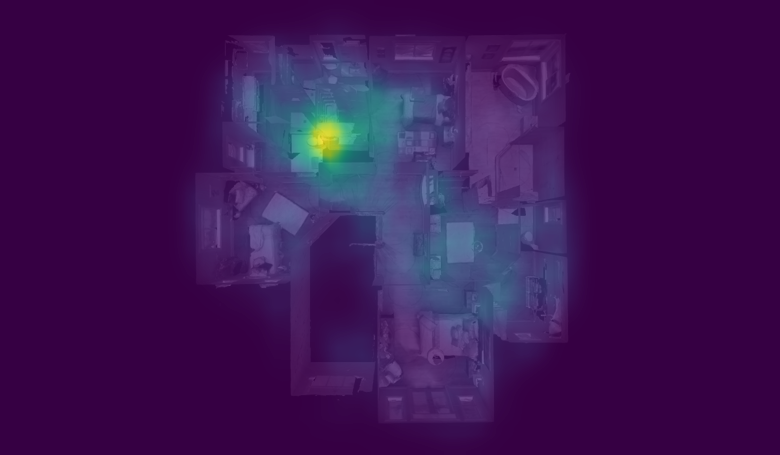}};
            \draw (1.2, 0.8) node[color=cyan,font=\small] {3.74m};
         \end{tikzpicture}
         \\
         \rotatebox{90}{DiaLoc-e} & 
         \begin{tikzpicture}
            \draw (0, 0) node[inner sep=0] 
            {\includegraphics[width=0.2\linewidth]{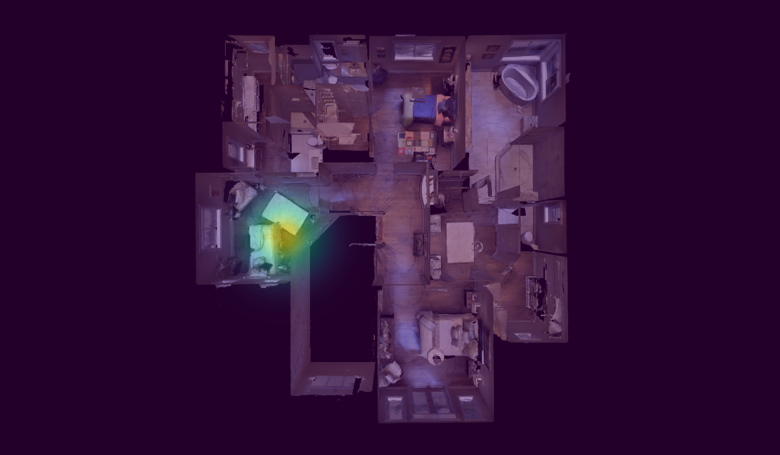}};
            \draw (1.4, 0.8) node[color=white,font=\small] {GT};
         \end{tikzpicture} 
         &
         \begin{tikzpicture}
            \draw (0, 0) node[inner sep=0] 
            {\includegraphics[width=0.2\linewidth]{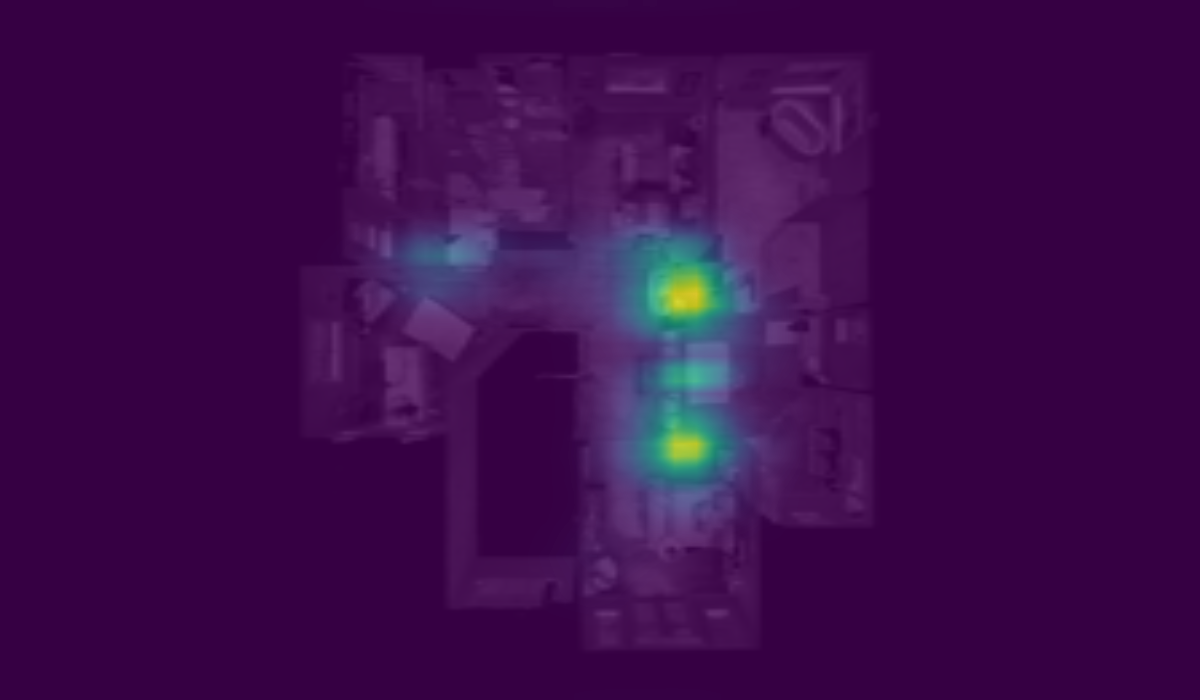}};
            \draw (1.2, 0.8) node[color=red,font=\small] {4.32m};
         \end{tikzpicture} 
         &
         \begin{tikzpicture}
            \draw (0, 0) node[inner sep=0] 
            {\includegraphics[width=0.2\linewidth]{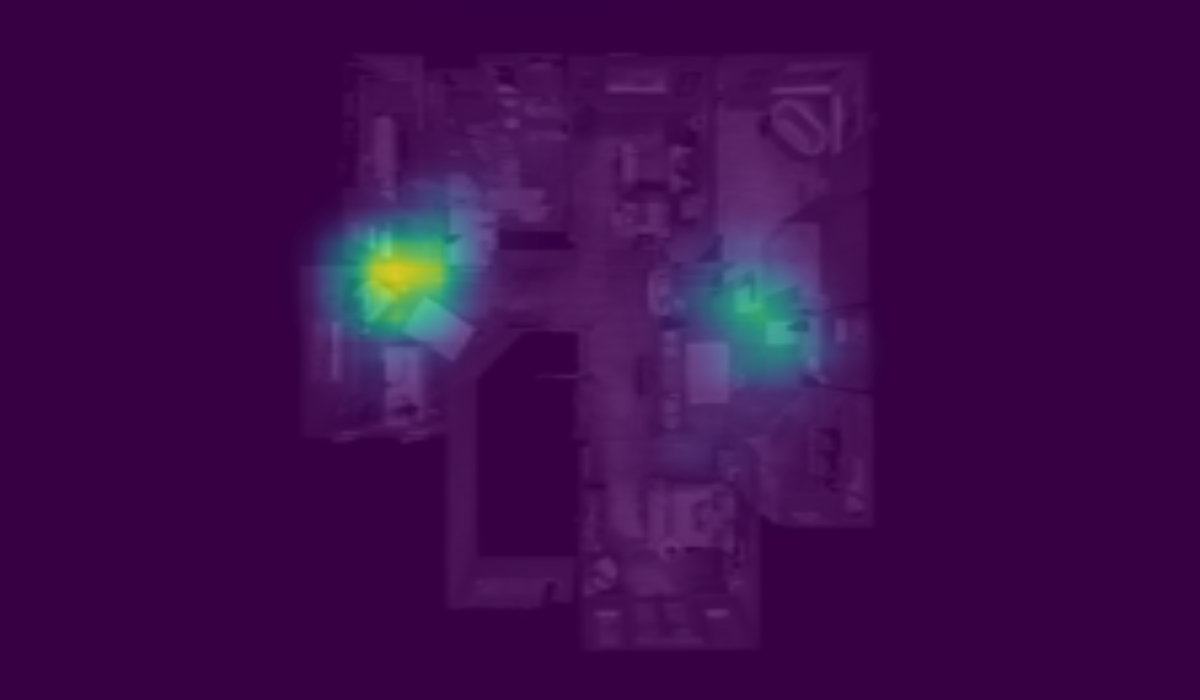}};
            \draw (1.2, 0.8) node[color=cyan,font=\small] {2.64m};
         \end{tikzpicture} 
         &
         \begin{tikzpicture}
            \draw (0, 0) node[inner sep=0] 
            {\includegraphics[width=0.2\linewidth]{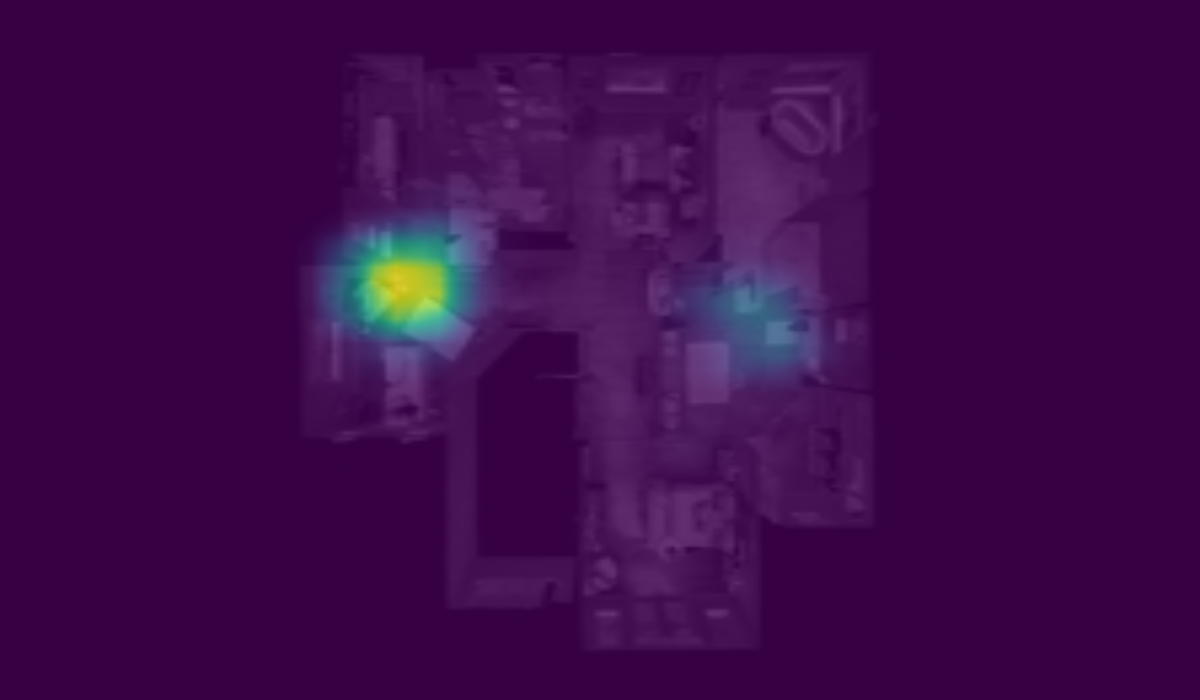}};
            \draw (1.2, 0.8) node[color=cyan,font=\small] {1.56m};
         \end{tikzpicture} 
         &
         \begin{tikzpicture}
            \draw (0, 0) node[inner sep=0] 
            {\includegraphics[width=0.2\linewidth]{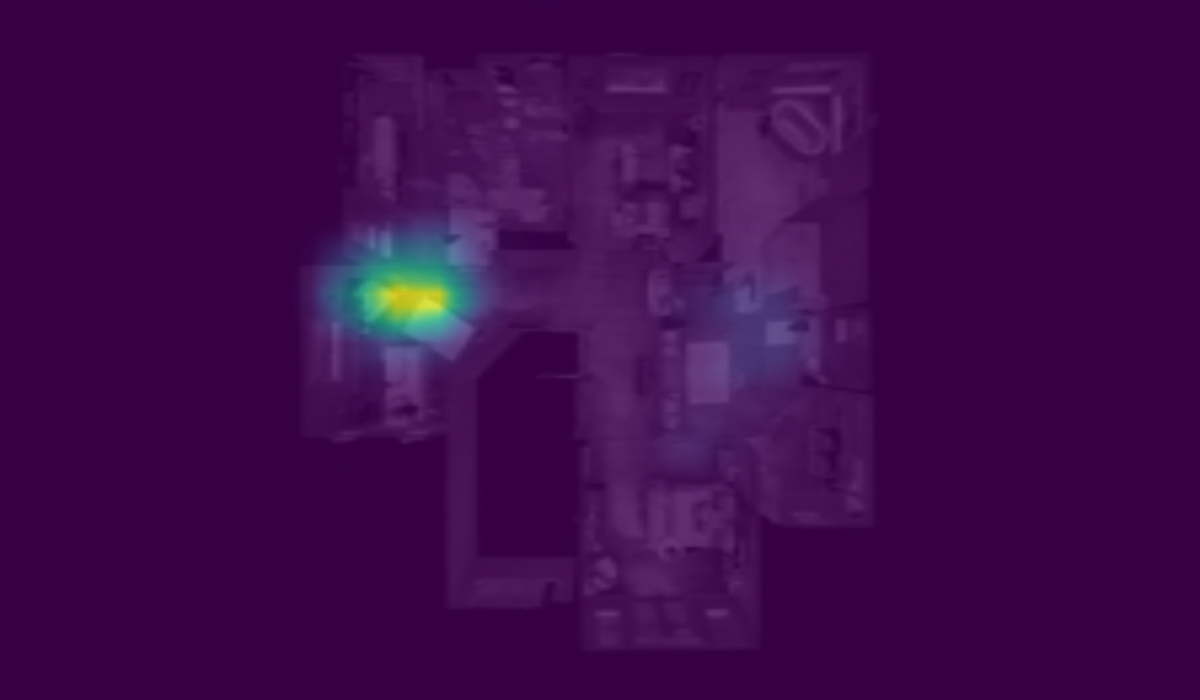}};
            \draw (1.2, 0.8) node[color=cyan,font=\small] {1.56m};
         \end{tikzpicture}
         \\
         & val-unseen 245 & {\color{red}Single-shot} & {\color{cyan}Multi-shot:} $1/T$ & $2/T$ & $3/T$ \\ \\
        
    \end{tabular}
    \caption{\textbf{Qualitative results of single-shot and multi-shot location predictions are presented.} In the first column, the visual map is displayed alongside its corresponding ground truth (GT) location. The second column displays the single-shot predictions, with LingUNet results above and DiaLoc results below. The last three columns showcase the multi-shot predictions. Regarding the results of valseen case 67, DiaLoc effectively corrects the prediction after the second turn, whereas LingUNet-ms produces noisy distributions.  For valUnseen case 245, both LingUNet and DiaLoc failed in the single-shot mode. Nevertheless, in the multi-shot mode, DiaLoc succeeds in refining the prediction. In contrast, LingUNet-ms converges towards an incorrect area. Localization Error (LE) is displayed for the predictions.}
    \vspace{-10pt}
    \label{fig:quali}
\end{figure*}
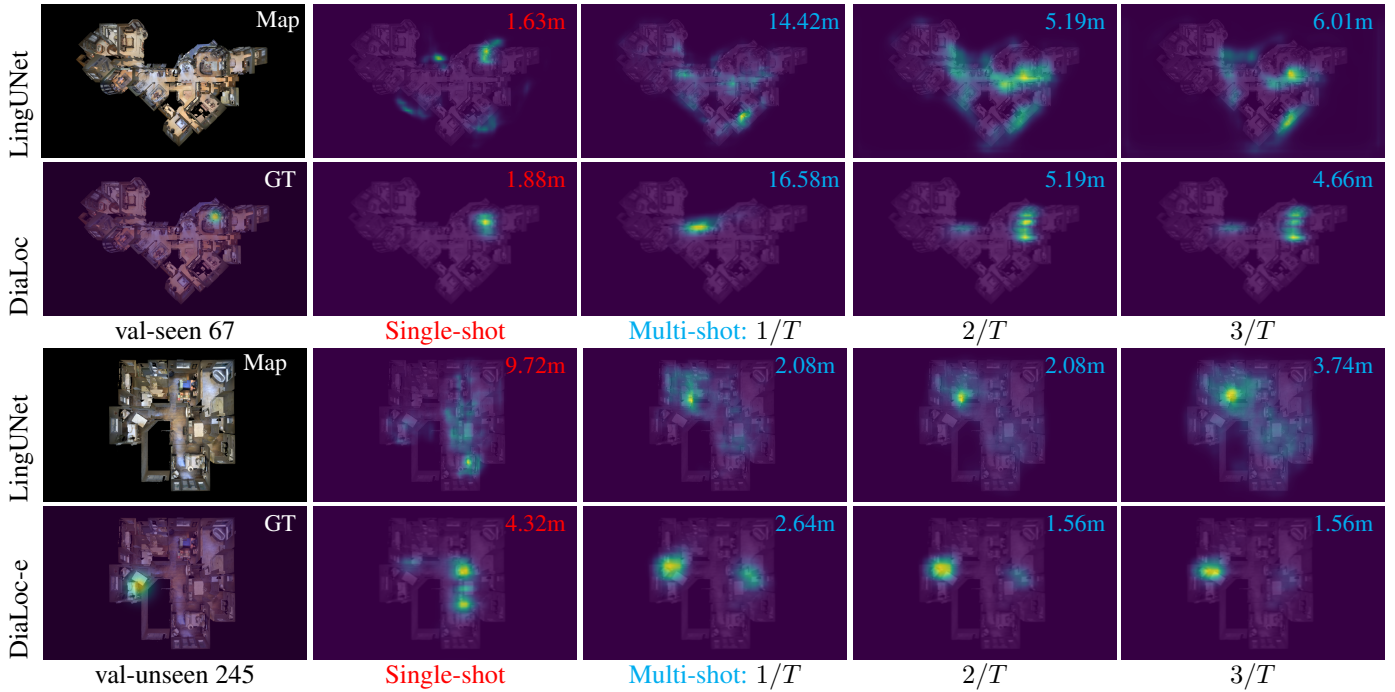
%%%%%%%%%%%%%%%%%%%%%%%%%%%%%%%%%%%%%%

\subsection{Comparison to the SOTA}

\paragraph{Single-shot and multi-shot.}
In this section, we undertake a comparative analysis of our method with state-of-the-art approaches in the LED task. 
For single-shot mode, LingUNet~\cite{hahn2020you} using complete dialog stands as the current SOTA method. We report the geodesic distance performance using the author-provided code. 
We also adapt BLIP~\cite{li2022blip} as BLIP-Res18 and BLIP(DiaLoc) to use dialogs as conditional input for image-based localization task. 
%In this context, both of our fusion variants become equivalent in this scenario and output just one prediction.  

In the multi-shot mode, we adapt LingUNet similarly to our DiaLoc-i and DiaLoc-e, enabling it to generate multi-shot predictions \st{TODO: more detail needed}{Specifically, the fusion is implemented by multiplying the output (H1 preceding the final MLP) with image features (F1). More details in Supp.Mat}. We refer to this method as LingUNet-i/e. Our proposed method is denoted as DiaLoc-i and DiaLoc-e, reflecting the different fusion variants \st{}{implicit and explicit respectively}. In alignment with \cite{hahn2020you}, we report Acc0 and Acc5 as evaluation metrics for both valSeen and valUnseen sets in Table~\ref{tab:sota}.

%Lastly, we show that our method benefits from pseudo GT produced by models trained on whole dialogs. 
%%%%%%%%%%%%%%%%%% SOTA: add euclidean, LUNet-msi %%%%%%%%%%%%%%%%%%%
\begin{table}[t]
%\captionsetup{font=scriptsize}
    \centering
    \small
    \begin{tabular}{c|c|c|c|c|c}
    \hline
    \multirow{2}{*}{Mode} & \multirow{2}{*}{Method} & \multicolumn{2}{c|}{valSeen} & \multicolumn{2}{c}{valUnseen} \\
      & & Acc0 & Acc5  & Acc0 & Acc5  \\
     \hline
     \multirow{3}{*}{\makecell{Single\\shot}}
         & LingUNet  & 19.87 & 59.29  & 6.16 & 33.33 \\
         & BLIP-Res18{$\diamond$} & 13.44 & 50.82  & 6.73 & 34.71 \\
         & BLIP(DiaLoc){$\dagger$} & \textbf{25.64} & \textbf{66.02}  & 7.02 & 40.41 \\
         
     \hline 
     \multirow{4}{*}{\makecell{Multi \\ shot}} 
      & LingUNet-i & 4.32 & 24.16  & 3.25 & 20.21  \\
      & LingUNet-e & 14.47 & 46.15 & 5.31 & 36.30  \\
      & DiaLoc-i   & 18.43 & 57.18 & 6.42 & 37.89  \\
      & DiaLoc-e   & 18.36 & 60.00 & \textbf{8.44} & \textbf{47.15} \\
      % TransMean: average heatmaps
     \hline
     \end{tabular}
    % \vspace{-10pt}
    \caption{\textbf{Evaluations on WAY dataset under single-shot and multi-shot scenarios.} We configure DiaLoc-i and DiaLoc-e to use $d=3, \alpha=0, \beta=1$. BLIP-Res18{$\diamond$} and BLIP(DiaLoc){$\dagger$} are adapated so that the visual branch is configured as the main backbone to predict heatmap.}
    % \vspace{-15pt}
    \label{tab:sota}
\end{table}
%%%%%%%%%%%%%%%%%%%%%%%%%%%%%%%%%%%%%%%%%%%%%%%%%%%

% %%%%%%%%%%%%%%%%%%%%%%%%%%%%%%%%%%%%%%%%%%%%%%%%%%%
% \begin{table}[ht]
%     \centering
%     \small
%     \begin{tabular}{c|c|c|c|c|c}
%     \hline
%     \multirow{2}{*}{Modes} & \multirow{2}{*}{Method} & \multicolumn{2}{c|}{valSeen} & \multicolumn{2}{c}{valUnseen} \\
%       &  & Acc0$\uparrow$ & Acc5$\uparrow$ & Acc0$\uparrow$ & Acc5$\uparrow$ \\
%      \hline
%      \multirow{3}{*}{\makecell{Single\\shot}} & \color{gray}LingUNet{$\diamond$}  & - & \color{gray}67.2 & - & \color{gray}63.6 \\
%                                                & LingUNet{$\dagger$}  & 19.87 & 59.29 & 6.16 & 33.33 \\
%                                                 & BLIP & \textbf{25.64} & \textbf{66.02} & 7.02 & 40.41 \\
%      \hline 
%      \multirow{3}{*}{\makecell{Multi \\ shot}} & LingUNet-ms    & 14.47 & 46.15 & 5.31 & 36.30 \\
%       & DiaLoc-i   & 18.43 & 57.18 & 6.42 & 37.89 \\
%       & DiaLoc-e   & 18.36 & 60.00 & \textbf{8.44} & \textbf{47.15} \\
%       % TransMean: average heatmaps
%      \hline
%     \end{tabular}
%     \vspace{5pt}
%     \caption{\textbf{Evaluations on WAY dataset under single-shot and multi-shot scenarios.} We configure DiaLoc-i and DiaLoc-e to use $d=3, \alpha=0, \beta=1$. LingUNet{$\diamond$} are based on Euclidean distance taken from \cite{hahn2020you} for reference only. LingUNet{$\dagger$} are reproduced when training on target floor only.}
%     \label{tab:sota}
% \end{table}
% %%%%%%%%%%%%%%%%%%%%%%%%%%%%%%%%%%%%%%%%%%%%%%%%%%%

%%%%%%%%%%%%%%%%%%%%%%%%%%%%%%%%%%%%%%%%%%%%%%%%%%%
\begin{figure}[t]
    \centering
    \includegraphics[trim={5 5 5 10},clip,width=0.98\linewidth]{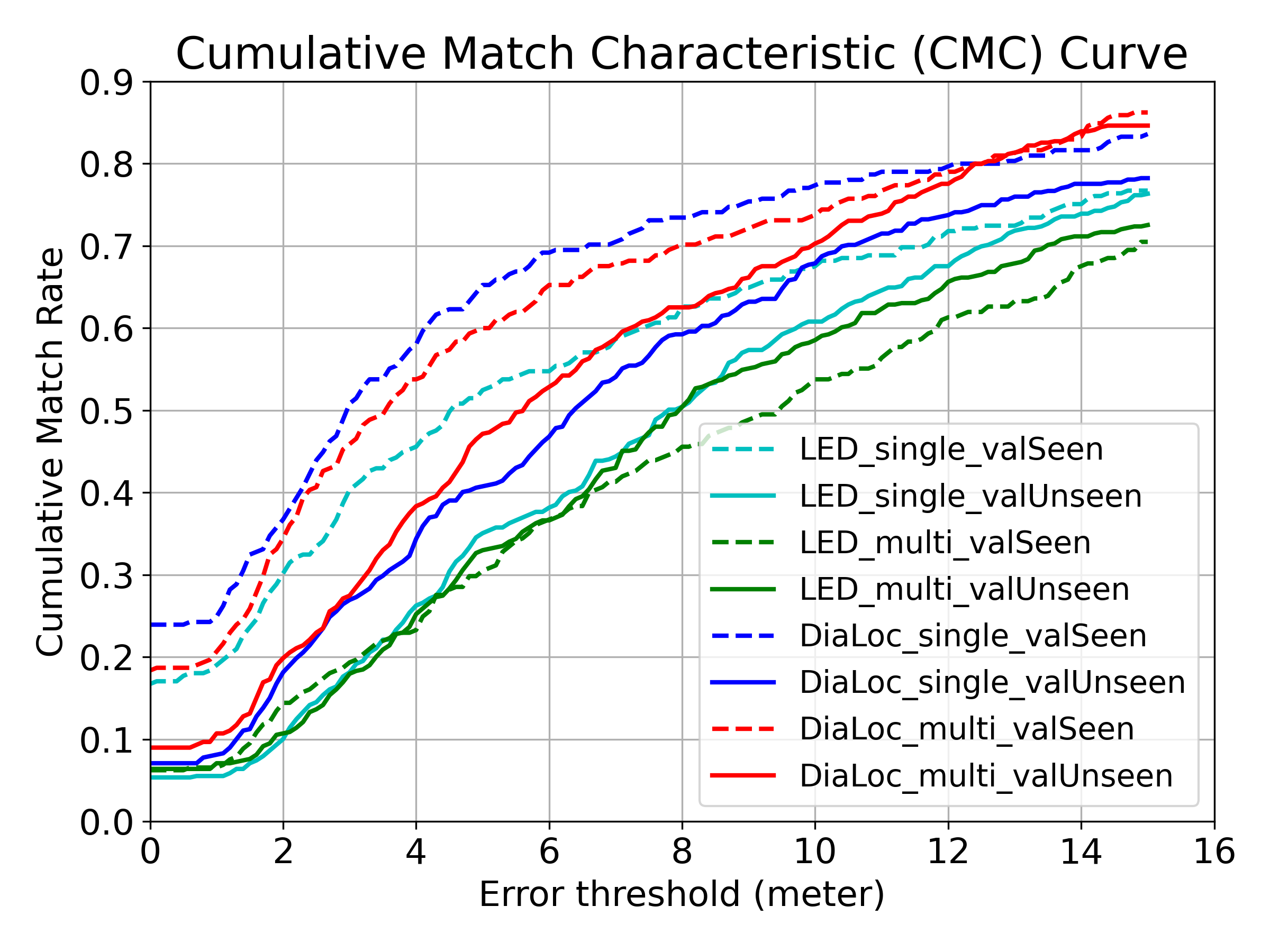}
    \caption{\textbf{CMC curves on WAY dataset.} We depict the CMC curves for both our DiaLoc and LingUNet for single-shot and multi-shot settings. DiaLoc consistently outperforms the baseline. X-axis denotes the error threshold for LE and the Y-axis denotes the success rate. }
    \label{fig:cmc}
\end{figure}
%%%%%%%%%%%%%%%%%%%%%%%%%%%%%%%%%%%%%%%%%%%%%%%%%%%
When utilizing the entire dialog both during training and inference, our method outperforms LingUNet by a substantial margin. The improvement in Acc5 on the valUnseen is 7.08. This performance enhancement can be attributed to the learning capabilities and flexibility of Transformers for both unimodal and multimodal encoders. 

In the context of the proposed multi-shot localization task, our evaluation focuses on final prediction accuracy. First, in comparison to the mode using complete dialogs, the performance of all methods exhibits a decline in the valSeen set. Our methods demonstrate competitive performance in valSeen set. Notably, DiaLoc-e surpasses all other methods in the valUnseen set, demonstrating robust generalization and reasoning capabilities. It is also noteworthy that LingUNet-ms performs well in the valUnseen set, even though its performance drops in valSeen. This observation suggests that the proposed iterative multi-shot approach effectively reduces the issue of over-fitting.

% computation: O(n^2) to O(n)
Considering computational complexity, there are significant savings when switching from the one-shot to the multi-shot scenario, despite multi-shot demanding multiple forward passes. Now, we compare the computational costs during the dialog embedding stage with Bert. For a dialog with $T$ turns, where each turn is encoded by $N$ tokens, the single-shot mode involves $O(T^2 N^{2})$ multiplications for self-attention computation. However, this complexity is reduced to $O(N^2)$ in the multi-shot mode.

%%%%%%%%%%%%%%%%%%%%%%%%%%%%%%%%%%%%%%%%%%%%%%%%%%%
\begin{figure*}[ht]
    \centering
    \includegraphics[trim={2 2 2 2},clip, width=0.4\linewidth]{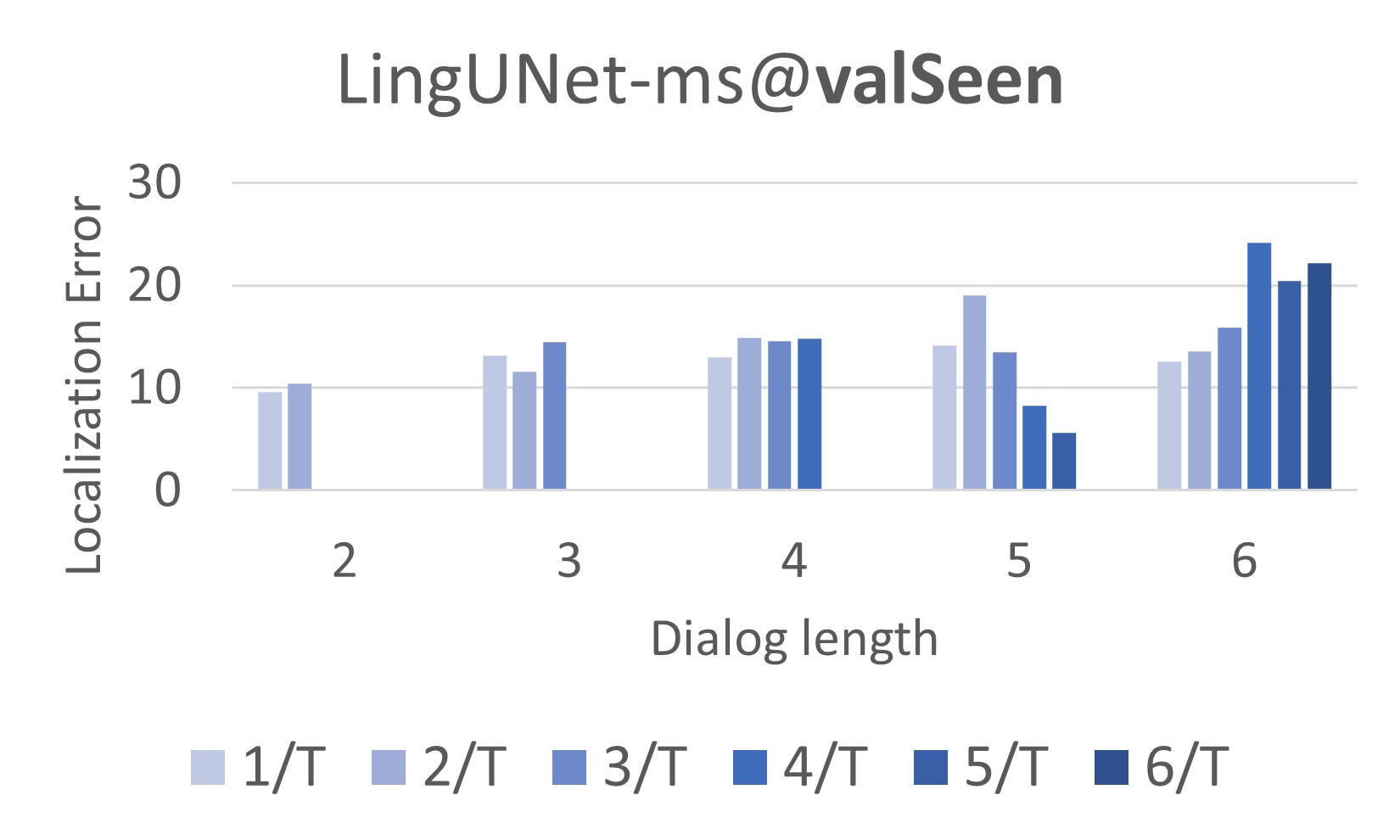}
    \includegraphics[trim={2 2 2 2},clip, width=0.4\linewidth]{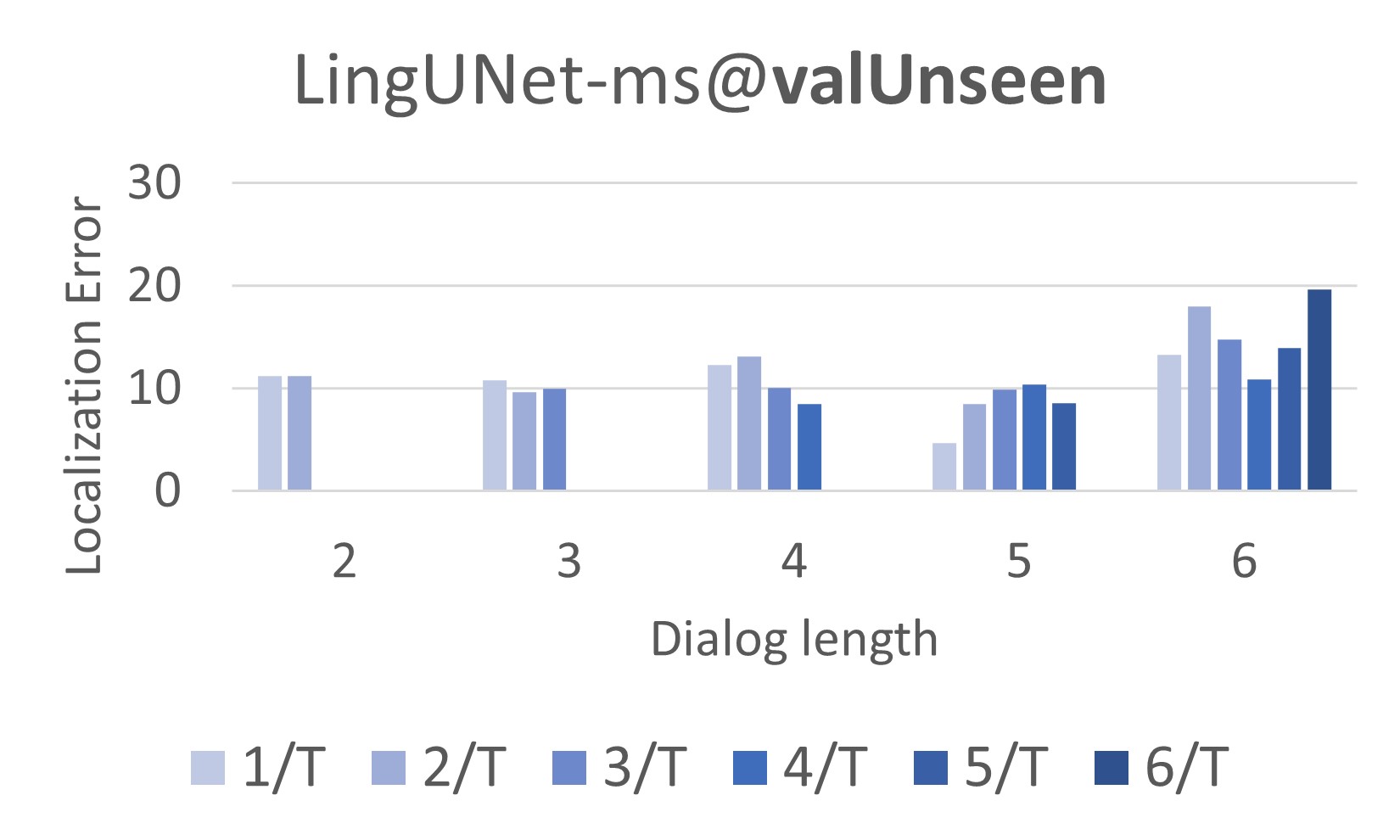} \\
    \includegraphics[trim={2 2 2 2},clip, width=0.4\linewidth]{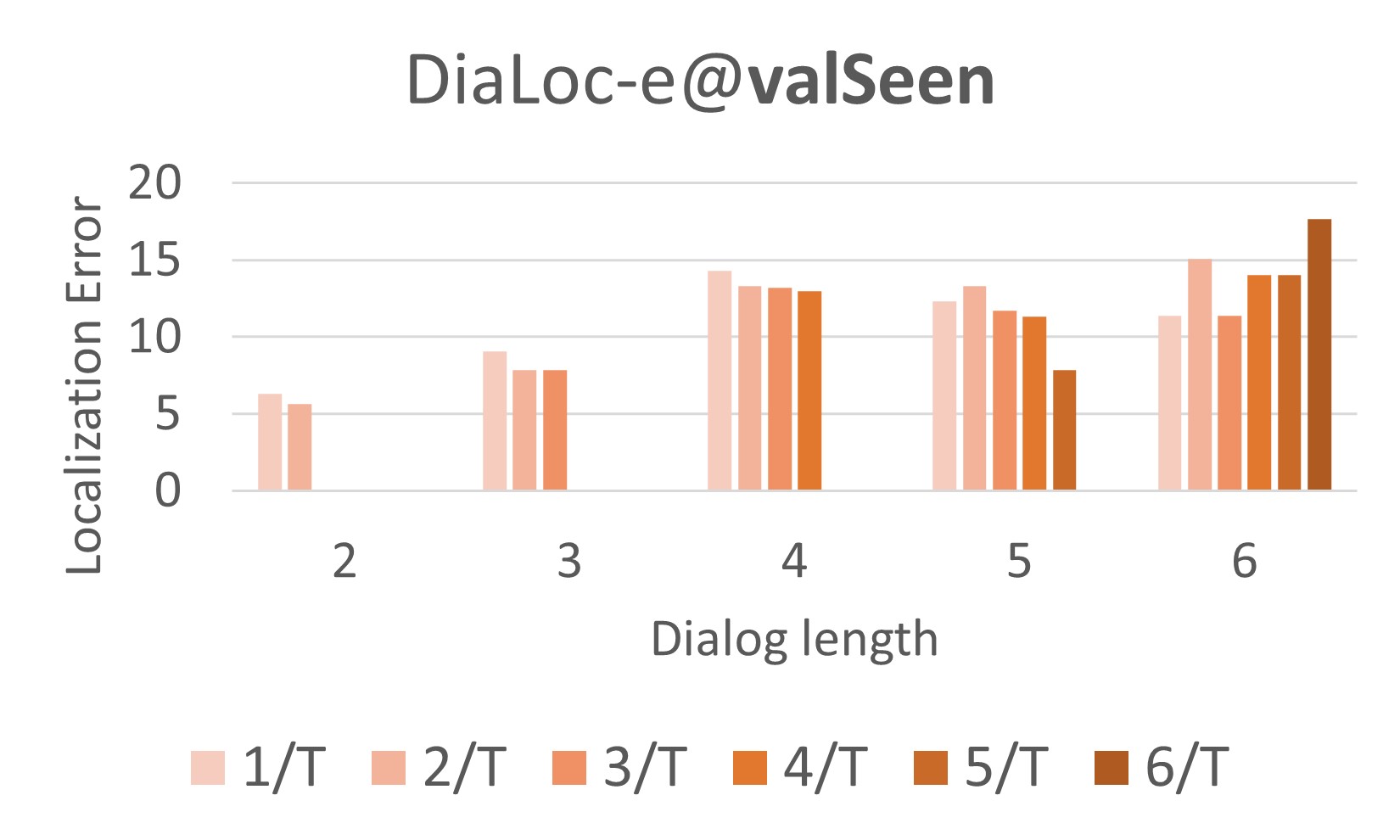}
    \includegraphics[trim={2 2 2 2},clip, width=0.4\linewidth]{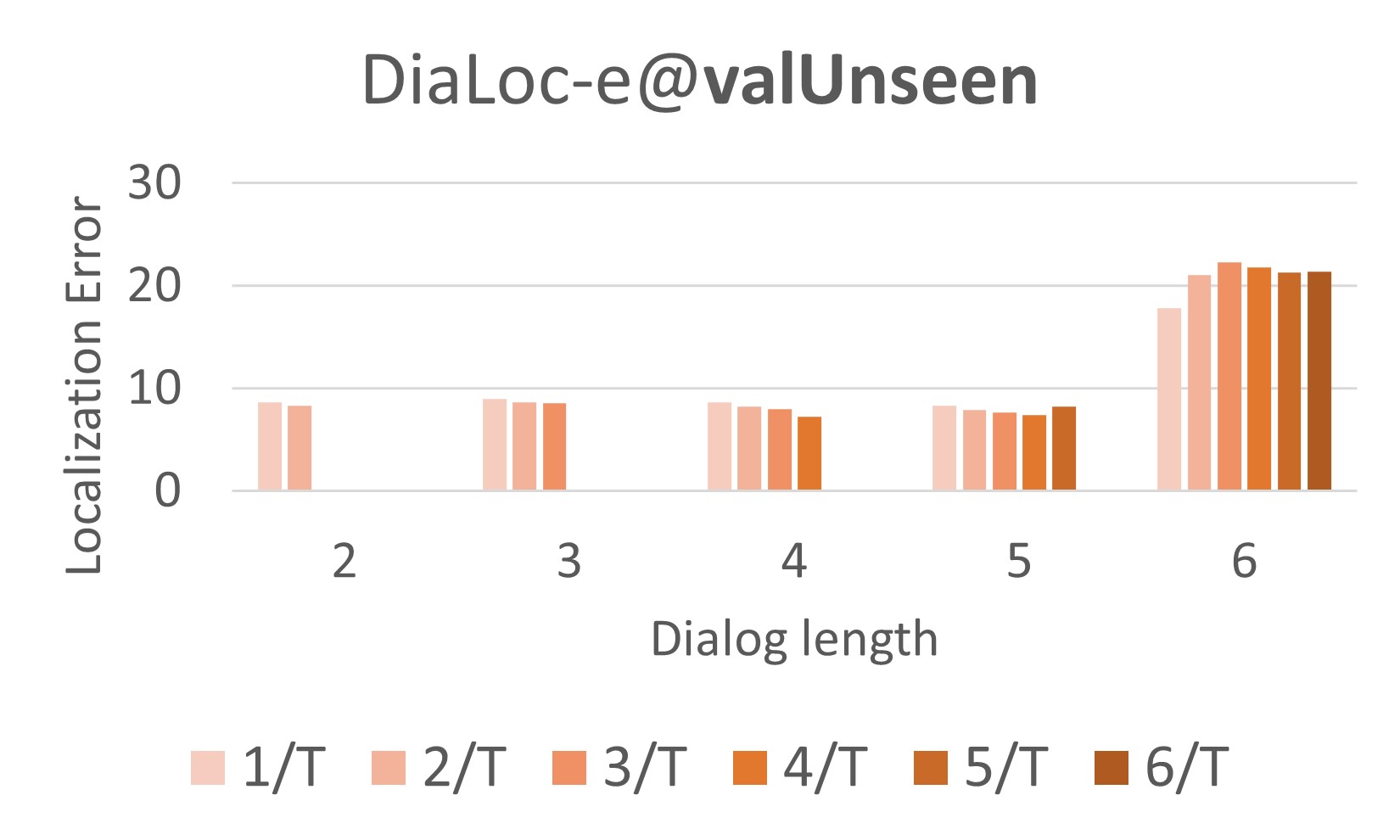}
    \caption{\textbf{Multi-shot localization error analysis.} To study performance across varied dialog length, we group samples based on their length $T$ and report LE for each group. Within each sub-plot, the LE at $t$ where $1 \le t\le T$ is detailed. Our method depicts a trend that more turns is helpful to localization while LingUNet shows mixed performance.   }
    \vspace{-10pt}
    \label{fig:finegrain}
\end{figure*}
%%%%%%%%%%%%%%%%%%%%%%%%%%%%%%%%%%%%%%%%%%%%%%%%%%%
\paragraph{CMC Curves.} 
To conduct a more comprehensive assessment of the performance of proposed method and the state-of-the-art method,
We present the Cumulative Match Characteristic (CMC) curves for both DiaLoc and LingUNet(ms) in Figure~\ref{fig:cmc}. DiaLoc consistently outperforms the baseline in both single-shot and multi-shot configurations. In valUnseen set, our multi-shot performance (red solid line) also surpasses that of the single-shot baseline (cyan solid line) and single-shot DiaLoc (blue solid line), showing improved performance in novel environments.

\paragraph{Qualitative results.}
In Figure~\ref{fig:quali}, we show two examples along with the corresponding predictions from the valSeen and valUnseen splits of WAY dataset \st{. In the odd rows, we display map (a) and the output of LingUNet (b), as well as outputs of LingUNet-ms (c-e) at different timesteps. In the even rows, we depict the target overlaid with map (a), the output of DiaLoc (b), and the outputs of the multi-shot DiaLoc variant.}{comparing LingUNet with DiaLoc.} Our single-shot method yields more precise and concentrated predictions. Regarding the multi-shot results, DiaLoc showcases the capability to rectify its previous predictions by leveraging the most recent dialog information as guidance.

\subsection{Multi-shot Predictions Analysis}
As one of the most attractive aspects, employing multi-shot localization holds the potential to early terminate the dialog in real-world searching and rescue applications.  
In this section, we analyze the performance of multi-shot methods using dialog up to timestep $t$. 
There exist several ways to determine the predicted locations. Examples encompass hard argmax and soft threshholding. Similar to single-shot evaluation, we use argmax to identify the top-1 location based on the predicted heatmap.
Alternatively,  soft threshholding identifying a set of top-K locations with probabilities surpassing the threshold. Subsequently, evaluation metrics such as Recall and Precision can be employed for measuring the performance. For simplicity, we adopt argmax as a straightforward way to evaluate the interim predictions, presenting the results in Figure~\ref{fig:finegrain}. \st{}{This will be unfair if multiple peaks show up, and one is true. It is worth looking at the prediction probability at ground-truth pixel.}

To comprehend performance across various dialog length $T$, we group samples from valSeen and valUnseen based on their respective dialog length. 
Each plot showcases the mean localization error (LE) for different sample groups. In each sub-plot, the LE at $t/T$ where $ 1 \le t\le T$ depicts the performance using incomplete dialog. 
Our method consistently demonstrates a trend wherein using more turns leads to decreased LE across valSeen and valUnseen split, while LingUNet exhibits varied behavior. Another notable observation is that both methods exhibit lower performance in the case of lengthier dialogs, particularly struggling with dialogs comprising $T=6$ turns. Introducing new turns does not yield performance improvements, and is possibly due to the unbalanced training data.

\section{Conclusion}
Localization using dialog is an important task in the field of Embodied AI and significant progress has been achieved thanks to the advancements in vision and language learning. In this work, we propose a novel approach DiaLoc for iterative dialog-based localization. The key idea is to overcome the limitations of existing methods that requires entire dialog for localization. We introduce DiaLoc, a multimodal localizer, to continually fuse visual and dialog inputs for accurate location prediction. Our work narrows the gap between simulation and real-world applications,  opening doors for future work on multimodal dialog generation and collaborative localization. 

%There are a few limitations worth emphasizing. Firstly, the proposed DiaLoc, employing a Transformer-based architecture, achieves state-of-the-art performance at the cost of high memory usage and computational expenses. Secondly, the evaluation dataset used in this study is imbalanced in terms of dialog length, and the performance on lengthy dialogs is unsatisfactory.
% %%%%%%%%%%%%%%%% TODO List %%%%%%%%%%%%%%%%%%
% {\color{red}
%  \begin{enumerate}    
%     \item Evaluations
%     \item Led-full, led-seq, both running on GPU1 --> quali fig, CMC
%     \item transformer base (run on SCAN4)  --> quali fig, CMC, SOTA tab

%     \begin{enumerate}
%         \item final pred only: le, acc
%         \item fine-grained eval: acc for different turns
%         \item intermediate preds: x-axis (turns), y-axis (le, thresh(set a threshold, all preds above it considered, min(le))) 
%     \end{enumerate}
%     \item Aux loss
%     \begin{enumerate}
%         \item weighted sum: $f^{(T-t)}$, choice of f (0, 0.5, 0.9, 1)
%         \item auxiliary loss: clipseg attention map
%         \item auxiliary loss: binary mse
%     \end{enumerate}
%     \item Test set
%     \begin{enumerate}
%         \item fullDialog, our, depth=1
%         \item seq dialog, our, depth=1
%         \item fullDialog, led
%         \item seq dialog, led
        
%     \end{enumerate}
%     \item New idea: denoising approach :)
% \end{enumerate}
% }
% %%%%%%%%%%%%%%%%%%%%%%%%%%%%%%%%%%%%%%%%%%%%%%%%%%%
\section{Supplementary Materials}
\begin{figure}[ht]
    \centering
    \includegraphics[width=0.95\linewidth]{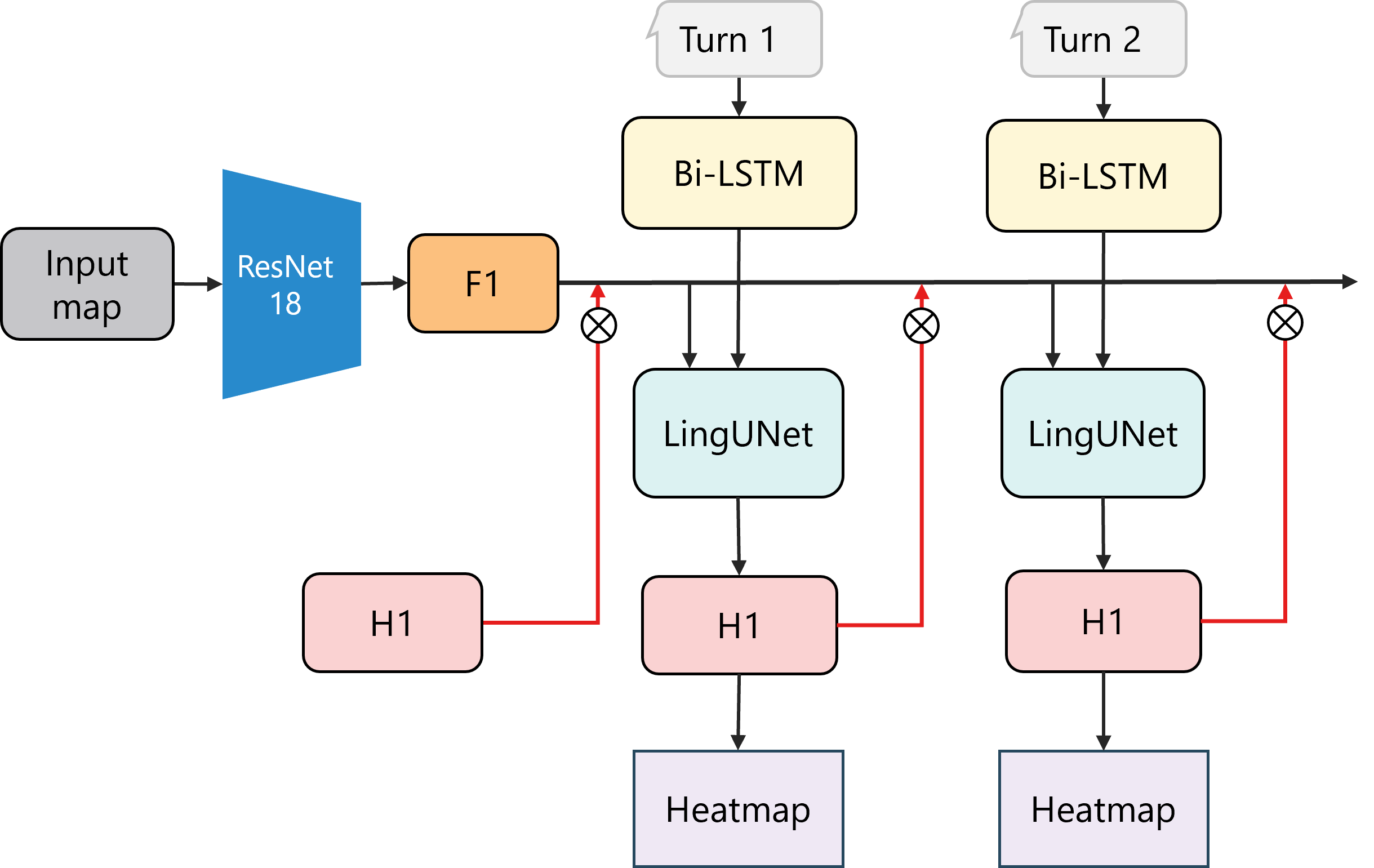}
    \caption{\textbf{LingUNet-ms: the adapted multi-shot multimodal LingUNet for iterative embodied dialog localization.}  }
    \label{fig:unet-ms}
\end{figure}

\paragraph{Multi-shot Adaptation of LingUNet.}
LingUNet was designed for single-shot dialog localization.  To enable it works for multi-shot scenario so that fair evaluation is possible under our iterative formulation, we make optimal modifications according to our proposed DiaLoc-e. 
Basically, the idea is to leverage hidden states of previous iteration for future predictions. The adapted LingUNet-ms is illustrated in Figure~\ref{fig:unet-ms}. Similar to our DiaLoc-e, the map embedding $\text{F1}$ generated by ResNet18 is fed to LingUNet alongside the dialog embedding to generate hidden states $\text{H1}$. The hidden states is then used as input to predict location heatmaps. In particular, at each timestep $t$, the hidden states of previous timestep $\text{H}1_{t-1}$ is fused with $\text{F1}$ to integrate the dynamic prior information.   
%%%%%%%%%%%%%%%%%%%%%%%%%%%%%%%%%%%%%%%%%%%%%%%%%%%%%%%%%%%%%
%%%%%%%%%%%%%%%%%%%%%%%%%%%%%%%%%%%%%%%%%%%%%%%%%%%%%%%%%%%%%

\paragraph{Dialog Augmentation using LLM.}
In this section, we provide additional details for leveraging LLM to augment localization dialogs in WAY dataset. 
We employ gpt-3.5-turbo-16k as the LLM instance to rewrite the ground-truth dialogs of training set. We use the prompt as `` Paraphrase the dialog". We set the temperature to 0.6 and the top-p to 0.5 in the API call. 

\begin{figure*}[t]
    \centering
    \includegraphics[width=0.95\linewidth]{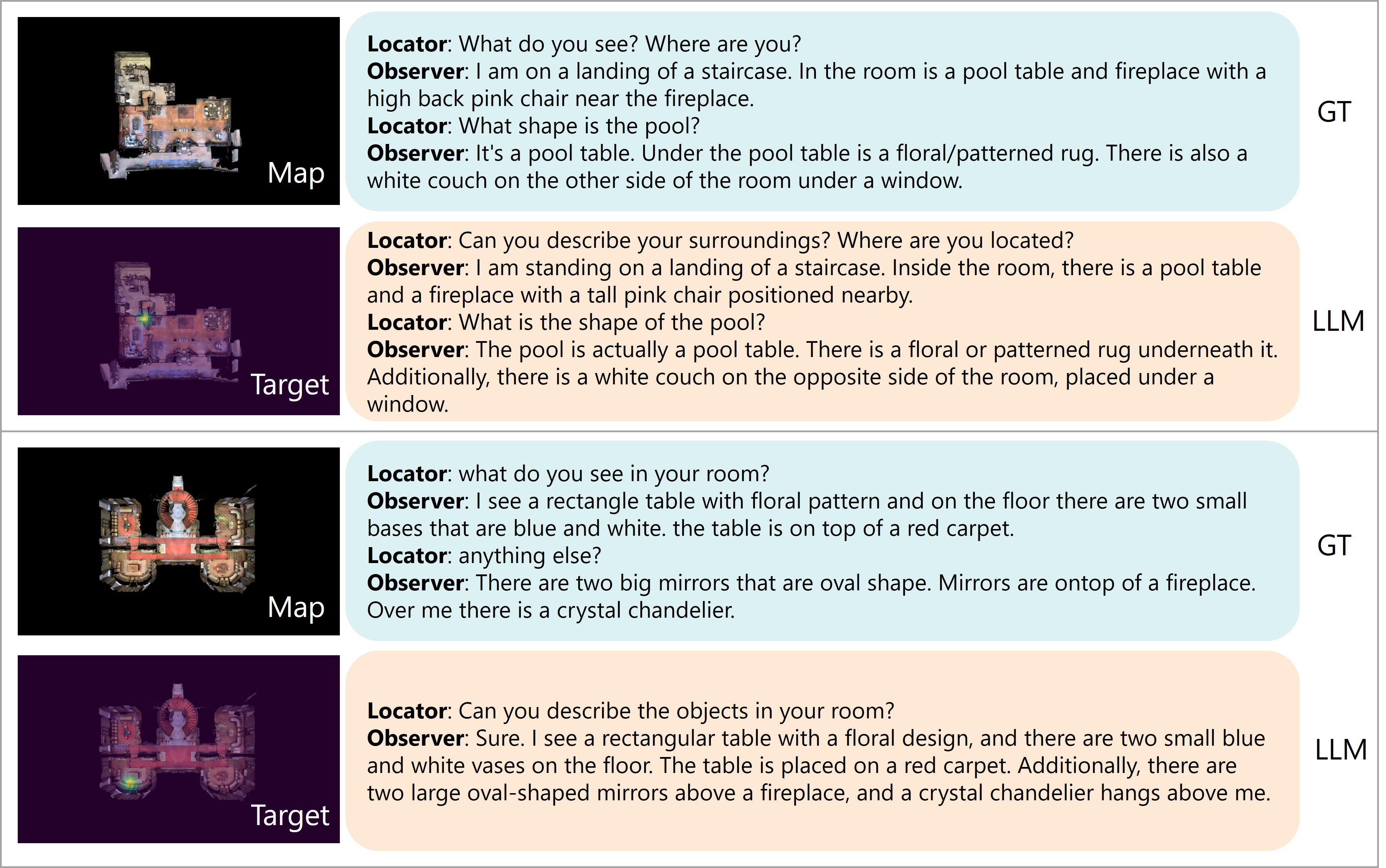}
    \caption{\textbf{Examples of ground-truth dialogs and augmented version using LLM}. }
    \label{fig:gpt}
\end{figure*}

In Figure~\ref{fig:gpt}, we show two examples from the train split of WAY. For each example, we display the top-down map and the corresponding target on the left. On the right size, the GT dialog is shown at the top within the blue box, and the para-phased version is shown inside the orange box. In both cases, we can see that GPT generates semantically consistent dialogs as the original version. In the second case, GPT reduced the length of the original dialog without changing the meaning. Note that, the GPT API does not use map information at all and is purely text-based.
%%%%%%%%%%%%%%%%%%%%%%%%%%%%%%%%%%%%%%%%%%%%%%%%%%%%%%%%%%%%%
%%%%%%%%%%%%%%%%%%%%%%%%%%%%%%%%%%%%%%%%%%%%%%%%%%%%%%%%%%%%%

\begin{figure*}[ht]
    \centering
    \includegraphics[trim={2 2 2 2},clip, width=0.43\linewidth]{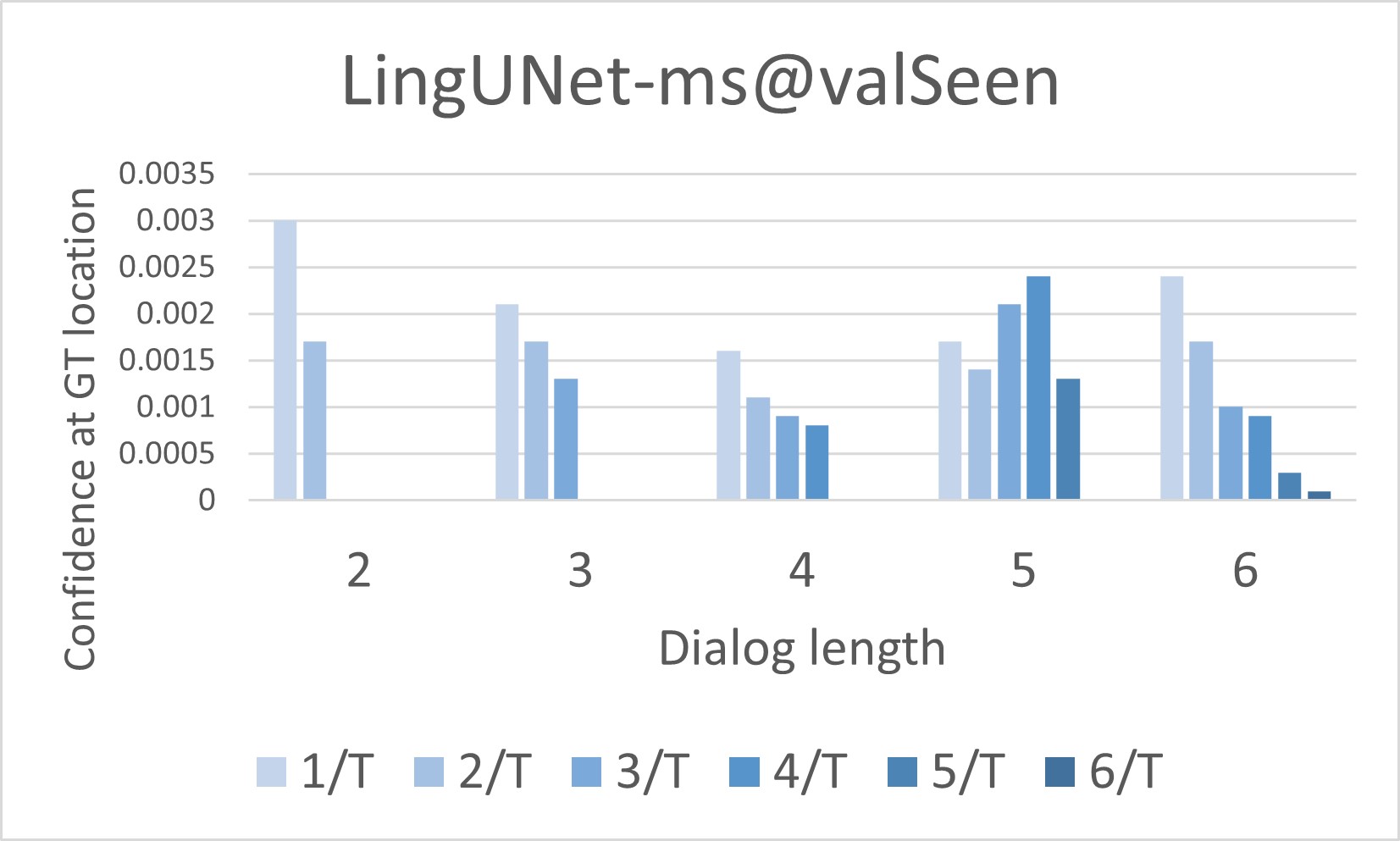}
    \includegraphics[trim={2 2 2 2},clip, width=0.43\linewidth]{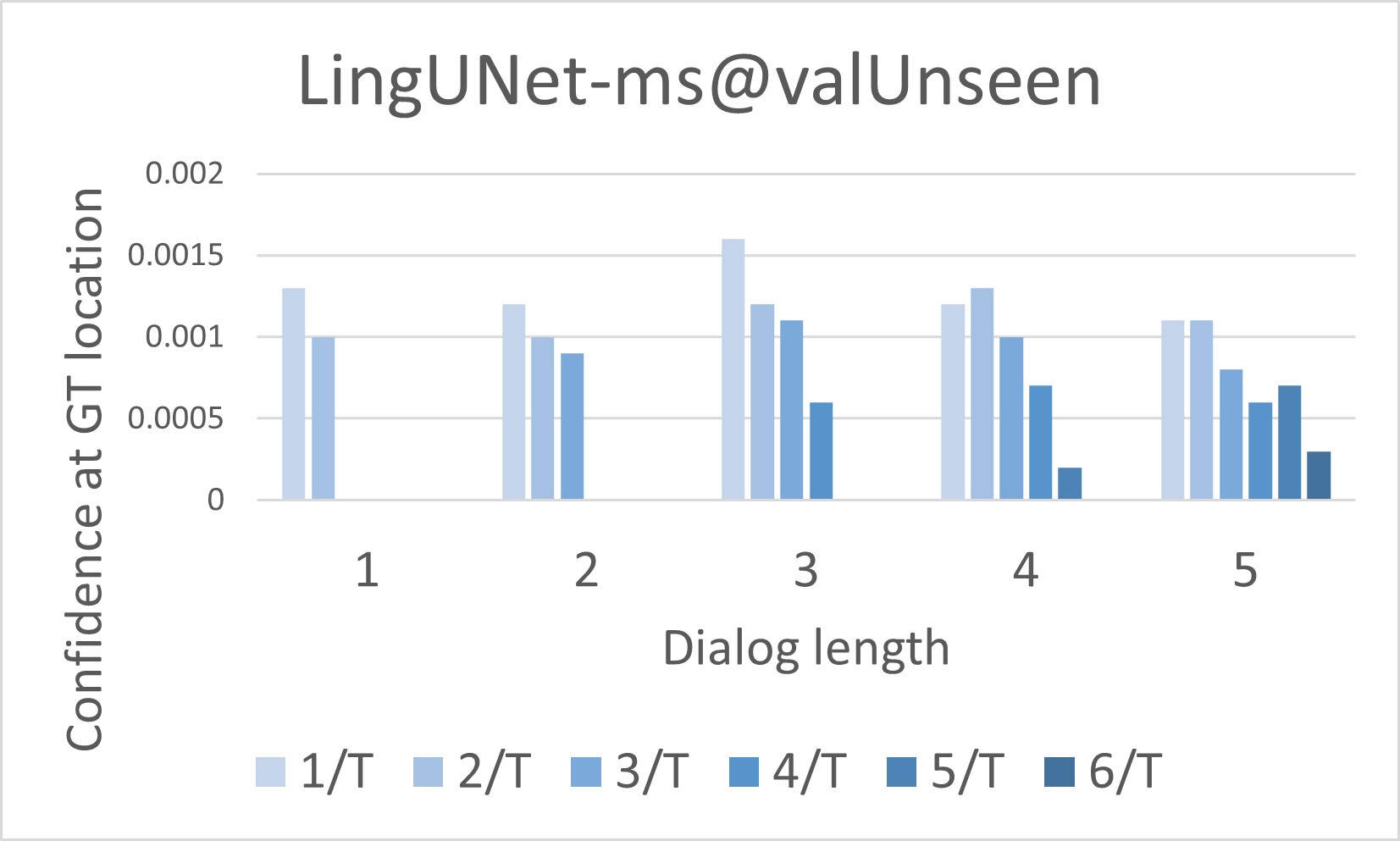} \\
    \includegraphics[trim={2 2 2 2},clip, width=0.43\linewidth]{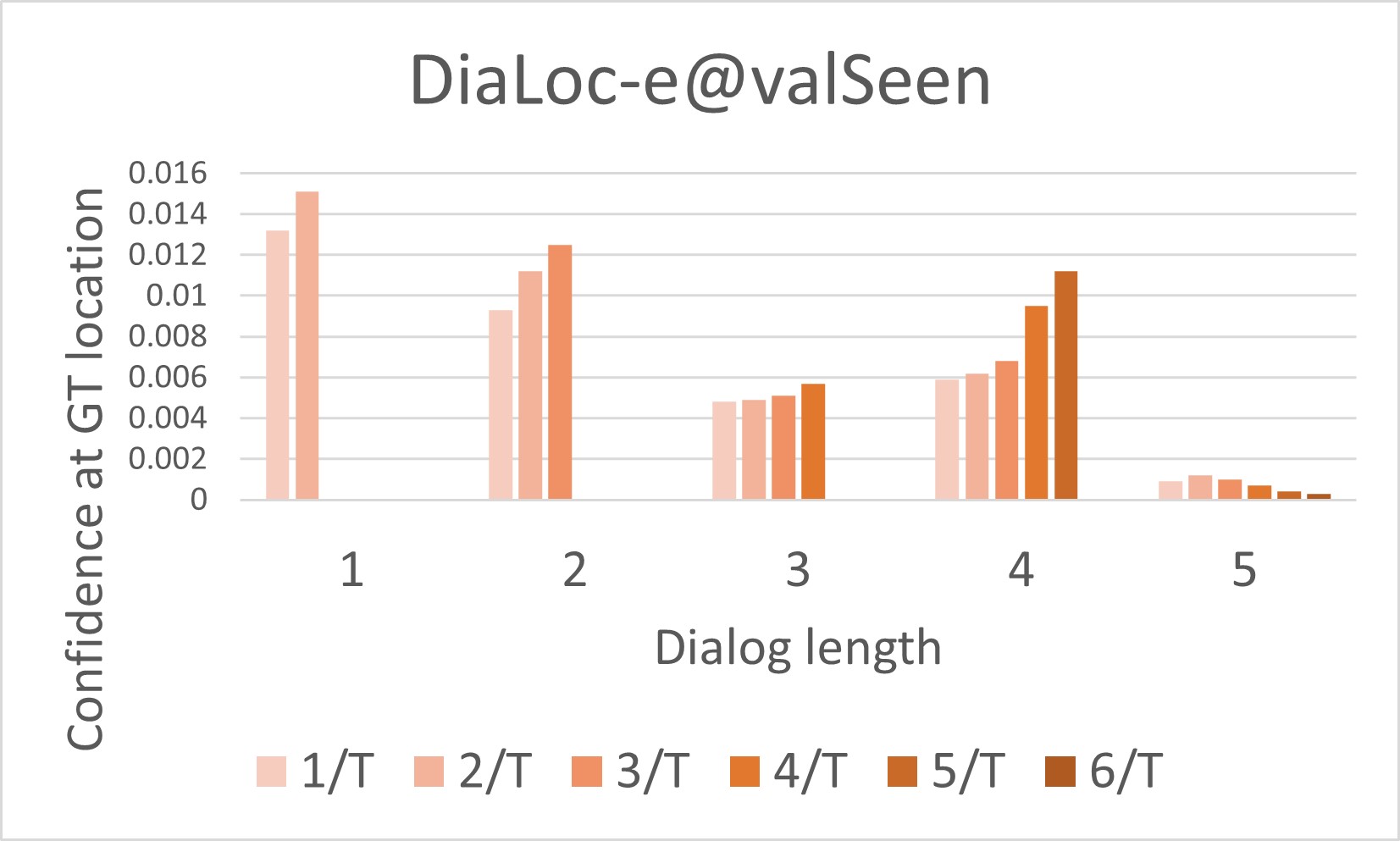}
    \includegraphics[trim={2 2 2 2},clip, width=0.43\linewidth]{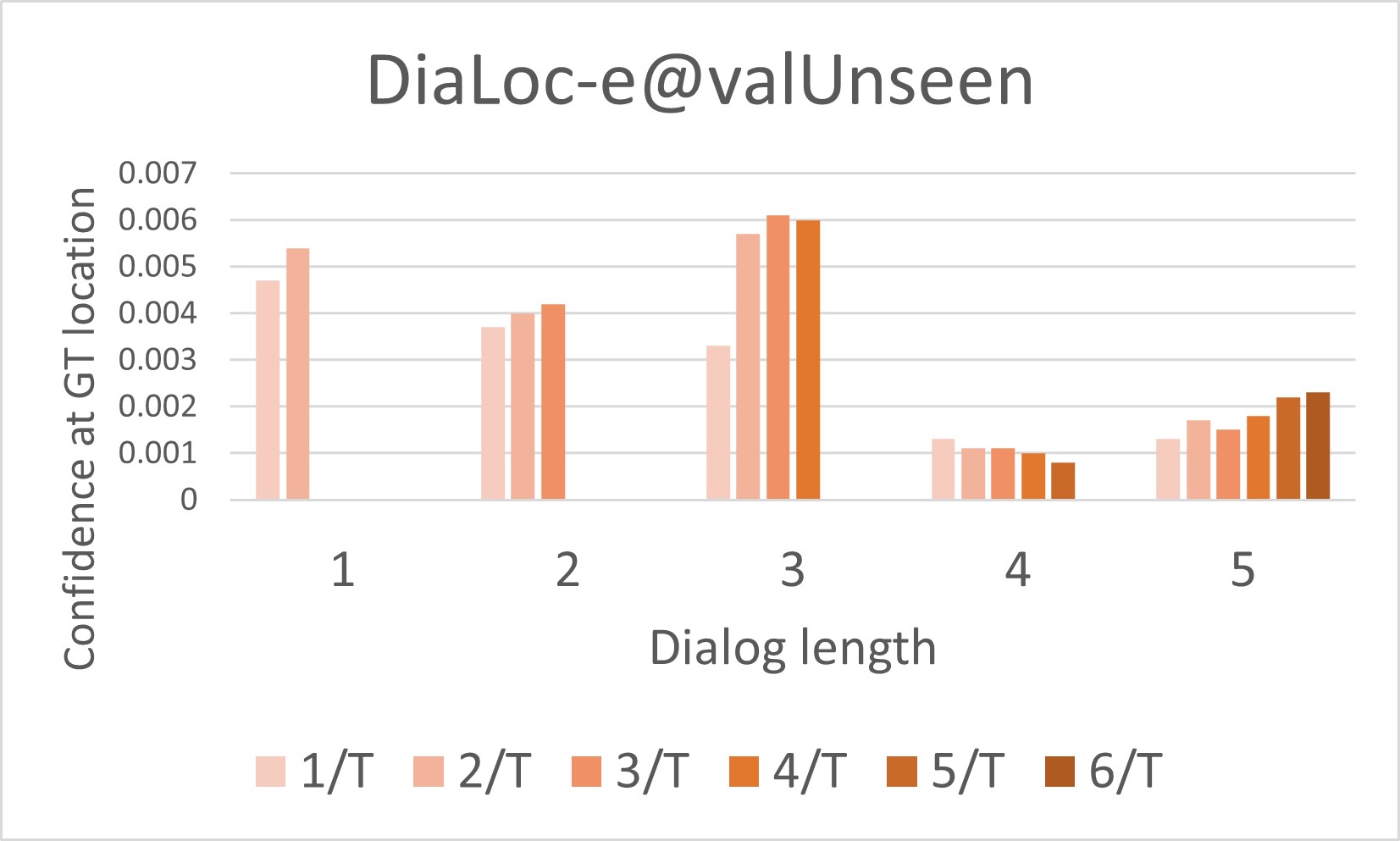}
    \caption{\textbf{Multi-shot prediction confidence analysis.} To study performance across varied dialog length, we group samples based on their length $T$ and report average prediction confidence for each group. Within each sub-plot, the PC at $t$ where $1 \le t\le T$ is detailed. Our method depicts a trend that more turns is helpful to increase the confidence at the desired location while LingUNet shows negative trend.   }
    \vspace{-10pt}
    \label{fig:confidence_plot}
\end{figure*}

\paragraph{Multi-shot analysis via prediction confidence.}
As one of the most attractive aspects, employing multi-shot localization holds the potential to early terminate the dialog in real-world searching and rescue applications.  
In this section, we analyze the performance of multi-shot methods using dialog up to timestep $t$. 
In addition to the localization error based on top-1 prediction, we employ prediction confidence as an alternative in this analysis. Localization error as a metric will be unfair in case of multiple peaks show up, and one is true positive. To overcome this issue, given heatmap prediction, we report the pixel-wise probability at ground-truth location in Figure~\ref{fig:confidence_plot}. In summary, our method depicts a trend that more turns is helpful to increase the confidence level at the desired location while LingUNet shows a negative trend. 

%%%%%%%%%%%%%%%%%%%%%%%%%%%%%%%%%%%%%%%%%%%%%%%%%%%
%%%%%%%%%%%%%%%%%%%%%%%%%%%%%%%%%%%%%%%%%%%%%%%%%%%

\paragraph{Inference runtime and memory usage.}
One of the limitations of the proposed approach is the memory usage and decreased inference efficiency. 
We evaluate DiaLoc-e (depth=1) in the multi-shot mode on the valSeen split of WAY. The evaluation results are shown in Table~\ref{tab:runtime}. 

\begin{figure*}[t]
    \centering
    \setlength{\tabcolsep}{1pt}
    \renewcommand{\arraystretch}{0.4}
    \begin{tabular}{p{10pt} c c c c c}
          % seen, 87
       \rotatebox{90}{LingUNet} & \begin{tikzpicture}
            \draw (0, 0) node[inner sep=0]
            {\includegraphics[width=0.2\linewidth]{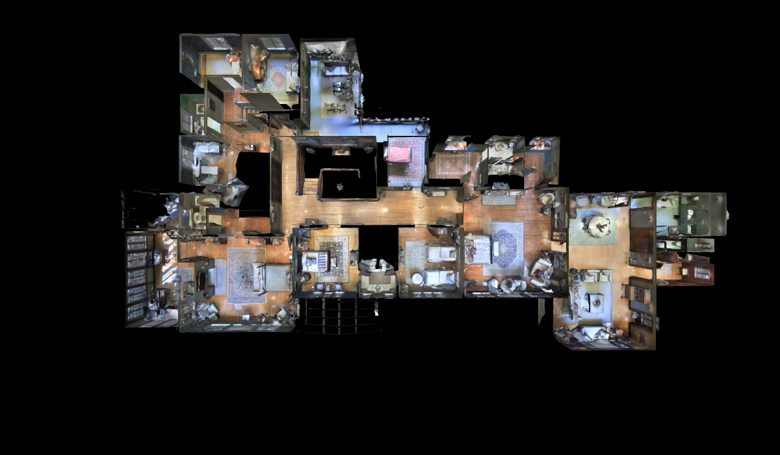}};
            \draw (1.4, 0.8) node[color=white,font=\small] {Map};
        \end{tikzpicture} &
         \includegraphics[width=0.2\linewidth]{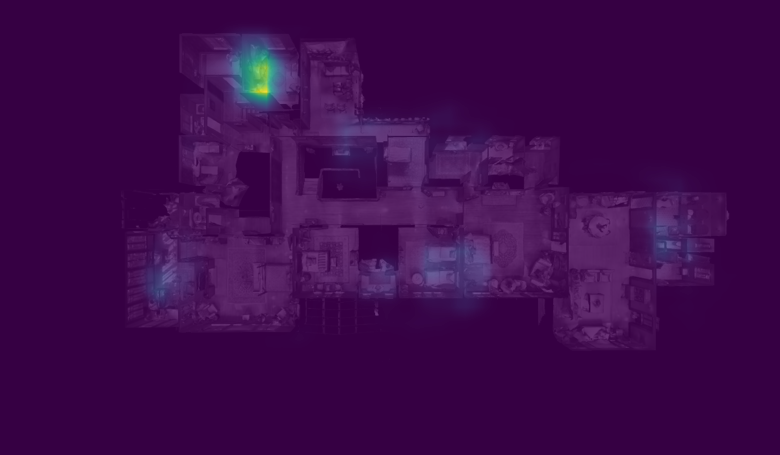} &
         \includegraphics[width=0.2\linewidth]{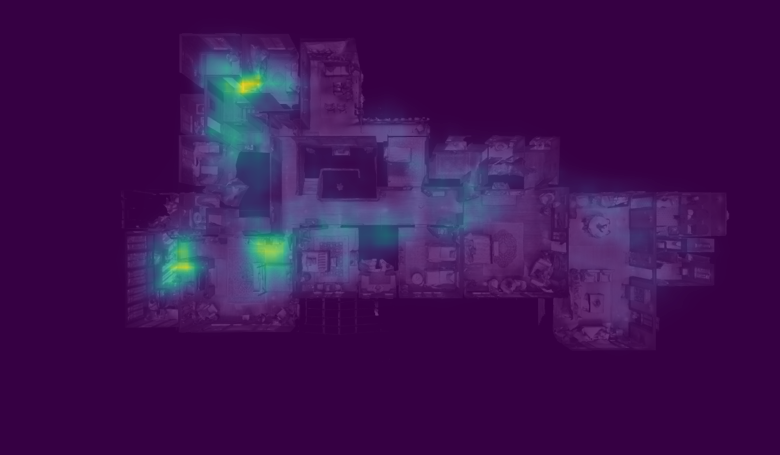} &
         \includegraphics[width=0.2\linewidth]{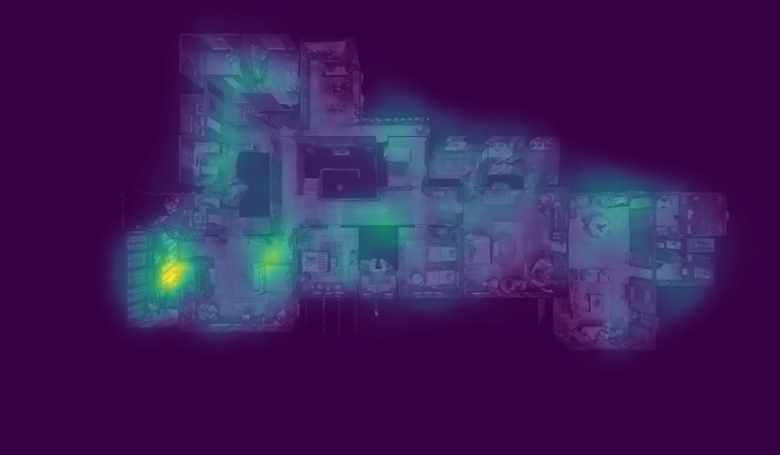} &
         \includegraphics[width=0.2\linewidth]{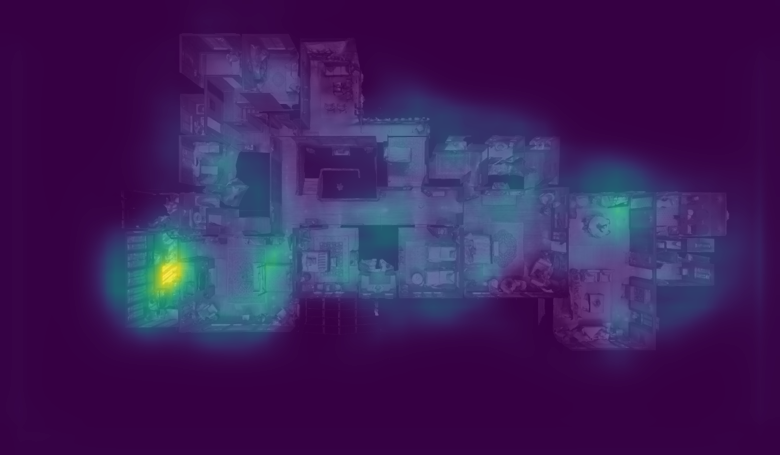} 
         \\
        \rotatebox{90}{DiaLoc} & \begin{tikzpicture}
            \draw (0, 0) node[inner sep=0]
            {\includegraphics[width=0.2\linewidth]{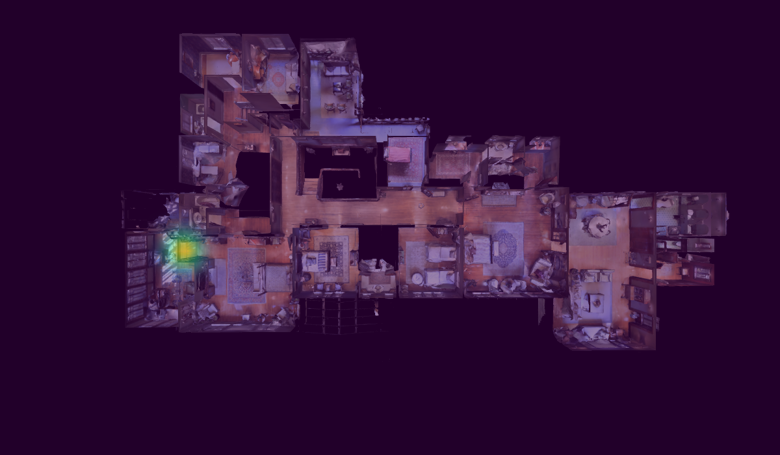}};
            \draw (1.4, 0.8) node[color=white,font=\small] {GT};
         \end{tikzpicture}  &
         \includegraphics[width=0.2\linewidth]{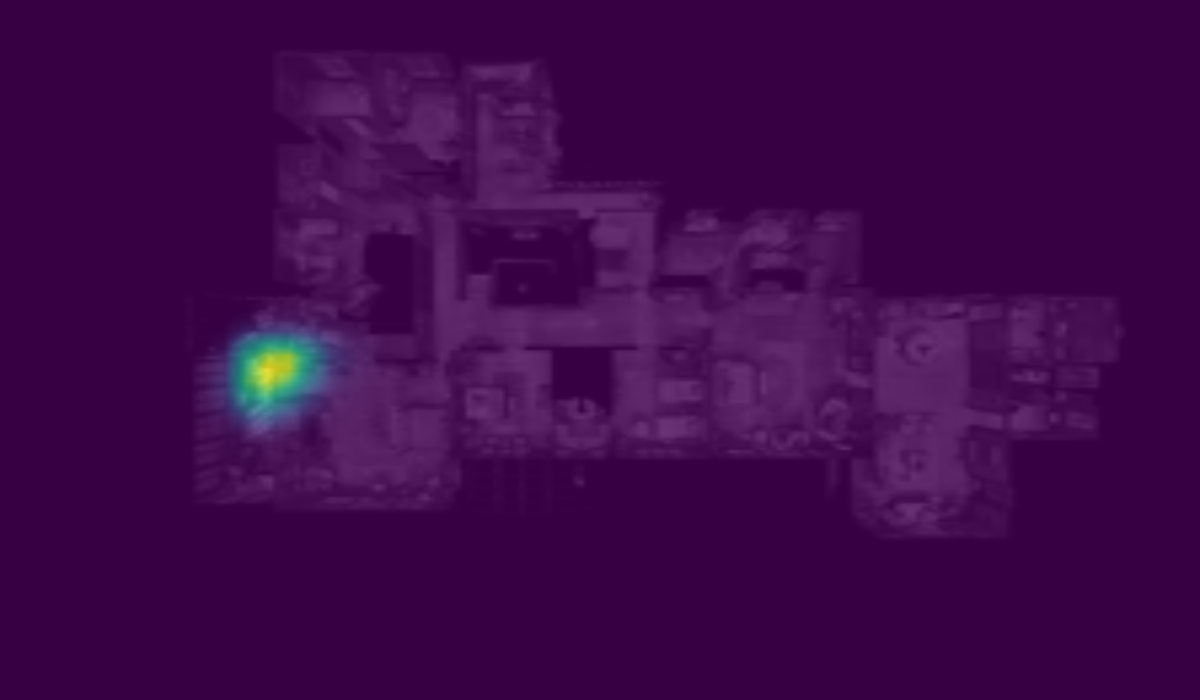} &
         \includegraphics[width=0.2\linewidth]{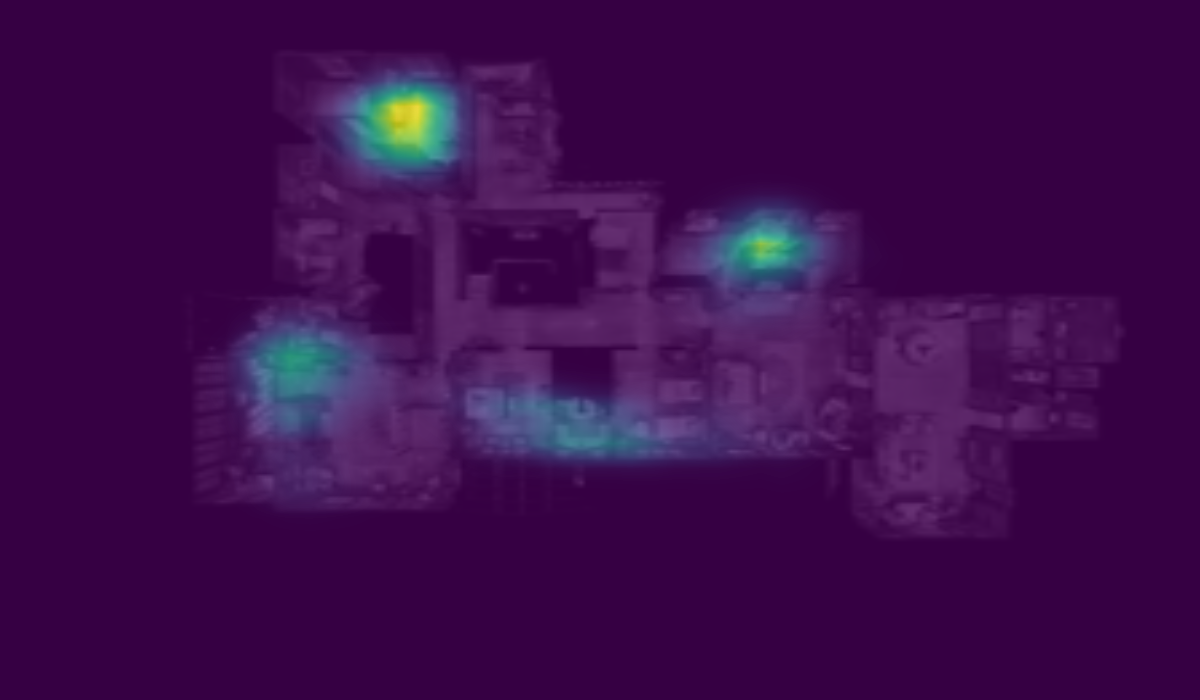} &
         \includegraphics[width=0.2\linewidth]{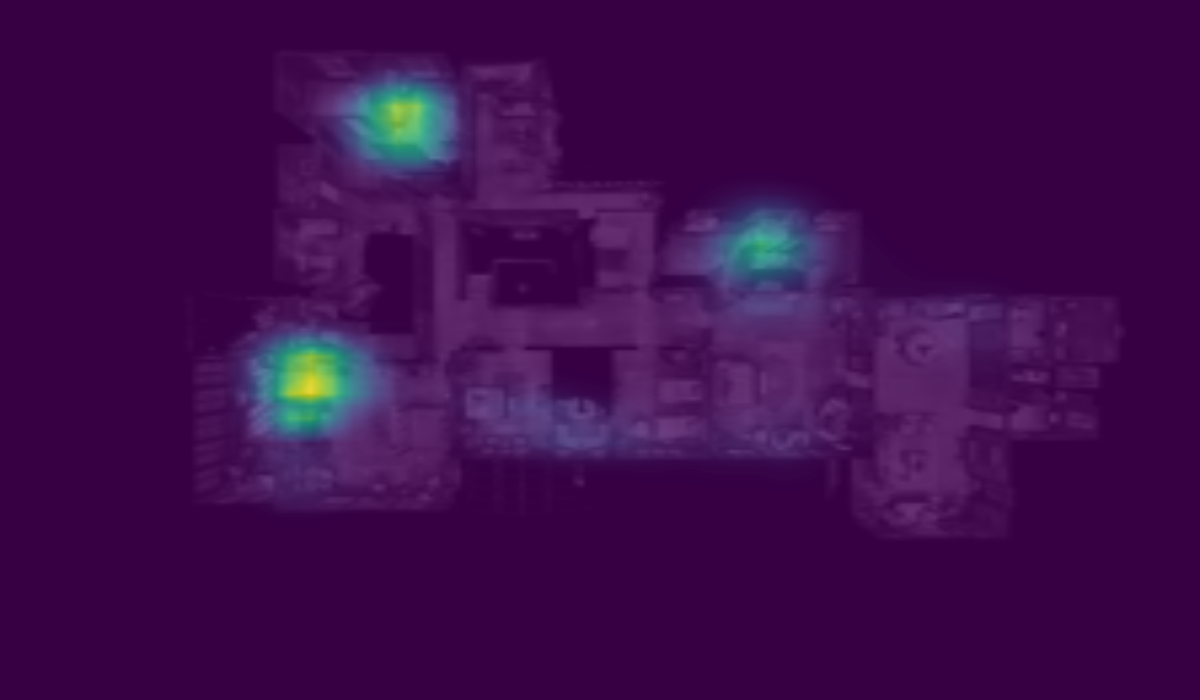} &
         \includegraphics[width=0.2\linewidth]{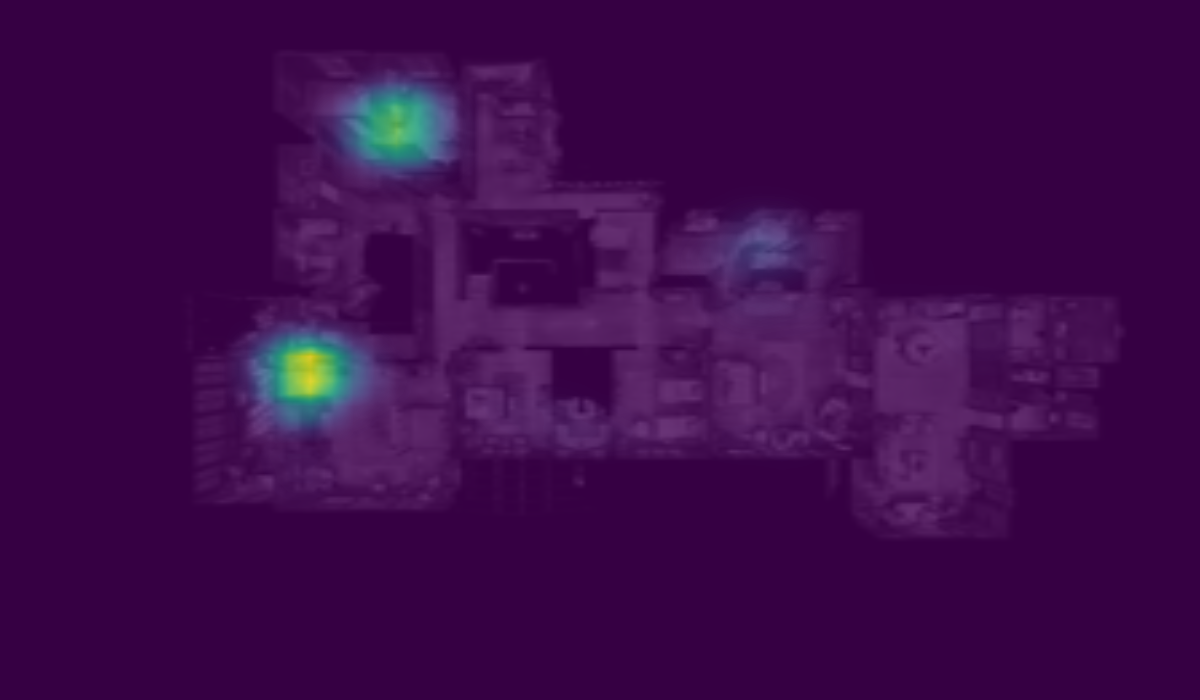} 
         \\
         & val-seen 87 & Single-shot & Multi-shot: $1/3$ & $2/3$ & $3/3$ \\
         % seen 176
         \rotatebox{90}{LingUNet} & \begin{tikzpicture}
         \draw (0, 0) node[inner sep=0] {\includegraphics[width=0.2\linewidth]{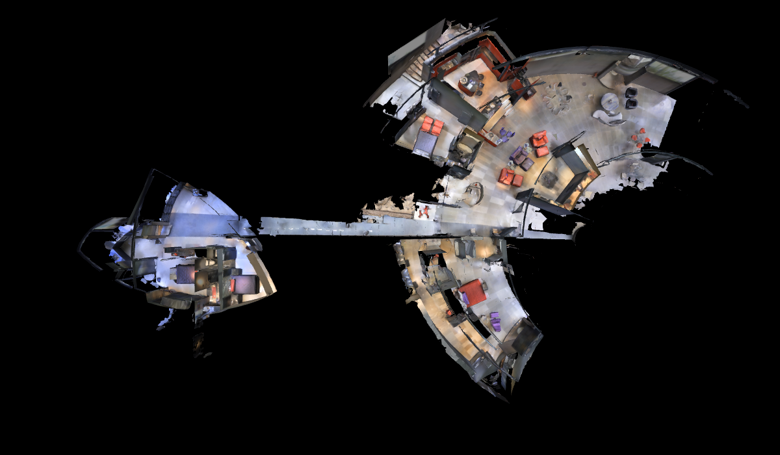}};
        \draw (-0.8, 0.8) node[color=white,font=\small] {Map};
        \end{tikzpicture} &
         \includegraphics[width=0.2\linewidth]{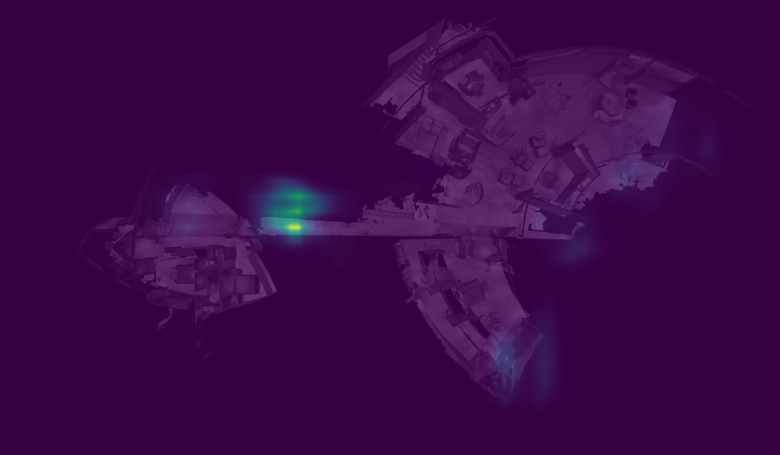} &
         \includegraphics[width=0.2\linewidth]{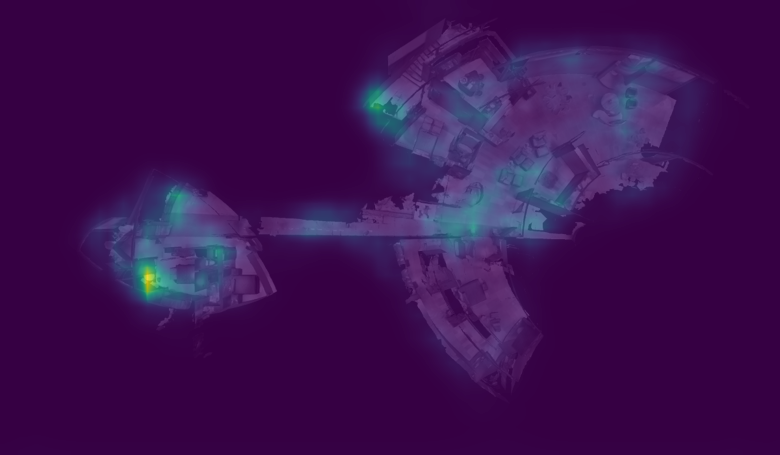} &
         \includegraphics[width=0.2\linewidth]{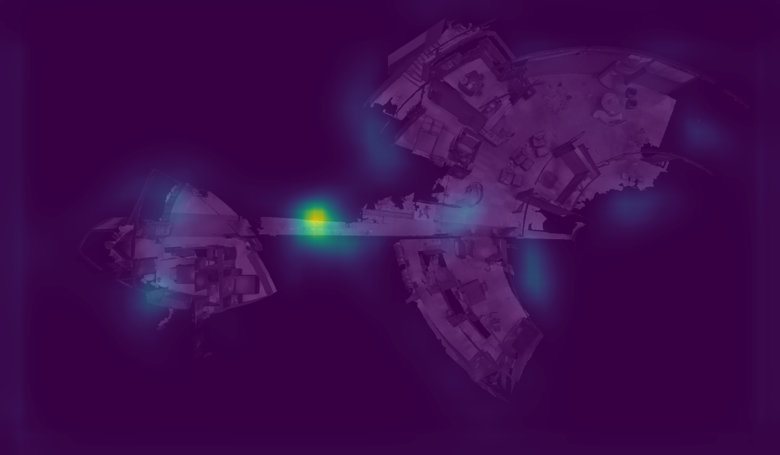} &
         \includegraphics[width=0.2\linewidth]{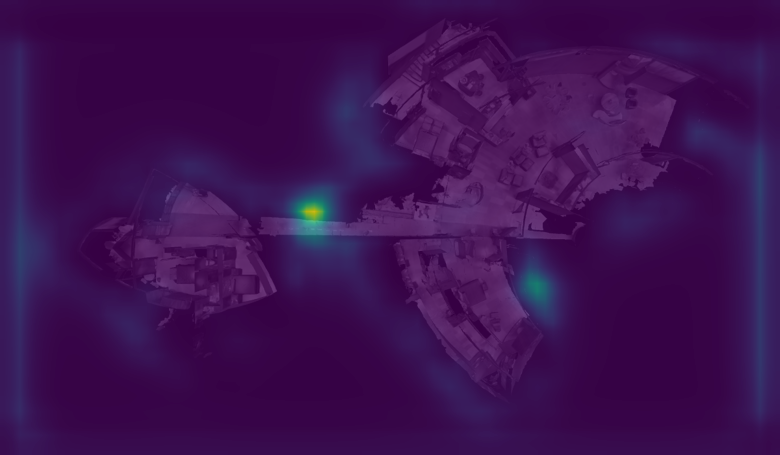} 
         \\
        \rotatebox{90}{DiaLoc} &  \includegraphics[width=0.2\linewidth]{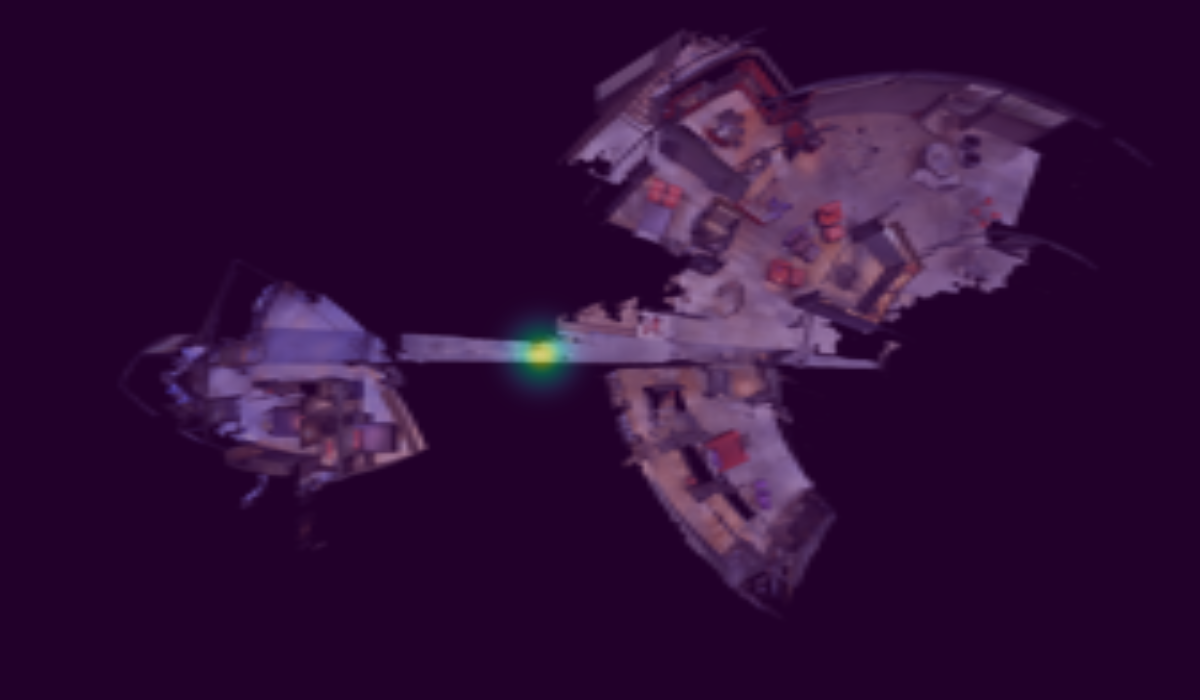} &
         \includegraphics[width=0.2\linewidth]{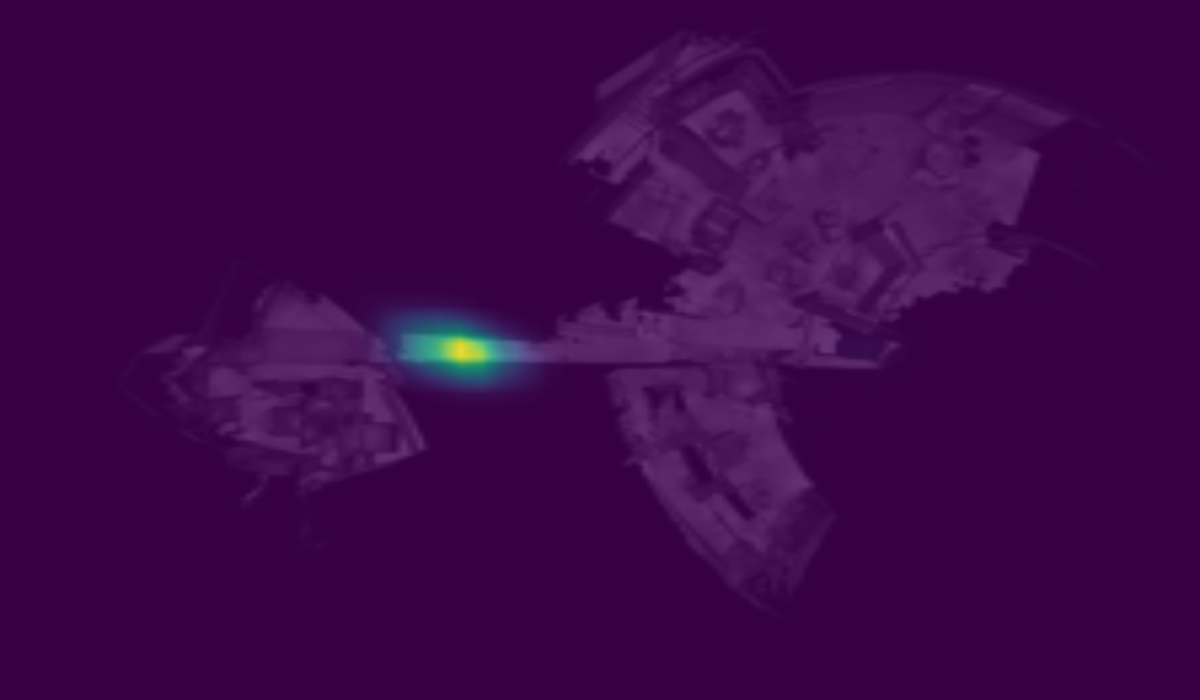} &
         \includegraphics[width=0.2\linewidth]{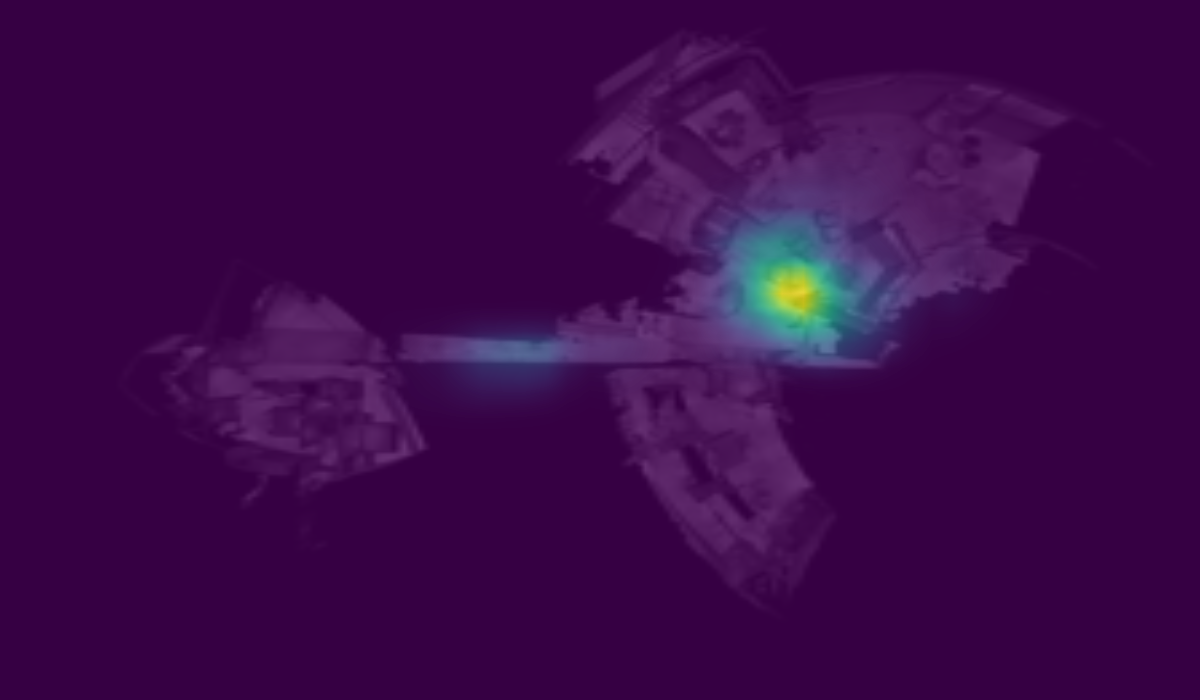} &
         \includegraphics[width=0.2\linewidth]{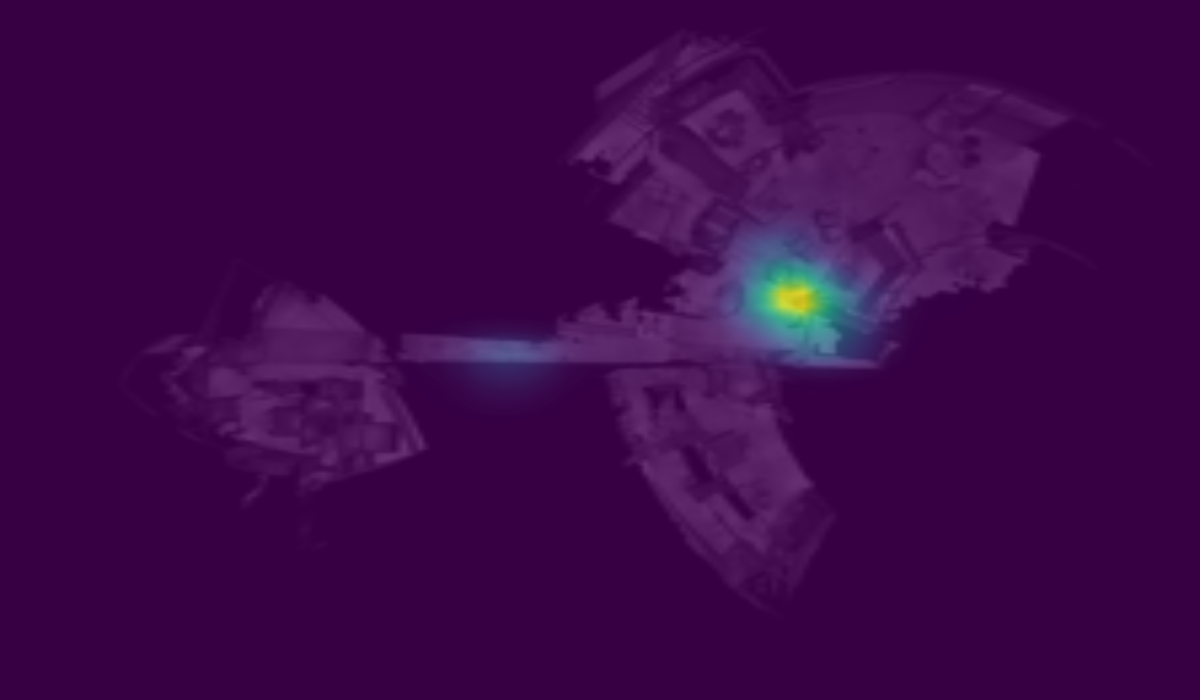} &
         \includegraphics[width=0.2\linewidth]{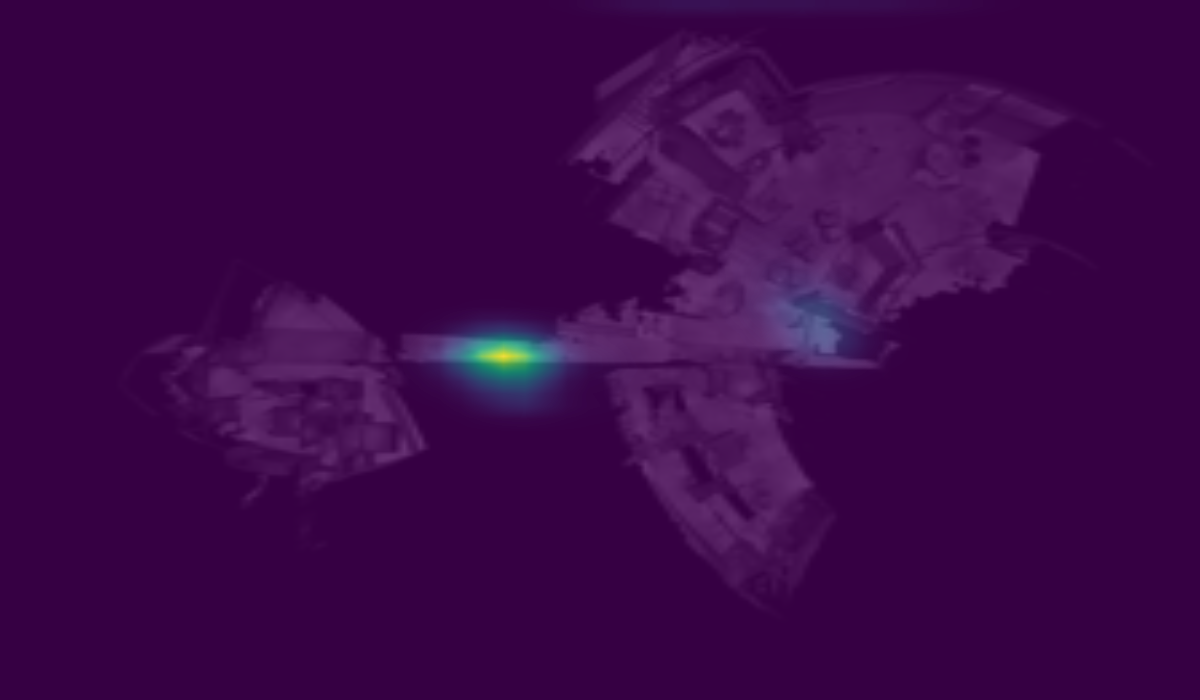} 
         \\
         & val-seen 176 & Single-shot & Multi-shot: $1/5$ & $3/5$ & $5/5$ \\ 

          % unseen, 245
         \rotatebox{90}{LingUNet} & \begin{tikzpicture}
            \draw (0, 0) node[inner sep=0] 
            {\includegraphics[width=0.2\linewidth]{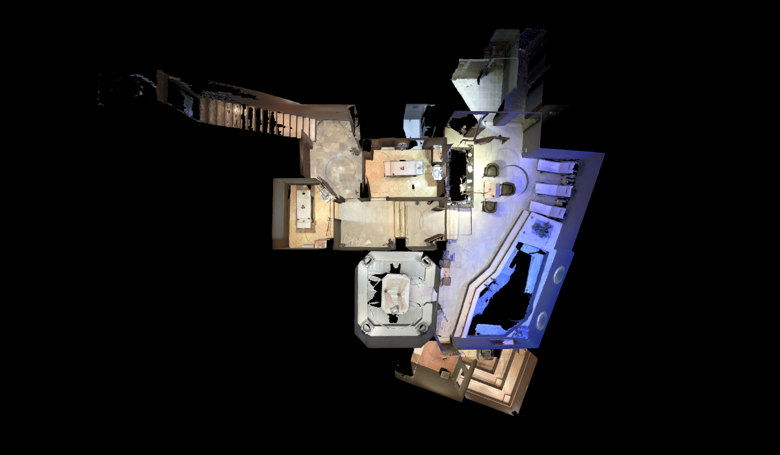}};
            \draw (1.4, 0.8) node[color=white,font=\small] {Map};
        \end{tikzpicture} &
         \includegraphics[width=0.2\linewidth]{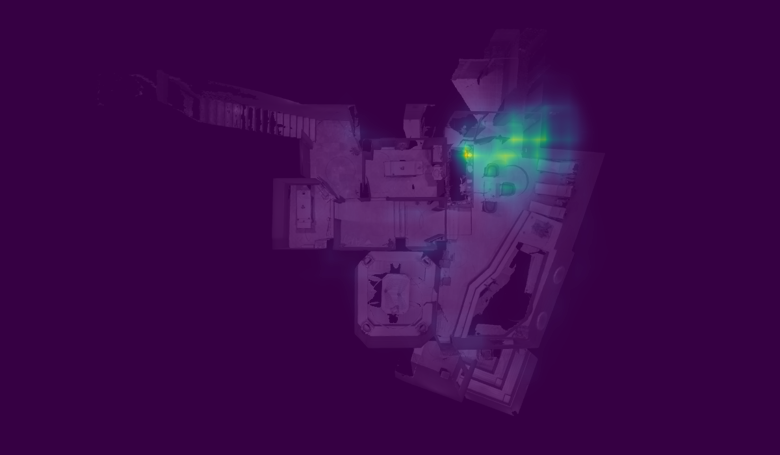} &
         \includegraphics[width=0.2\linewidth]{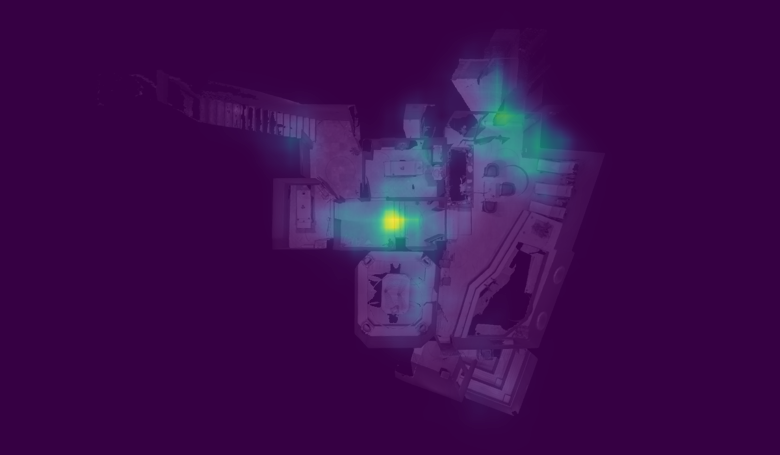} &
         \includegraphics[width=0.2\linewidth]{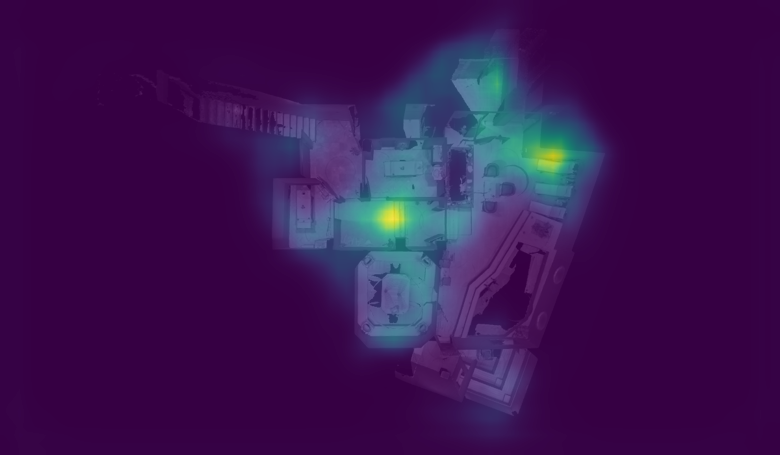} &
         \includegraphics[width=0.2\linewidth]{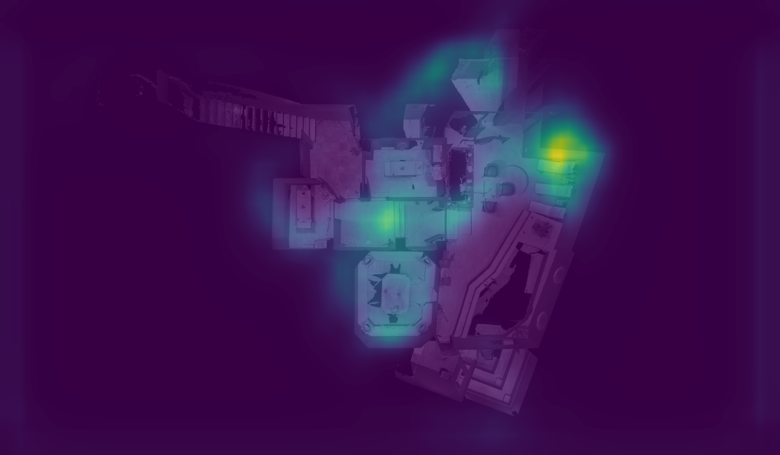} 
         \\
         \rotatebox{90}{DiaLoc} & \begin{tikzpicture}
            \draw (0, 0) node[inner sep=0] 
            {\includegraphics[width=0.2\linewidth]{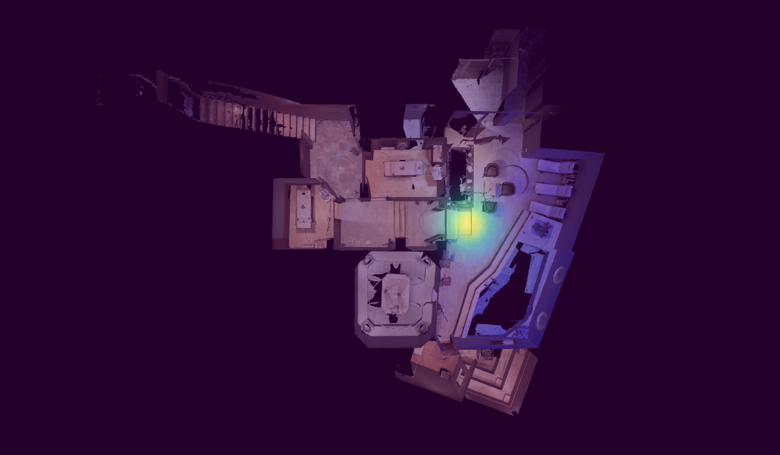}};
            \draw (1.4, 0.8) node[color=white,font=\small] {GT};
         \end{tikzpicture} &
         \includegraphics[width=0.2\linewidth]{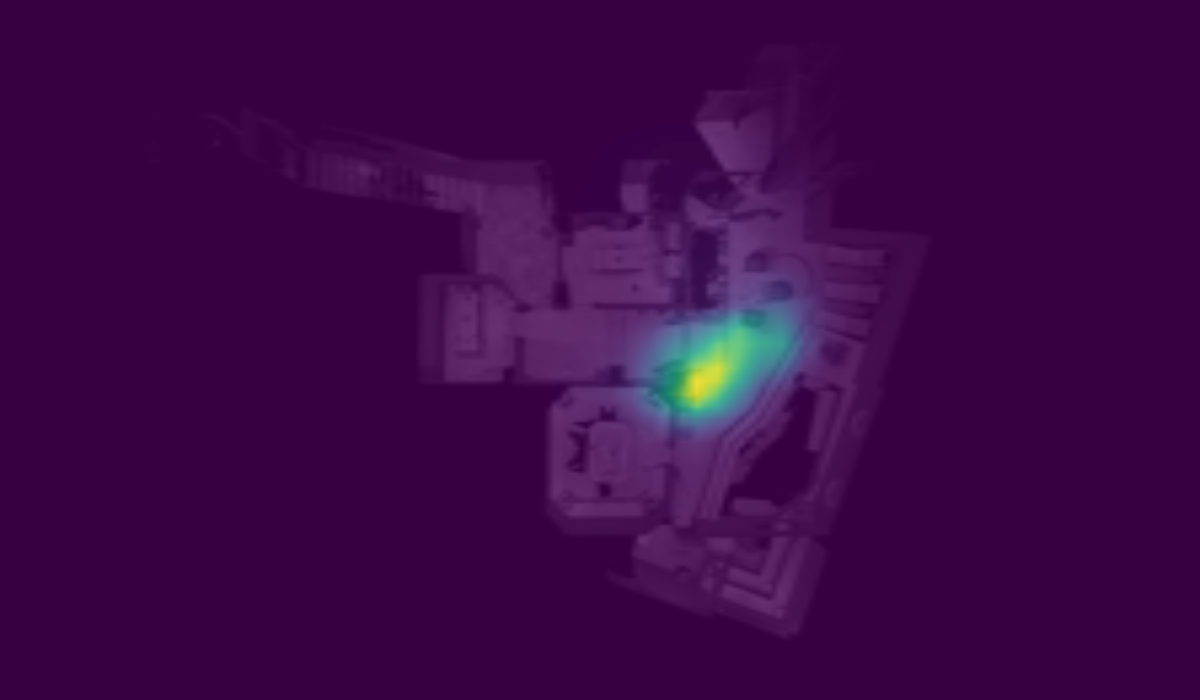} &
         \includegraphics[width=0.2\linewidth]{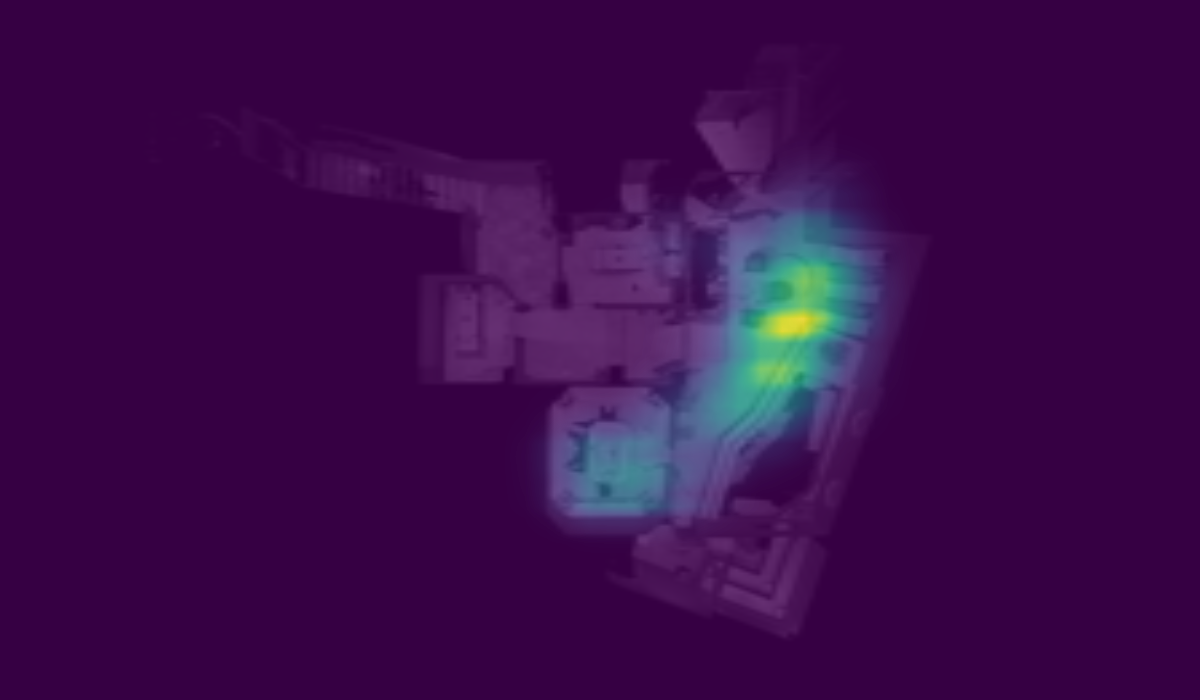} &
         \includegraphics[width=0.2\linewidth]{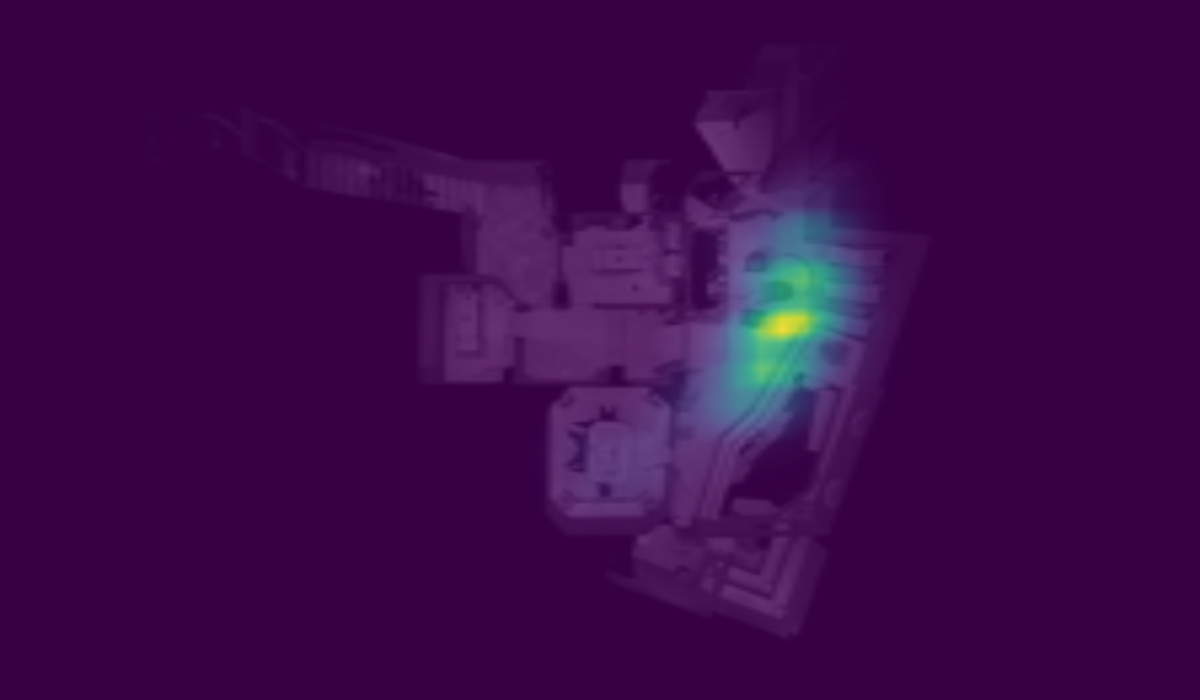} &
         \includegraphics[width=0.2\linewidth]{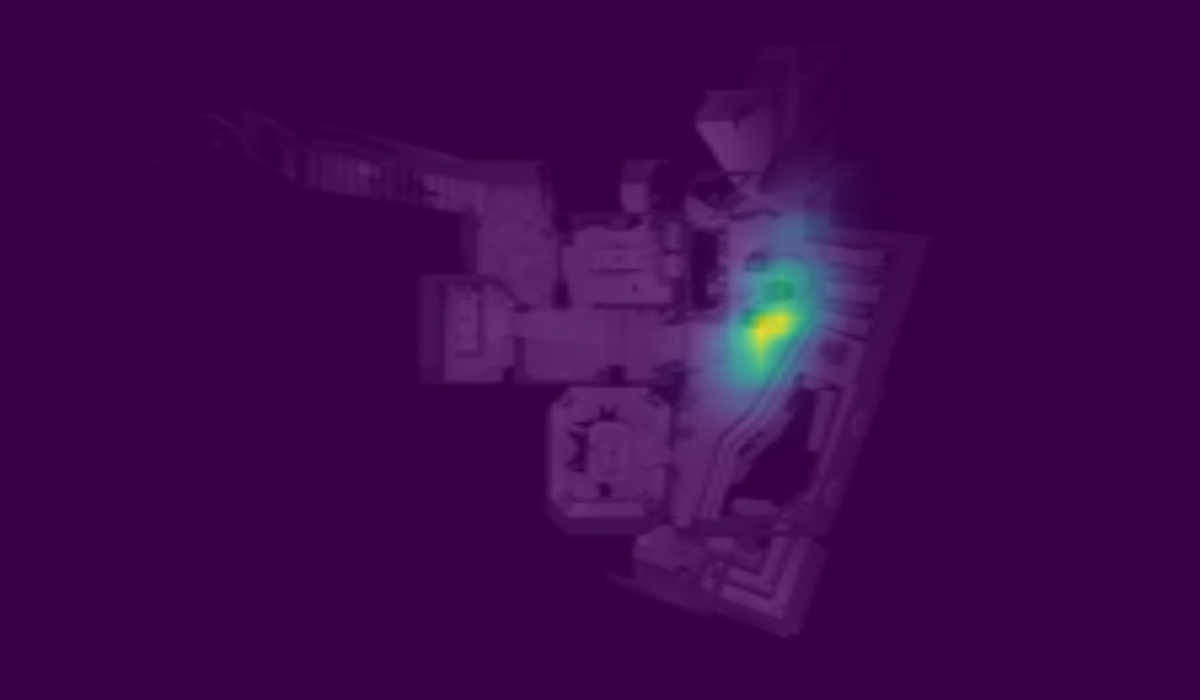} 
         \\
         & val-unseen 224 & Single-shot & Multi-shot: $1/3$ & $2/3$ & $3/3$ \\ \\

          % unseen, 327
         \rotatebox{90}{LingUNet} & \begin{tikzpicture}
            \draw (0, 0) node[inner sep=0] 
            {\includegraphics[width=0.2\linewidth]{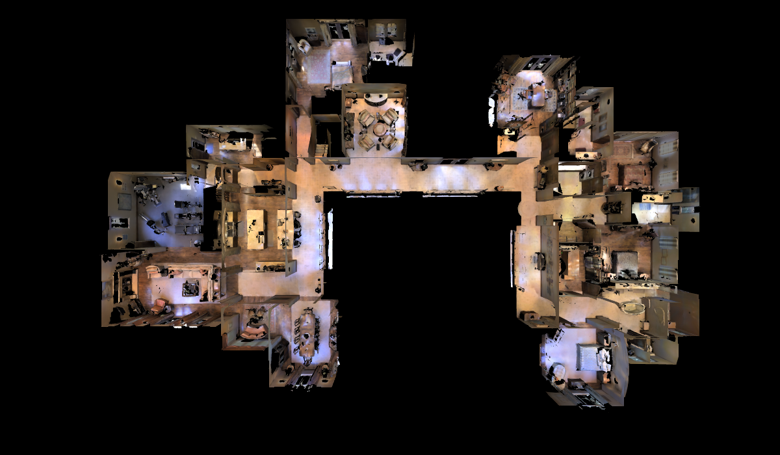}};
            \draw (1.4, 0.8) node[color=white,font=\small] {Map};
        \end{tikzpicture} &
         \includegraphics[width=0.2\linewidth]{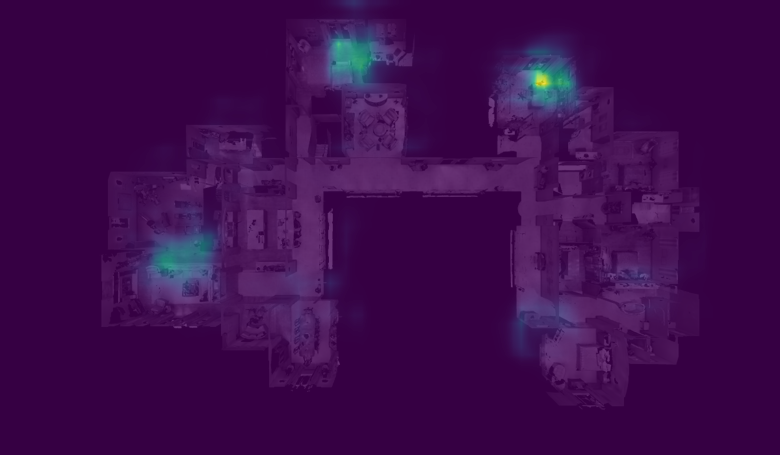} &
         \includegraphics[width=0.2\linewidth]{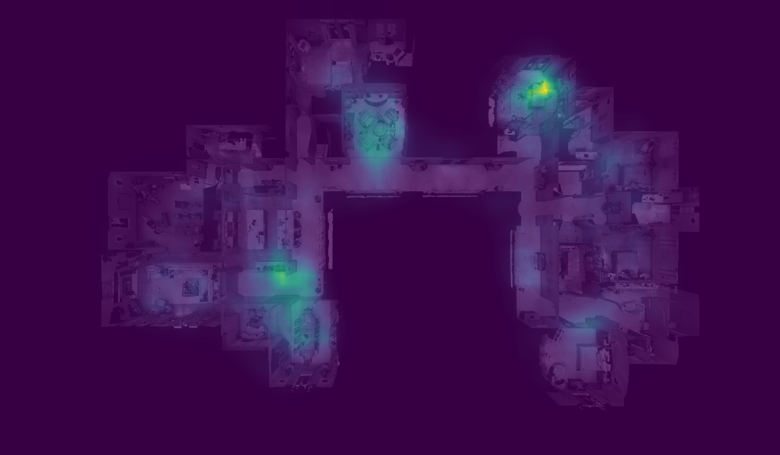} &
         \includegraphics[width=0.2\linewidth]{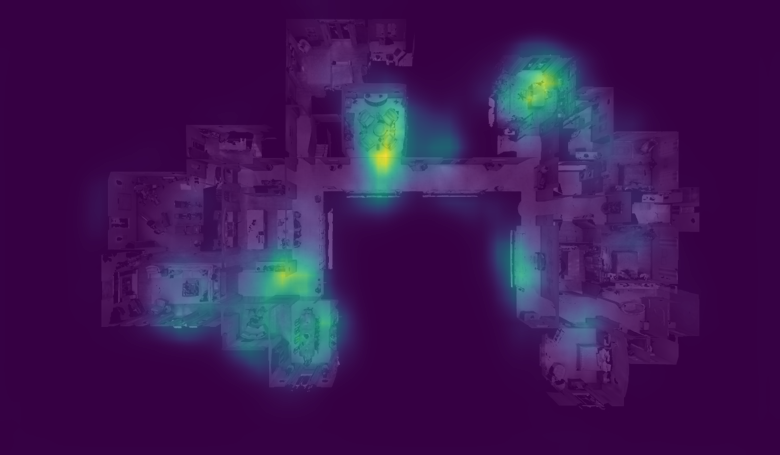} &
         \includegraphics[width=0.2\linewidth]{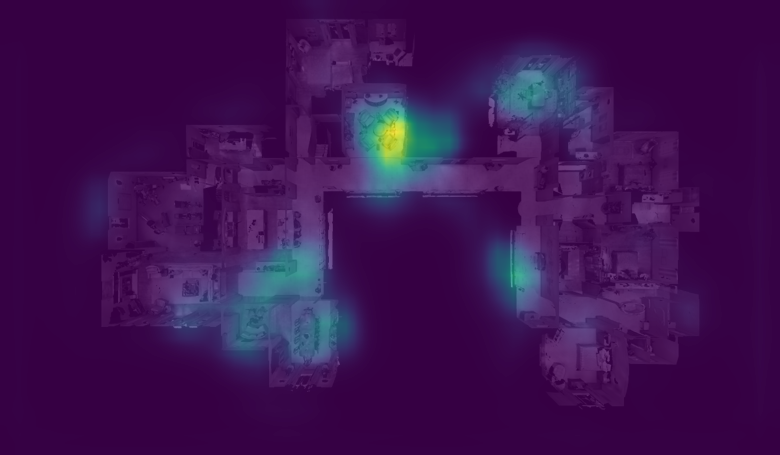} 
         \\
         \rotatebox{90}{DiaLoc} & \begin{tikzpicture}
            \draw (0, 0) node[inner sep=0] 
            {\includegraphics[width=0.2\linewidth]{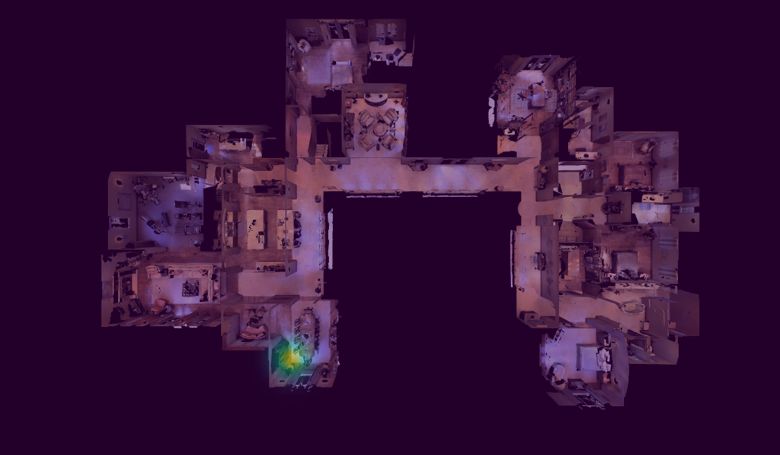}};
            \draw (1.4, 0.8) node[color=white,font=\small] {GT};
         \end{tikzpicture} &
         \includegraphics[width=0.2\linewidth]{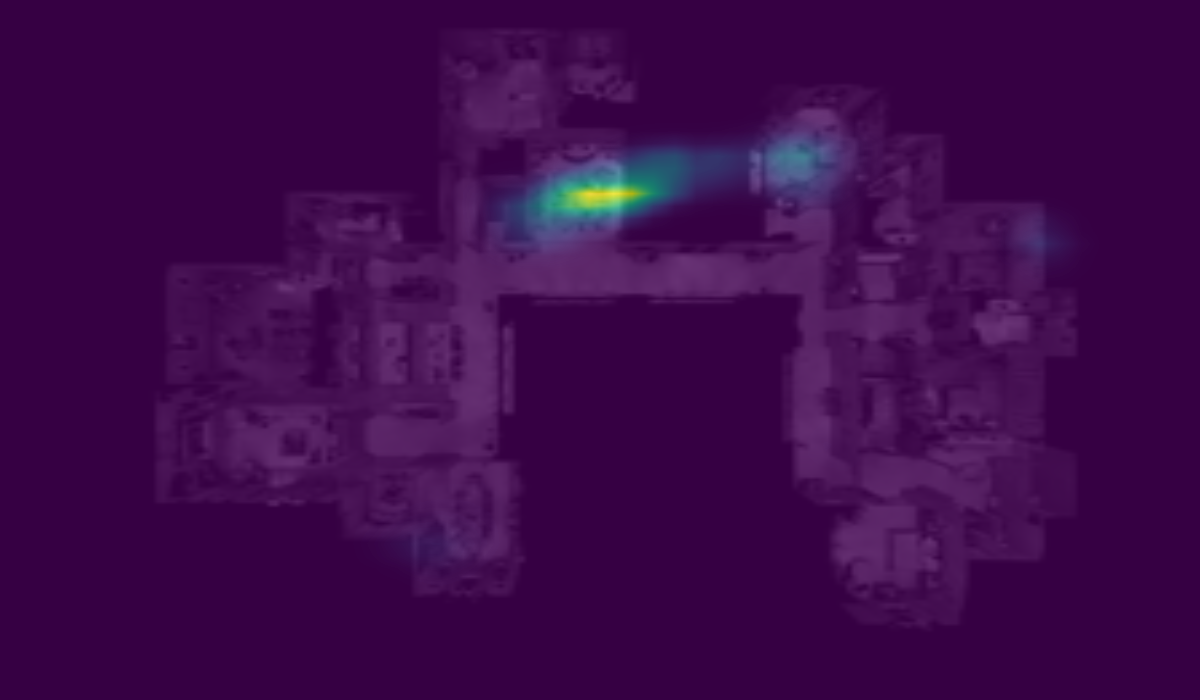} &
         \includegraphics[width=0.2\linewidth]{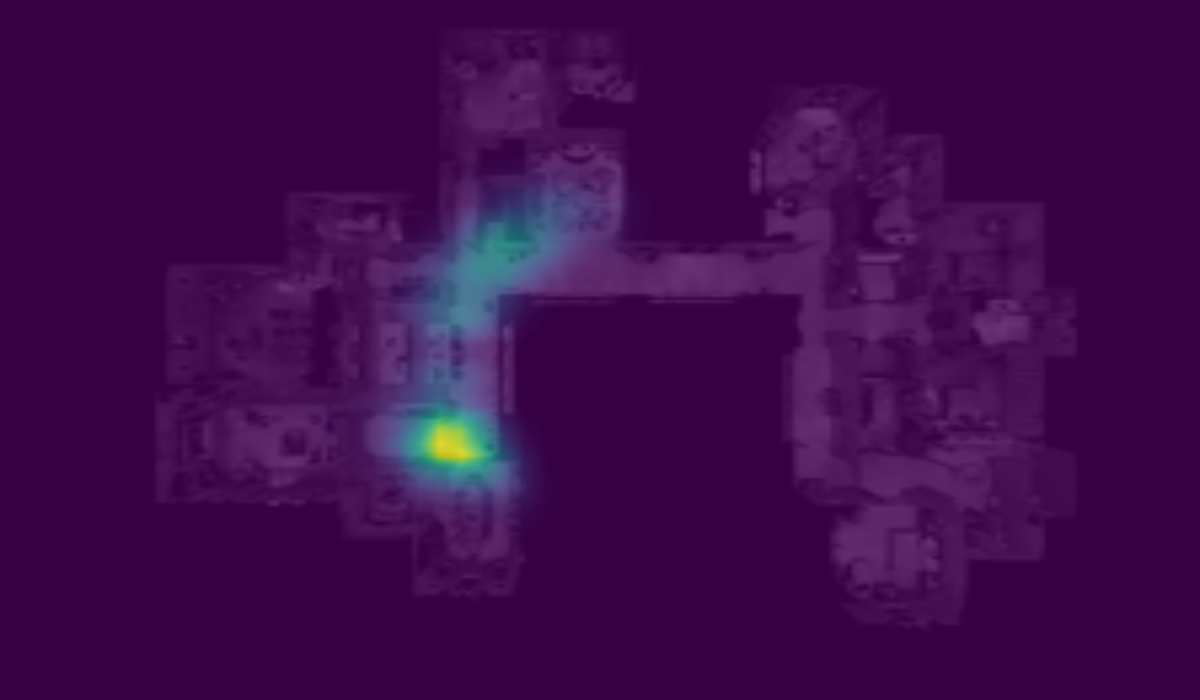} &
         \includegraphics[width=0.2\linewidth]{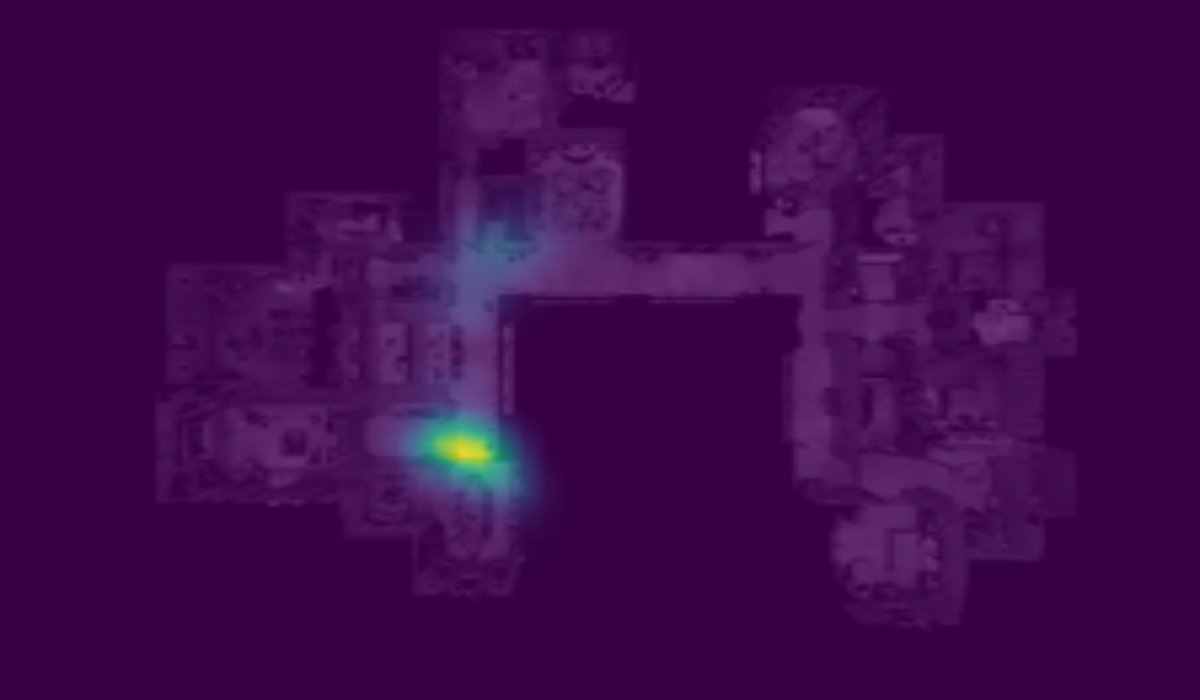} &
         \includegraphics[width=0.2\linewidth]{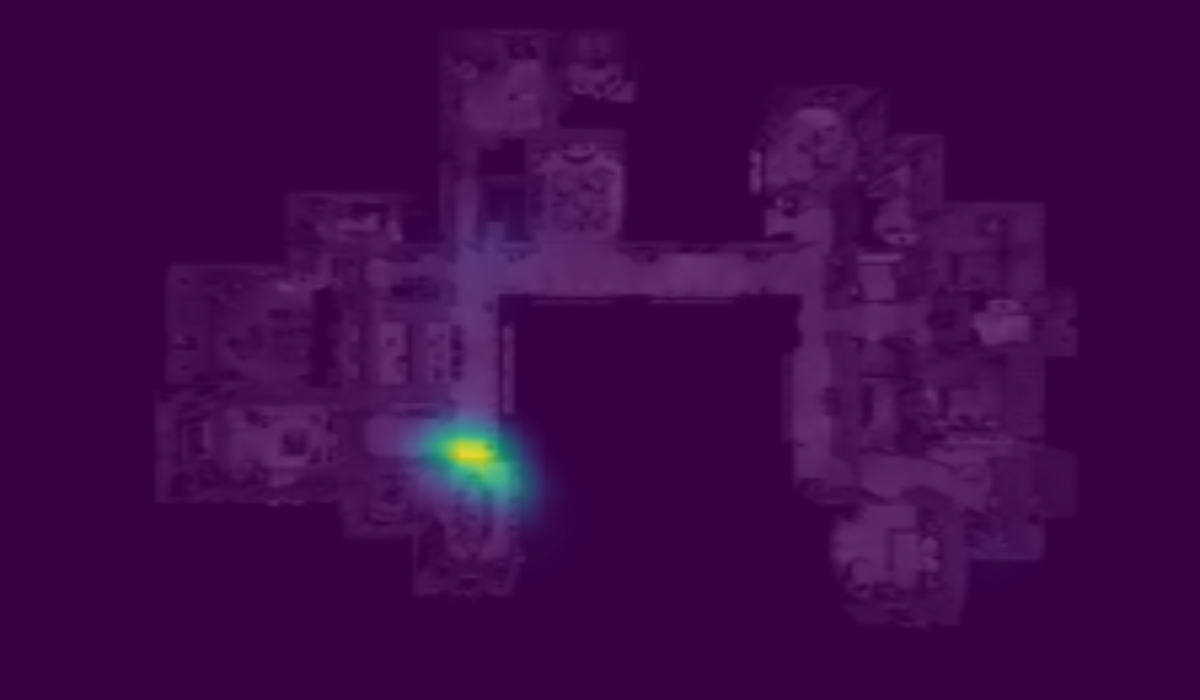} 
         \\
         & val-unseen 327 & Single-shot & Multi-shot: $1/3$ & $2/3$ & $3/3$ \\
    \end{tabular}
    \caption{\textbf{Qualitative results of single-shot and multi-shot location predictions are presented.} In the first column, the top-down map is displayed alongside its corresponding ground truth (GT) location. The second column displays the single-shot predictions, with LingUNet results above and DiaLoc results below. The last three columns showcase the multi-shot predictions.}
    \vspace{-10pt}
    \label{fig:quali2}
\end{figure*}

%%%%%%%%%%%%%%%%%%%%%%%%%%%%%%%%%%%%%%%%%%%%%%%%%%%
\begin{table}[t]
\centering
\begin{tabular}{c|c|c|c}
\hline

\textbf{Method} & \makecell{Image Size \\ (height x width)} & \makecell{Runtime \\(second)} & \makecell{GPU Usage\\ (MiB)} \\
\hline
LingUNet-ms & 455 x 780 & 0.768  & 1,273 \\
DiaLoc-e    & 224 x 224 & 1.374  & 3,875 \\
\hline
\end{tabular}
\vspace{5pt}
\caption{\textbf{Average runtime and memory usage comparison at inference time.} Batch size is set to 1 for both methods. A Nvidia Titan RTX 24gb is used for the benchmarking. }
\label{tab:runtime}
\end{table}
%%%%%%%%%%%%%%%%%%%%%%%%%%%%%%%%%%%%%%%%%%%%%%%%%%%
\paragraph{Additional visualizations.}
In Fig.\ref{fig:quali2}, we visualize additional localization predictions comparing LingUNet and the proposed method. For both methods, single-shot and multi-shot variants are evaluated. We now share our insights from these representative examples. 
\begin{itemize}
    \item \textbf{Val-seen 87:} DiaLoc predicts the correct location while LingUNet failed in the single-shot mode. LingUNet-ms produces noisy but correct predictions. In contrast, DiaLoc is capable of generating concentrated multi-modal (not to be confused with \emph{multimodal learning}) predictions. 
    \item \textbf{Val-seen 176:} Both approaches give acceptable predictions in the single-shot mode. In multi-shot mode, DiaLoc recovers from its initial incorrect prediction,  out of two possible guesses.
    \item \textbf{Val-unseen 224:} DiaLoc succeeds while LingUNet fails in the single-shot. In multi-shot mode, LingUNet-ms generates noisy predictions, while DiaLoc continually refining its prediction and converging towards the correct location in the end. 
    \item \textbf{Val-unseen 327:} In the single-shot mode, both methods failed. In the multi-shot mode, LingUNet-ms converged to a few locations, but none matches the GT. For our DiaLoc, the predictions are incrementally refined to the right area. 
\end{itemize}
%%%%%%%%%%%%%%%% Seen Quali %%%%%%%%%%%%%%%
%%%%%%%%%%%%%%%%%%%%%%%%%%%%%%%%%%%%%%

{
    \small
    \bibliographystyle{ieeenat_fullname}
    \bibliography{egbib}

\begin{thebibliography}{24}
\providecommand{\natexlab}[1]{#1}
\providecommand{\url}[1]{\texttt{#1}}
\expandafter\ifx\csname urlstyle\endcsname\relax
  \providecommand{\doi}[1]{doi: #1}\else
  \providecommand{\doi}{doi: \begingroup \urlstyle{rm}\Url}\fi

\bibitem[Alayrac et~al.(2022)Alayrac, Donahue, Luc, Miech, Barr, Hasson, Lenc,
  Mensch, Millican, Reynolds, et~al.]{alayrac2022flamingo}
Jean-Baptiste Alayrac, Jeff Donahue, Pauline Luc, Antoine Miech, Iain Barr,
  Yana Hasson, Karel Lenc, Arthur Mensch, Katherine Millican, Malcolm Reynolds,
  et~al.
\newblock Flamingo: a visual language model for few-shot learning.
\newblock \emph{Advances in Neural Information Processing Systems},
  35:\penalty0 23716--23736, 2022.

\bibitem[Anderson et~al.(2018)Anderson, Wu, Teney, Bruce, Johnson,
  S{\"u}nderhauf, Reid, Gould, and Van Den~Hengel]{anderson2018vision}
Peter Anderson, Qi Wu, Damien Teney, Jake Bruce, Mark Johnson, Niko
  S{\"u}nderhauf, Ian Reid, Stephen Gould, and Anton Van Den~Hengel.
\newblock Vision-and-language navigation: Interpreting visually-grounded
  navigation instructions in real environments.
\newblock In \emph{Proceedings of the IEEE conference on computer vision and
  pattern recognition}, pages 3674--3683, 2018.

\bibitem[Banerjee et~al.(2021)Banerjee, Thomason, and
  Corso]{banerjee2021robotslang}
Shurjo Banerjee, Jesse Thomason, and Jason Corso.
\newblock The robotslang benchmark: Dialog-guided robot localization and
  navigation.
\newblock In \emph{Conference on Robot Learning}, pages 1384--1393. PMLR, 2021.

\bibitem[Brown et~al.(2020)Brown, Mann, Ryder, Subbiah, Kaplan, Dhariwal,
  Neelakantan, Shyam, Sastry, Askell, et~al.]{brown2020language}
Tom Brown, Benjamin Mann, Nick Ryder, Melanie Subbiah, Jared~D Kaplan, Prafulla
  Dhariwal, Arvind Neelakantan, Pranav Shyam, Girish Sastry, Amanda Askell,
  et~al.
\newblock Language models are few-shot learners.
\newblock \emph{Advances in neural information processing systems},
  33:\penalty0 1877--1901, 2020.

\bibitem[Chang et~al.(2017)Chang, Dai, Funkhouser, Halber, Niessner, Savva,
  Song, Zeng, and Zhang]{chang2017matterport3d}
Angel Chang, Angela Dai, Thomas Funkhouser, Maciej Halber, Matthias Niessner,
  Manolis Savva, Shuran Song, Andy Zeng, and Yinda Zhang.
\newblock Matterport3d: Learning from rgb-d data in indoor environments.
\newblock \emph{arXiv preprint arXiv:1709.06158}, 2017.

\bibitem[Devlin et~al.(2018)Devlin, Chang, Lee, and Toutanova]{devlin2018bert}
Jacob Devlin, Ming-Wei Chang, Kenton Lee, and Kristina Toutanova.
\newblock Bert: Pre-training of deep bidirectional transformers for language
  understanding.
\newblock \emph{arXiv preprint arXiv:1810.04805}, 2018.

\bibitem[Driess et~al.(2023)Driess, Xia, Sajjadi, Lynch, Chowdhery, Ichter,
  Wahid, Tompson, Vuong, Yu, et~al.]{driess2023palm}
Danny Driess, Fei Xia, Mehdi~SM Sajjadi, Corey Lynch, Aakanksha Chowdhery,
  Brian Ichter, Ayzaan Wahid, Jonathan Tompson, Quan Vuong, Tianhe Yu, et~al.
\newblock Palm-e: An embodied multimodal language model.
\newblock \emph{arXiv preprint arXiv:2303.03378}, 2023.

\bibitem[Hahn and Rehg(2022)]{hahn2022transformer}
Meera Hahn and James~M Rehg.
\newblock Transformer-based localization from embodied dialog with large-scale
  pre-training.
\newblock \emph{arXiv preprint arXiv:2210.04864}, 2022.

\bibitem[Hahn et~al.(2020)Hahn, Krantz, Batra, Parikh, Rehg, Lee, and
  Anderson]{hahn2020you}
Meera Hahn, Jacob Krantz, Dhruv Batra, Devi Parikh, James~M Rehg, Stefan Lee,
  and Peter Anderson.
\newblock Where are you? localization from embodied dialog.
\newblock \emph{arXiv preprint arXiv:2011.08277}, 2020.

\bibitem[Huang et~al.(2015)Huang, Xu, and Yu]{huang2015bidirectional}
Zhiheng Huang, Wei Xu, and Kai Yu.
\newblock Bidirectional lstm-crf models for sequence tagging.
\newblock \emph{arXiv preprint arXiv:1508.01991}, 2015.

\bibitem[Kolmet et~al.(2022)Kolmet, Zhou, O{\v{s}}ep, and
  Leal-Taix{\'e}]{kolmet2022text2pos}
Manuel Kolmet, Qunjie Zhou, Aljo{\v{s}}a O{\v{s}}ep, and Laura Leal-Taix{\'e}.
\newblock Text2pos: Text-to-point-cloud cross-modal localization.
\newblock In \emph{Proceedings of the IEEE/CVF Conference on Computer Vision
  and Pattern Recognition}, pages 6687--6696, 2022.

\bibitem[Li et~al.(2022{\natexlab{a}})Li, Weinberger, Belongie, Koltun, and
  Ranftl]{li2022languagedriven}
Boyi Li, Kilian~Q Weinberger, Serge Belongie, Vladlen Koltun, and Rene Ranftl.
\newblock Language-driven semantic segmentation.
\newblock In \emph{International Conference on Learning Representations},
  2022{\natexlab{a}}.

\bibitem[Li et~al.(2021)Li, Selvaraju, Gotmare, Joty, Xiong, and
  Hoi]{li2021align}
Junnan Li, Ramprasaath Selvaraju, Akhilesh Gotmare, Shafiq Joty, Caiming Xiong,
  and Steven Chu~Hong Hoi.
\newblock Align before fuse: Vision and language representation learning with
  momentum distillation.
\newblock \emph{Advances in neural information processing systems},
  34:\penalty0 9694--9705, 2021.

\bibitem[Li et~al.(2022{\natexlab{b}})Li, Li, Xiong, and Hoi]{li2022blip}
Junnan Li, Dongxu Li, Caiming Xiong, and Steven Hoi.
\newblock Blip: Bootstrapping language-image pre-training for unified
  vision-language understanding and generation.
\newblock In \emph{International Conference on Machine Learning}, pages
  12888--12900. PMLR, 2022{\natexlab{b}}.

\bibitem[Liao et~al.(2022)Liao, Xie, and Geiger]{liao2022kitti}
Yiyi Liao, Jun Xie, and Andreas Geiger.
\newblock Kitti-360: A novel dataset and benchmarks for urban scene
  understanding in 2d and 3d.
\newblock \emph{IEEE Transactions on Pattern Analysis and Machine
  Intelligence}, 45\penalty0 (3):\penalty0 3292--3310, 2022.

\bibitem[Loshchilov and Hutter(2017)]{loshchilov2017decoupled}
Ilya Loshchilov and Frank Hutter.
\newblock Decoupled weight decay regularization.
\newblock \emph{arXiv preprint arXiv:1711.05101}, 2017.

\bibitem[Manuel et~al.(2022)Manuel, Faied, Krishnan, and
  Paulik]{manuel2022robot}
Melvin~P Manuel, Mariam Faied, Mohan Krishnan, and Mark Paulik.
\newblock Robot platooning strategy for search and rescue operations.
\newblock \emph{Intelligent Service Robotics}, pages 1--12, 2022.

\bibitem[Misra et~al.(2018)Misra, Bennett, Blukis, Niklasson, Shatkhin, and
  Artzi]{misra2018mapping}
Dipendra Misra, Andrew Bennett, Valts Blukis, Eyvind Niklasson, Max Shatkhin,
  and Yoav Artzi.
\newblock Mapping instructions to actions in 3d environments with visual goal
  prediction.
\newblock \emph{arXiv preprint arXiv:1809.00786}, 2018.

\bibitem[Queralta et~al.(2020)Queralta, Taipalmaa, Pullinen, Sarker, Gia,
  Tenhunen, Gabbouj, Raitoharju, and Westerlund]{queralta2020collaborative}
Jorge~Pena Queralta, Jussi Taipalmaa, Bilge~Can Pullinen, Victor~Kathan Sarker,
  Tuan~Nguyen Gia, Hannu Tenhunen, Moncef Gabbouj, Jenni Raitoharju, and Tomi
  Westerlund.
\newblock Collaborative multi-robot search and rescue: Planning, coordination,
  perception, and active vision.
\newblock \emph{Ieee Access}, 8:\penalty0 191617--191643, 2020.

\bibitem[Radford et~al.(2021)Radford, Kim, Hallacy, Ramesh, Goh, Agarwal,
  Sastry, Askell, Mishkin, Clark, et~al.]{radford2021learning}
Alec Radford, Jong~Wook Kim, Chris Hallacy, Aditya Ramesh, Gabriel Goh,
  Sandhini Agarwal, Girish Sastry, Amanda Askell, Pamela Mishkin, Jack Clark,
  et~al.
\newblock Learning transferable visual models from natural language
  supervision.
\newblock In \emph{International conference on machine learning}, pages
  8748--8763. PMLR, 2021.

\bibitem[Shah et~al.(2023)Shah, Osi{\'n}ski, Levine, et~al.]{shah2023lm}
Dhruv Shah, B{\l}a{\.z}ej Osi{\'n}ski, Sergey Levine, et~al.
\newblock Lm-nav: Robotic navigation with large pre-trained models of language,
  vision, and action.
\newblock In \emph{Conference on Robot Learning}, pages 492--504. PMLR, 2023.

\bibitem[Thomason et~al.(2020)Thomason, Murray, Cakmak, and
  Zettlemoyer]{thomason2020vision}
Jesse Thomason, Michael Murray, Maya Cakmak, and Luke Zettlemoyer.
\newblock Vision-and-dialog navigation.
\newblock In \emph{Conference on Robot Learning}, pages 394--406. PMLR, 2020.

\bibitem[Vaswani et~al.(2017)Vaswani, Shazeer, Parmar, Uszkoreit, Jones, Gomez,
  Kaiser, and Polosukhin]{vaswani2017attention}
Ashish Vaswani, Noam Shazeer, Niki Parmar, Jakob Uszkoreit, Llion Jones,
  Aidan~N Gomez, {\L}ukasz Kaiser, and Illia Polosukhin.
\newblock Attention is all you need.
\newblock \emph{Advances in neural information processing systems}, 30, 2017.

\bibitem[Wang et~al.(2023)Wang, Bao, Dong, Bjorck, Peng, Liu, Aggarwal,
  Mohammed, Singhal, Som, et~al.]{wang2023image}
Wenhui Wang, Hangbo Bao, Li Dong, Johan Bjorck, Zhiliang Peng, Qiang Liu, Kriti
  Aggarwal, Owais~Khan Mohammed, Saksham Singhal, Subhojit Som, et~al.
\newblock Image as a foreign language: Beit pretraining for vision and
  vision-language tasks.
\newblock In \emph{Proceedings of the IEEE/CVF Conference on Computer Vision
  and Pattern Recognition}, pages 19175--19186, 2023.

\end{thebibliography}
}


\end{document}